\UseRawInputEncoding
\documentclass[11pt]{article}
\usepackage{CJKnumb}
\usepackage{amsmath, amsfonts, bm,cite}
\usepackage{newtxtext} 
\usepackage{newtxmath} 
\usepackage{graphicx}
\usepackage{color}
\usepackage{multirow}
\usepackage{fancyhdr}
\usepackage{array}
\usepackage{makeidx, chngpage}
\usepackage{enumerate}
\usepackage{titlesec}
\usepackage[T1]{fontenc}
\usepackage{float}
\usepackage{url}
\usepackage{caption}
\usepackage{booktabs}
\usepackage{ascii}
\usepackage[margin=1.5cm]{geometry}
\usepackage{bm}
\usepackage{indentfirst}
\usepackage{textcomp}
\usepackage{graphicx}
\usepackage{hyperref} 
\hypersetup{
    colorlinks=true,        
    linkcolor=blue,         
    citecolor=blue,          
    urlcolor=purple         
}
\usepackage{mwe}
\usepackage{subfigure}
\usepackage{appendix}
\usepackage{mathtools}

\usepackage{amstext}
\usepackage{fancyhdr}
\usepackage{tabularx}

\usepackage{amsthm}
\usepackage{pifont}
\newcommand{\cmark}{\ding{51}}%
\newcommand{\xmark}{\ding{55}}%

\newtheorem{assumption}{Assumption}

\usepackage{lineno}
\nolinenumbers

\RequirePackage[table]{xcolor} 
\RequirePackage{soul} 
\RequirePackage{multirow}

\newlength{\defbaselineskip}
\newcommand{\bitem}{\begin{itemize}} 
\newcommand{\eitem}{\end{itemize}}
\newcommand{\be}{\begin{equation}}
\newcommand{\ee}{\end{equation}}
\newcommand{\benu} {\begin{enumerate}}
\newcommand{\eenu} {\end{enumerate}}
\newcommand{\ba}{\begin{eqnarray}}
\newcommand{\ea}{\end{eqnarray}}
\newcommand{\bdes}{\begin{description}}
\newcommand{\edes}{\end{description}}

\title{A Continual Validation, Updating, and Decision-Making Framework for \\ Self-Adaptive Digital Twins: A Case Study in Additive Manufacturing}

\author{Yi-Ping Chen$^{1}$, Ying-Kuan Tsai$^{1,2}$, Seul Lee$^{1}$, Vispi Karkaria$^{1}$, Daniel Apley$^{3}$, and Wei Chen$^{1}$ \thanks{Corresponding author, weichen@northwestern.edu, Department of Mechanical Engineering, Northwestern University, Evanston, IL, 60208}\\ \small{$^{1}$ Department of Mechanical Engineering, Northwestern University}\\
\small{$^{2}$ Ingram School of Engineering, Texas State University}\\
\small{$^{3}$ Department of Industrial Engineering and Management Science, Northwestern University}\\
}

\date{}

\begin{document}

\graphicspath{{Graphs/}}

\maketitle

\begin{abstract}

\textcolor{black}{Digital Twins support real-time decision-making through predictive models of physical systems, yet their reliability can deteriorate as operating conditions evolve. Maintaining predictive fidelity under such changes is challenging, particularly when model adaptation relies on limited streaming data and updated models must be evaluated before deployment. Here, we present an online adaptation framework that integrates gradient-based drift detection, parameter-efficient fine-tuning, and statistical model validation within model predictive control. Changes in prediction-loss gradients trigger batch-wise model updates using Low-Rank Adaptation, while a paired moving-block bootstrap test evaluates predictive improvement before deploying the updated model. The framework is evaluated using a stochastic dynamical system with abrupt changes and a simulated directed energy deposition process with gradual changes in material properties. Results show improvements in predictive accuracy and uncertainty estimation during evaluated post-adaptation intervals, alongside improved reference tracking and reduced constraint violations in the manufacturing case study compared with the investigated baselines. These findings demonstrate the potential of integrating continual validation with model adaptation while acknowledging the trade-off between responsiveness and predictive reliability in evolving Digital Twins.}

\end{abstract}
\textbf{Keywords}: Digital Twin; Concept Drift Detection; Continual Learning; Parameter-Efficient Fine-Tuning; Online Validation; Robust Model Predictive Control; Trustworthy Machine Learning

\tableofcontents



\section{Introduction}

\subsection{Challenges of Maintaining Model Accuracy for Digital Twins}

A Digital Twin is a virtual representation of a physical system that continuously synchronizes and evolves with its physical counterpart through bidirectional data exchange, reflecting the current state, history, and anticipated behavior of the system \cite{grieves2016digital, national2023foundational, van2023digital, karkaria2025optimization}. By assimilating streaming sensor data, it supports real-time state estimation and multi-step prediction for informed decision-making under uncertainty \cite{galvez2025decision, chen2026uncertainty, hamzah2026modular}, including examples such as performance optimization \cite{chen2025real}, fault diagnosis \cite{zheng2022emergence}, predictive maintenance \cite{dwight2025maintenance, karkaria2025ai}, and adaptive control \cite{chen2025adaptive}. Recent advances in machine learning provide expressive, data-driven surrogates of complex nonlinear dynamics \cite{thelen2022comprehensive, karkaria2024towards, yu2024learning}, enabling rapid evaluation of ``what-if'' scenarios without costly high-fidelity simulation \cite{galvez2025decision,lee20262026} and turning Digital Twins from static monitors into proactive, intelligent systems.

\textcolor{black}{What differentiates Digital Twins from Digital Shadows or Digital Models is their capability of supporting real-time decision-making \cite{national2023foundational}}. When integrated with decision-making frameworks such as dynamic programming \cite{karkaria2025ai}, reinforcement learning (RL) \cite{tsai2026digital,tsai2026reinforcement}, or model predictive control (MPC) \cite{chen2025real}, the Digital Twin becomes a closed-loop, real-time optimization engine in which the surrogate is the predictive core subject to operational constraints and safety requirements \cite{es2024methods}. Unlike feedback control that relies on error correction, such frameworks perform proactive optimization over a horizon. MPC's receding-horizon structure \cite{rawlings2017model, hewing2020learning} complements this paradigm: at each step, the virtual twin supplies state estimates and predictions, MPC solves a constrained optimization for control actions \cite{chen2025real, chen2026uncertainty}, and continual re-optimization adapts to disturbances and evolving conditions. MPC's repeated real-time optimization places stringent demands on model efficiency, so conventional formulations often rely on linear state-space models solved via convex quadratic programming (QP) \cite{tsai2023robust,tsai2025control}. Increasingly, neural networks (NNs) and other nonlinear approximators surrogate complex dynamics \cite{park2023simultaneous, huang2022lstm, jung2023model, wu2020process}, expanding MPC to highly nonlinear, spatio-temporal processes and broadening Digital Twin systems. 

\textcolor{black}{The effectiveness of model-based decision-making also critically depends on the fidelity of the predictive model. Therefore, maintaining synchronization between the virtual counterpart of the Digital Twin and the physical system is essential \cite{dutta2021new, de2026digital}. In practice, state observations are often partial and limited \cite{chen2025targeted}, for example due to sparse sensing, measurement noise, unmeasured disturbances, or latent material and environmental variations. When such factors cannot be explicitly incorporated as state variables, or reliably inferred from available sensor data, model-plant mismatch inevitably arises. These uncertainties may stem from state-dependent parameter variations \cite{chen2021unknown}, unmodeled dynamics \cite{hewing2019cautious, ren2022bias}, structural simplifications, or gradual system degradation over time \cite{huang2023novel}. As a result, static models become progressively misaligned with the evolving physical process. This challenge has motivated methods that aim to continuously update or correct the system model during operation to improve control performance.}

\textcolor{black}{Continual adaptation of predictive models has been studied from two related perspectives. The first is the emerging literature on \emph{self-adaptive Digital Twins}, which explicitly considers how the virtual model should evolve as the physical system changes over its life cycle. Existing studies have adopted different adaptation strategies. Some first detect changes in the system or model performance and trigger model updating only when adaptation is considered necessary \cite{modesto2025towards,karkaria2025ai,gupta2024twin}. For instance, Karkaria \emph{et al.} \cite{karkaria2025ai} developed a Digital Twin framework integrating maximum concept discrepancy (MCD) to detect distributional shift for the newly available data, and fine-tune the model for the remaining useful life (RUL) of tires to support decision-making for predictive maintenance. Similarly, Modesto \emph{et al.} \cite{modesto2025towards} implemented the Kolmogorov-Smirnov moving window (KSWIN) to detect covariate drift to further trigger the updating of the surrogate model of the transportation systems, which is represented by graph neural networks (GNNs). Others update the predictive model whenever new operational data become available \cite{chen2025adaptive, fu2024development, tang2026lifelong}. For example, Tang \emph{et al.} \cite{tang2026lifelong} map the response of the system into a latent space via an autoencoder, and learn the degradation pattern using LSTM to guide the lifelong updating of the Digital Twin surrogate, and Chen \emph{et al.} \cite{chen2025adaptive} use the immediate feedback data from the physical measurement from the incremental forming process to update the predictive model for sheet deformation, modeled by a Koopman operator, using recursive least squares. Despite these recent efforts, research on self-adaptation of learning-based Digital Twins within a closed-loop decision-making framework remains relatively limited, and existing studies typically address only individual aspects of the adaptation process rather than enabling autonomous and continual operation throughout the system life cycle.}

\textcolor{black}{The second is a much broader body of related applications found in \emph{adaptive MPC}, where the predictive model is continuously synchronized with the physical system using streaming measurements \cite{patrinos2013accelerated,heirung2015mpc}. For conventional MPC, this synchronization can often be achieved efficiently because the underlying models have relatively compact parameterizations that can be updated using methods such as Kalman filtering \cite{vafamand2022adaptive} or recursive least squares \cite{calderon2021koopman}. Extending this paradigm to learning-based MPC and Digital Twins is substantially more difficult when the predictive model is a highly parameterized DNN. Updating such models at every time step can introduce considerable computational cost and, when only limited streaming data are available, may lead to overfitting or catastrophic forgetting. Moreover, adaptive MPC primarily seeks to maintain satisfactory closed-loop control performance. Consequently, adaptation can favor a model that is sufficiently accurate within the current operating region for local control \cite{mamakoukas2019local}, without necessarily preserving the broader predictive fidelity expected from a Digital Twin across different operating regimes.}

\textcolor{black}{These observations conclude that a self-adapting Digital Twin must autonomously manage the predictive model throughout its operational life cycle, including determining when adaptation is necessary, updating the model reliably from limited streaming data, and verifying that the updated model is sufficiently trustworthy before deployment. Based on these requirements, three closely coupled tasks remain to be addressed simultaneously:}
    
\begin{itemize}
    \item \emph{When to update the model}: For learning-based predictive models, updating at every time step can introduce substantial computational overhead and a risk of overfitting, potentially hindering real-time implementation and stability. Conversely, infrequent updates may delay the response to distributional drift, leading to performance degradation and reduced robustness \cite{gou2024sample}. Therefore, a drift detection mechanism is needed to determine \emph{when} model updates should be triggered, balancing computational efficiency with adaptability to evolving system dynamics. \textcolor{black}{For autonomous Digital Twin operation, such a mechanism should also accommodate multivariate model outputs and allow its detection criterion to be automatically reset or recalibrated after adaptation so that monitoring can continue throughout the system life cycle.}

    \item \emph{How to update the model}: Updating a highly parameterized model, such as a DNN with numerous tunable weights, from limited streaming data is inherently challenging. In particular, when the incoming data undergo abrupt changes or contain measurement noise, naive iterative fine-tuning can cause unstable adaptation, overfitting to corrupted samples, or catastrophic forgetting, i.e., erosion of previously acquired knowledge and degradation of performance on previously well-modeled operating regimes. Data- and parameter-efficient fine-tuning strategies are therefore needed for DNN-based MPC.

    \item \emph{Is the update successful}: Digital Twins must be continuously validated throughout their operation and life cycle \cite{national2023foundational}, especially when the virtual counterpart relies on a data-driven model updated from limited batches of streaming data. Given the aforementioned risks, deploying an inadequately validated updated model for safety-critical decision-making can lead to severe performance degradation or even system-level failure. Systematic post-update validation is therefore indispensable to ensure reliability and trustworthiness.
\end{itemize}

\subsection{When to Update the Digital Twin? -- Updating Models in Existing Closed-Loop Systems}

Determining when to update the model in real time intelligently remains an open challenge \cite{abbasi2024understanding, rabczuk2026scientific}. Update strategies are broadly instance-wise, batch-wise, periodic, or event-triggered. Instance-wise schemes update at every step via gradient descent \cite{hedjar2013adaptive, akpan2011nonlinear, wu2019real, casagrande2023online}, recursive least squares \cite{akpan2011nonlinear}, or online Gaussian-process updates \cite{wu2023adaptive, hewing2019cautious}. These are responsive but raise memory cost and degrade long-term recall \cite{abdoune2025digital}, motivating cost-aware updating \cite{gou2024sample}. Batch-wise and periodic schemes retrain after a fixed number of samples \cite{bieker2020deep, bao2020online} or at a fixed interval \cite{yang2020model}, sometimes with threshold-based retraining of lightweight models \cite{desale2023concept, yang2021lightweight}. These improve efficiency but can delay the response to abrupt changes. Event-triggered schemes, by integrating drift detection mechanisms, update only when drift is detected \cite{felipe2020drift}. 

\textcolor{black}{To determine when model adaptation is necessary, concept drift detection provides a systematic approach to identifying changes in the underlying data distribution. A substantial body of research has investigated concept drift detection in machine learning and statistical process monitoring \cite{gama2014survey, lu2018learning, hoens2012learning}}. Let $\mathcal{D}_t = \{(x_i, y_i)\}_{i=1}^{n}$ denote the joint data distribution at time $t$, where $x \in \mathbb{R}^d$ is the input vector and $y \in \mathbb{R}^m$ is the output. Concept drift refers to a change in the distribution $P_t(x, y)$ over time \cite{gama2014survey}:
\begin{equation}
    \exists\, t_0 < t_1 : \quad P_{t_0}(x, y) \neq P_{t_1}(x, y).
    \label{eq:drift_def}
\end{equation}

In practice, the broad family of drifts can originate from a shift in the input marginal distribution $p_t(x)$ only (covariate shift), a change in the conditional output distribution $p_t(y \mid x)$ (concept drift), or both simultaneously. In this work, we are only focusing on concept drift, where the relationship between the input and the output has changed.

{\color{black}Existing concept drift detection methods can be broadly categorized into error-based and distribution-based approaches. Error-based methods, such as the Drift Detection Method (DDM) \cite{gonccalves2014comparative}, Exponentially Weighted Moving Average (EWMA) \cite{crowder1992ewma}, and Cumulative Sum (CUSUM) \cite{biau2007quality}, monitor changes in prediction errors or related statistics. While computationally efficient for streaming data, these methods often rely on scalar statistics that may obscure changes in individual components of multivariate outputs. They may also exhibit delayed detection when distributional changes do not immediately increase prediction errors, such as under gradual drift \cite{zhang2023concept}. Alternatively, distribution-based methods, including Adaptive Windowing (ADWIN) \cite{bifet2007learning}, Maximum Mean Discrepancy (MMD) \cite{borgwardt2006integrating}, and the Kolmogorov--Smirnov test, identify drift by comparing statistical properties of reference and recent data windows. Although these methods can capture distributional changes beyond prediction errors, their computational and sample requirements may increase with dimensionality, while window-based comparisons can struggle with gradual drift \cite{desale2023concept, yang2021lightweight}. These limitations become more pronounced when monitoring models that predict multiple statistical quantities, such as quantile regression models, where a single observation cannot provide a separate ground-truth target for each predicted quantile. Consequently, monitoring such models requires a mechanism that can jointly assess changes across multiple predictive outputs using the limited feedback available during operation.}

\textcolor{black}{An alternative approach proposed by Zhang and Apley \cite{zhang2023concept} tracks the Fisher score vector, i.e.\ the gradient of the log-likelihood with respect to the model parameters, enabling parameter-level attribution of not only \emph{when} drift occurs but also \emph{which} parts of the input--output relationship have changed. By monitoring the gradient of a likelihood-based objective, this approach can capture changes in the underlying predictive relationship without requiring a separate observation for each model output. More broadly, the score-based monitoring principle can be extended to differentiable supervised learning loss, provided that the corresponding loss can be evaluated using streaming observations. This makes it applicable to models with multiple outputs.}

Moreover, many existing applications with drift detectors, however, assume that operation can be halted and the detector reinitialized on identified "normal" data \cite{zhao2007adaptive}, which is impractical for autonomous, continuously operating Digital Twins and anchors the detector to outdated concepts. This motivates continual, data-efficient online re-initialization and, more broadly, the integration of formal drift detection with continual decision-making as a critical element for proactive, drift-aware decision-making in Digital Twins.

\subsection{How to Update the Digital Twin?}

\subsubsection{Catastrophic Forgetting and Continual Learning}

Once drift is detected, adapting the virtual twin to a new distribution must accommodate evolving dynamics while avoiding overfitting to limited streaming data and undesirable loss of useful prior structure \cite{desale2023concept, kirkpatrick2017overcoming, aleixo2023catastrophic}. Catastrophic forgetting is studied under continual learning, where regularization-, replay-, optimization-, and representation-based methods mitigate forgetting. These mostly preserve performance across multiple tasks, whereas under concept drift, the aim is rather to overwrite outdated knowledge for plasticity. Here, "forgetting" for the purpose of adapting to new concepts means erasing still-relevant structural representations (i.e., the physics or the pre-trained network as the backbones) and degrading robustness.

Recent advances in parameter-efficient fine-tuning (PEFT) provide a promising direction for efficient model updating from streaming data \cite{han2024parameter}. PEFT methods adapt large NNs by updating only a subset or a low-dimensional subspace of the parameters, reducing computational cost while improving training stability. Among these, Low-Rank Adaptation (LoRA) \cite{hu2022lora} is a representative reparameterization-based technique that introduces trainable low-rank matrices in parallel to the pretrained weights, offering parameter efficiency, low inference latency, and stable optimization. Despite their success in large-scale learning tasks, to the best of our knowledge, the application of such methods to NN-based model adaptation for dynamical systems and the real-time decision-making of Digital Twins has not yet been investigated.

\subsubsection{Adaptation in MPC and Digital Twins}
From a control-design perspective, adaptive MPC can be realized in three principal forms \cite{sun2024adaptive}. The first updates the predictive model directly via online system identification \cite{na2020output}. The second adapts controller parameters, such as tuning the weighting matrices in nonlinear MPC \cite{kostadinov2020online} or analytically adjusting the control horizon \cite{turki2017analytical}, without modifying the underlying model. The third integrates RL with MPC, where RL tunes the control parameters online while preserving the MPC structure \cite{sun2024adaptive, vrabie2009neural, chaubey2026adaptive}. Related approaches further combine offline-trained deep RL with nonlinear MPC to calibrate model parameters, ensuring synchronization between the digital and physical systems \cite{berg2025digital}. In this work, the focus is placed exclusively on model-level adaptation, as it directly determines the fidelity between the virtual twin and the physical system.

Model-level adaptation typically estimates parameters under uncertainty via optimization \cite{zhang2025self}, recursive least squares \cite{heirung2017dual}, Kalman filtering \cite{bouffard2012learning, dutta2021new, chen2021unknown}, or divergence-based correction \cite{huang2023novel}, with AM applications calibrating temperature-dependent properties \cite{van2025real} or Koopman models \cite{chen2025adaptive}. Naive recursive estimation, however, risks instability and overfitting to streaming data, yielding only locally accurate models rather than global ones.

To mitigate overfitting and forgetting, several strategies constrain updates to specific components: residual learning of physics--data discrepancies \cite{halaly2024continuous, jiahao2023online, chee2022knode, hewing2019cautious}, layer-wise online updating \cite{gasparino2023unmatched}, and adaptive bias correction \cite{jiahao2023online}. These methods all fine-tune the model by preserving the pretrained structure and minimizing the number of tunable parameters, highlighting the importance of data- and parameter-efficient fine-tuning. Motivated by these works, the present work advocates integrating PEFT-based methods into Digital Twin adaptation, particularly when NNs are used as surrogates, enabling controlled, efficient, and robust model updates from limited streaming data while balancing plasticity and stability.

\subsection{Is the Update Successful?}

In safety-critical and high-consequence systems, model updating must be accompanied by systematic validation to ensure that adaptation truly improves, rather than degrades, predictive and control performance. In the context of trustworthy scientific machine learning, verification and validation (V\&V) are regarded not as one-time procedures but as iterative processes of credibility assessment spanning problem formulation, model development, deployment, and domain shift \cite{jakeman2025verification}. In particular, distributional shift between the training and operational domains necessitates continuous validation to maintain model reliability.

Yet most works evaluate updates offline or retrospectively, e.g.\ via held-out transfer-learning performance \cite{bao2020online} or post-hoc classification tests \cite{trovato2026continual}, which benchmark accuracy but do not safeguard real-time updating in closed-loop systems or guarantee stability and constraint satisfaction during deployment. A notable exception is the Lyapunov-based MPC of \cite{wu2019real}, which gates updates on stability conditions and accumulated-error triggers. Overall, real-time validation mechanisms that assess model performance after updating, at test time, remain scarce. Developing principled criteria to certify the success of model updates in Digital Twin systems is therefore an important research direction.

\textcolor{black}{Among the existing literature, the studies most closely related to our scope are our previous work~\cite{karkaria2025ai} and Wu et al.~\cite{wu2019real}. The former integrates MCD drift detection with uncertainty-aware LoRA for model adaptation in service-cycle tire maintenance using periodic inspection feedback. However, its evaluation focuses on predictive improvements before and after fine-tuning, rather than demonstrating repeated adaptation, validation, and model deployment within a continuously operating feedback-control loop. The latter combines Lyapunov-based and accumulated-error triggers with online RNN updating to maintain closed-loop stability and predictive accuracy, providing a systematic framework for model adaptation during operation. However, frequent step-wise model updating may introduce computational overhead and risks of overfitting, while its Lyapunov-based criterion requires a suitable Lyapunov function and stabilizing controller for the underlying dynamical system. These requirements make the approach challenging to generalize to other supervised-learned models, such as multi-step-ahead predictors that are studied in this work. Building on these directions, there remains a need for a generalizable framework that integrates drift detection, parameter-efficient model adaptation, and systematic post-update validation within real-time decision-making, enabling Digital Twins to autonomously maintain predictive fidelity throughout their operational life cycle.}

\subsection{Research Objective}

A central research gap lies in the absence of a trustworthy framework that enables Digital Twins to support real-time decision-making while remaining reliable and adaptive throughout their life cycle. Although machine learning-enabled MPC has substantially improved predictive accuracy for control, most existing efforts concentrate on local model performance and overlook the need for drift detection and systematic validation of continuously updated models. This deviates from the purpose of a Digital Twin, whose target is to maintain the faithfulness of the virtual model to the physical system, not solely to solve a control problem well. From a trustworthy-AI perspective, updating the deployed models without validation raises fundamental concerns regarding credibility and safety. While the control community has traditionally emphasized performance and stability, the statistical-learning literature has long recognized drift management and validation as essential components of adaptive systems. This reveals a critical gap at the intersection of drift detection, continual learning with validation, and real-time decision-making of Digital Twins over their life cycle.

To the best of our knowledge, no existing framework simultaneously addresses the following three challenges:
\begin{enumerate}
    \item how to integrate drift detection into real-time decision-making frameworks with online reset and adaptation mechanisms for multivariate NN surrogate models, providing insight into when distributional shifts occur;
    \item how to conduct streaming-data continual learning in a parameter- and data-efficient manner; and
    \item how to embed online validation within continual learning to safeguard adaptation and improve model trustworthiness.
\end{enumerate}

\begin{figure}[h!]
    \centering
    \includegraphics[width=1\linewidth]{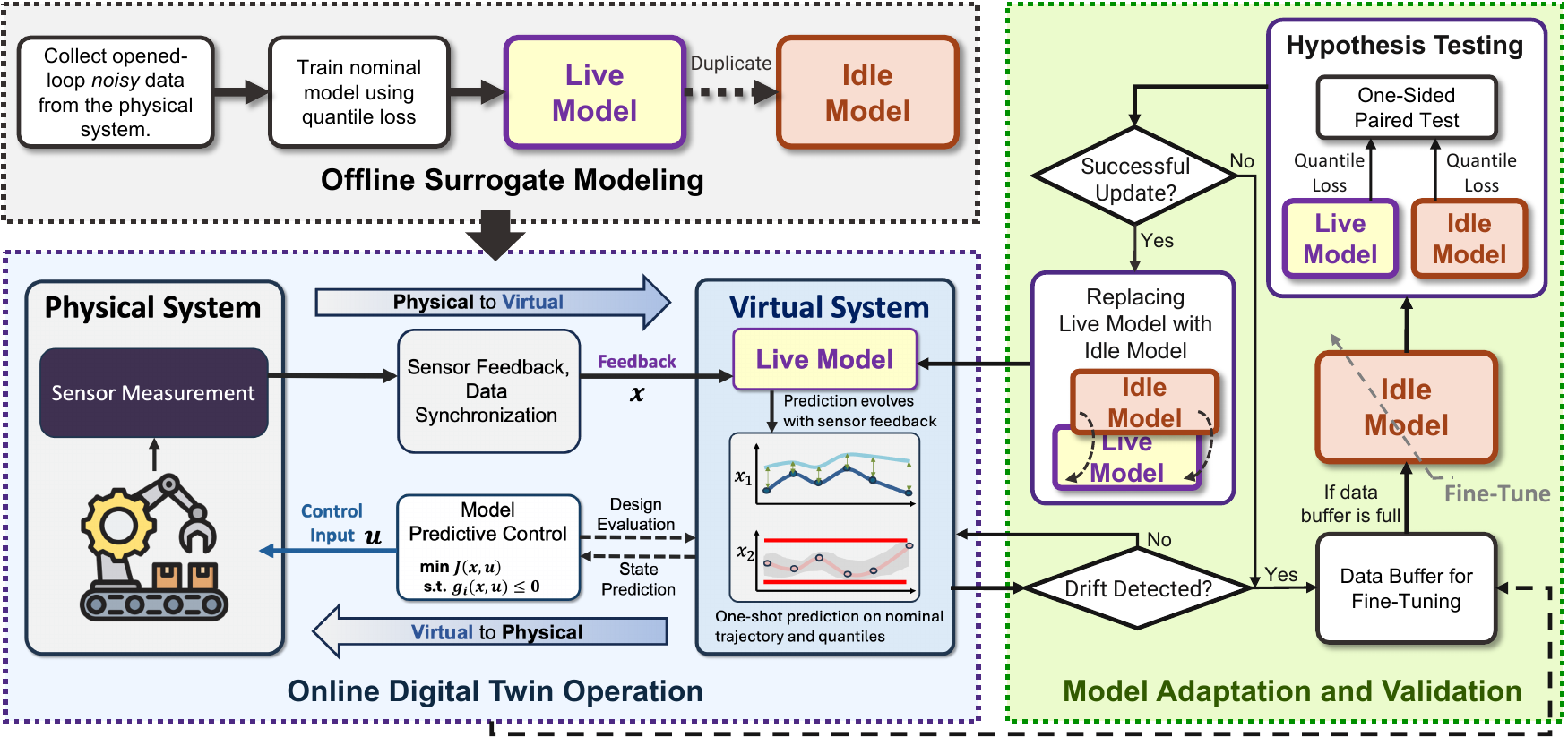}
    \caption{The overall framework for the Adaptive Digital Twin. The gray box and the blue box on the left-hand side were developed in our previous work \cite{chen2026uncertainty}, and the green box on the right-hand side highlights the novelty of this work. The illustration also shows that the proposed method can serve as a non-intrusive add-on to existing Digital Twin frameworks.}
    \label{fig:overall_framework}
\end{figure}

To address these challenges, we propose an adaptive Digital Twin framework for handling concept drift in real-time decision-making, as illustrated in Figure~\ref{fig:overall_framework}. Building upon our prior work on uncertainty-aware Digital Twins~\cite{chen2026uncertainty}, which represents a closed-loop Digital Twin system (left-hand side of Figure~\ref{fig:overall_framework}), the present work extends this architecture with a generalized adaptation layer (right-hand side of Figure~\ref{fig:overall_framework}) that can be implemented across NN-based Digital Twin systems. The key distinction from existing work is that our framework embeds an additional level of uncertainty awareness and validation to ensure that the model is fine-tuned successfully and automatically.

The primary contribution of this work lies not in the individual components but in their integration into a unified framework that systematically couples drift detection, continual learning, and statistical validation within real-time decision-making, forming a comprehensive pipeline for the continual adaptation of NN-based Digital Twins that maintains both predictive accuracy and trustworthiness throughout the operational life cycle, even in the presence of previously unseen distributional shifts. The selected methods and implementations for each module are further \textcolor{black}{modified to maximize generalizability to parametric supervised-learned surrogate models, enhancing applicability across NN-based Digital Twin systems beyond the specific architectures and case studies considered here.} Specifically, the elements that contribute to this framework are as follows:

\begin{itemize}
    \item \textbf{\textcolor{black}{Generalizable multivariate
    } drift-detection mechanism} --- determining \emph{when to update the model}: \textcolor{black}{A gradient--based multivariate drift-detection method, modified from the score-based method \cite{zhang2023concept} is developed to identify concept drift in NN surrogate models during Digital Twin operation. To enable life-long and real-time implementations, adjustments to the original methods are introduced, including} diagonal covariance approximation and threshold estimation via bootstrapping, making the approach computationally tractable for online deployment. The method is model-agnostic and applicable to general supervised-learned, differentiable models, providing detection capability for concept drift.

    \item \textbf{Parameter-efficient fine-tuning} --- determining \emph{how to update the model}: To the best of our knowledge, this is the first work to introduce LoRA as a parameter- and data-efficient fine-tuning strategy for online surrogate-model adaptation in Digital Twin systems and adaptive MPC governed by highly nonlinear dynamics, enabling rapid updates from limited streaming data while mitigating catastrophic forgetting. \textcolor{black}{Through benchmarking against commonly-used fine-tuning strategies, we show that LoRA is competitive with, and in the settings considered slightly superior to, head and full fine-tuning, and we adopt it as the default fine-tuning method within this framework.}

    \item \textbf{Online validation} --- determining \emph{whether the update is successful}:  An online validation procedure based on the \textcolor{black}{one-sided paired moving-block bootstrapping (MBB) test} is introduced to assess whether the adapted model achieves a significant reduction in predictive loss relative to the current live model. This test ensures that model replacement is justified by measurable predictive improvement, providing a safeguard against deploying an inadequately updated surrogate.

\end{itemize}

The rest of the paper is organized as follows. Section~\ref{sec:tech_background} introduces the technical foundations of the underlying closed-loop Digital Twin \textcolor{black}{which are mainly used to develop the testbeds introduced in our previous work \cite{chen2026uncertainty}}, including MPC and its robust extension, surrogate modeling via TiDE, and uncertainty-aware learning through quantile regression. Section~\ref{sec:method} presents the Adaptive Digital Twin framework, introducing \textcolor{black}{gradient}--based drift detection and LoRA together with their online implementations, and detailing their integration with robust MPC for continual decision-making under evolving operating conditions. Section~\ref{sec:score_based_ex} evaluates \textcolor{black}{the effectiveness of online model adaptation through a series of benchmark studies.} Section~\ref{sec:AM} demonstrates \textcolor{black}{the implementation of the framework on a directed energy deposition (DED) Digital Twin operating under concept drift via a high-fidelity physics-based simulator.} Finally, Section~\ref{sec:closure} summarizes the main findings and current limitations, and outlines directions for future research.

\section{Technical Background}
\label{sec:tech_background}
This section reviews the three building blocks of the closed-loop, uncertainty-aware Digital Twin upon which the proposed framework is constructed: decision-making through MPC and its robust extension, surrogate modeling of dynamical systems via the Time-series Dense Encoder (TiDE), and quantile regression for uncertainty-aware learning. \textcolor{black}{Note that these are the fundamentals for the development of the closed-loop, uncertainty-aware Digital Twins introduced in our previous work \cite{chen2026uncertainty}, whereas technical details related to the proposed method in this work are elaborated in Section~\ref{sec:method}}. 

\subsection{Model Predictive Control and Robust Model Predictive Control}

{\color{black}

Model predictive control (MPC) is a constrained optimal control method that repeatedly solves a finite-horizon optimization problem using a predictive model of the system \cite{rawlings2017model}. At each sampling instant, MPC optimizes a sequence of future control inputs to minimize a prescribed objective, typically balancing reference tracking and control effort while satisfying operational constraints. Only the first control input is applied to the physical system, and the optimization is repeated using updated measurements, as illustrated in Figure~\ref{fig:MPC_illustration}(a) and (b).

\begin{figure}[h!]
\centering
\includegraphics[width=1\linewidth]{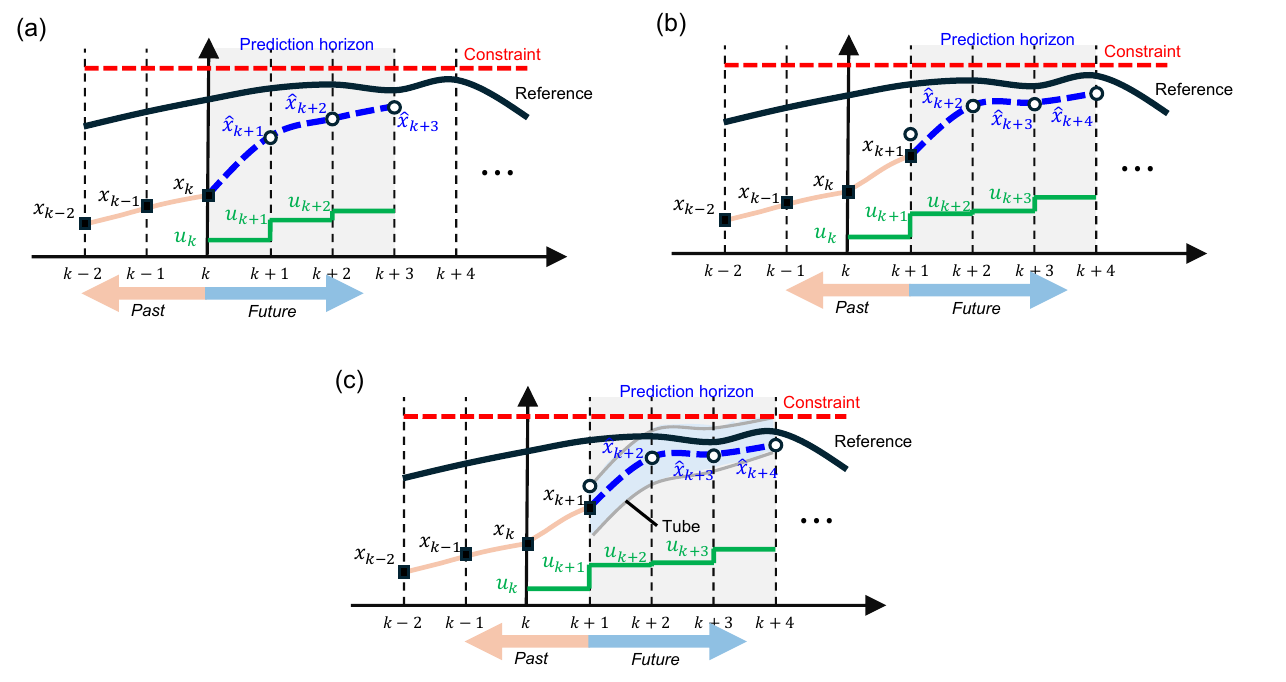}
\caption{Illustration of MPC and robust MPC. (a) MPC optimization at time $k$. (b) Receding-horizon optimization at time $k+1$. (c) Tube-based robust MPC accounting for uncertainty in state predictions. The green lines represent the planned control input sequences, while the blue dashed lines represent the nominal state predictions. The blue shaded region in (c) illustrates the uncertainty tube surrounding the nominal trajectory.}
\label{fig:MPC_illustration}
\end{figure}

Conventional MPC relies on nominal predictions and may violate constraints under disturbances and model uncertainty. Robust MPC explicitly accounts for this uncertainty. Consider a discrete-time nonlinear system subject to disturbances,
\begin{equation}
  \mathbf{x}_{k+1} = f_{w}(\mathbf{x}_k, \mathbf{u}_k, \mathbf{w}_k),
  \label{eq:MPC} 
\end{equation}
where $\mathbf{x}_k \in \mathbb{R}^{n}$ and $\mathbf{u}_k \in \mathbb{R}^{m}$ are the state and applied input, $\mathbf{w}_k \in \mathbb{R}^{p}$ is an exogenous disturbance, and $\mathcal{X}$ and $\mathcal{U}$ denote the admissible state and input sets. Robust MPC enforces the constraints for all admissible disturbances when these are bounded, and through chance constraints when they are unbounded and stochastic. A common realization is tube-based robust MPC (Figure~\ref{fig:MPC_illustration}(c))~\cite{tsai2023robust, tsai2025control}, which optimizes a nominal trajectory, corrects deviations of the actual state through an ancillary feedback law, and tightens the constraints by the set of admissible deviations (the tube) using the Pontryagin difference $\ominus$, so that satisfying the tightened constraints implies satisfaction of the original ones whenever the tube bounds the actual deviations. This work adopts the data-driven tube-based robust MPC of our previous work~\cite{chen2026uncertainty}, which characterizes the tube through multi-step quantile predictions, as formulated in Section~\ref{sec:multistep_robust_mpc}.

}

\subsection{Time-Series Dense Encoder}
\label{sec:tide}
The surrogate must support fast inference for repeated MPC evaluations and accommodate exogenous control inputs \cite{chen2026uncertainty}. We adopt TiDE \cite{das2023long}, the \textbf{Ti}me-series \textbf{D}ense \textbf{E}ncoder, an encoder--decoder of stacked linear residual blocks that attains competitive accuracy while being 5--10$\times$ faster at inference than Transformer counterparts. Moreover, it supports simultaneous multi-step prediction, avoiding error propagation while learning temporal dependencies, which suits partially observable systems and parallelizes the prediction of future state trajectories.
\begin{figure}[h!]
    \centering
    \includegraphics[width=0.5\linewidth]{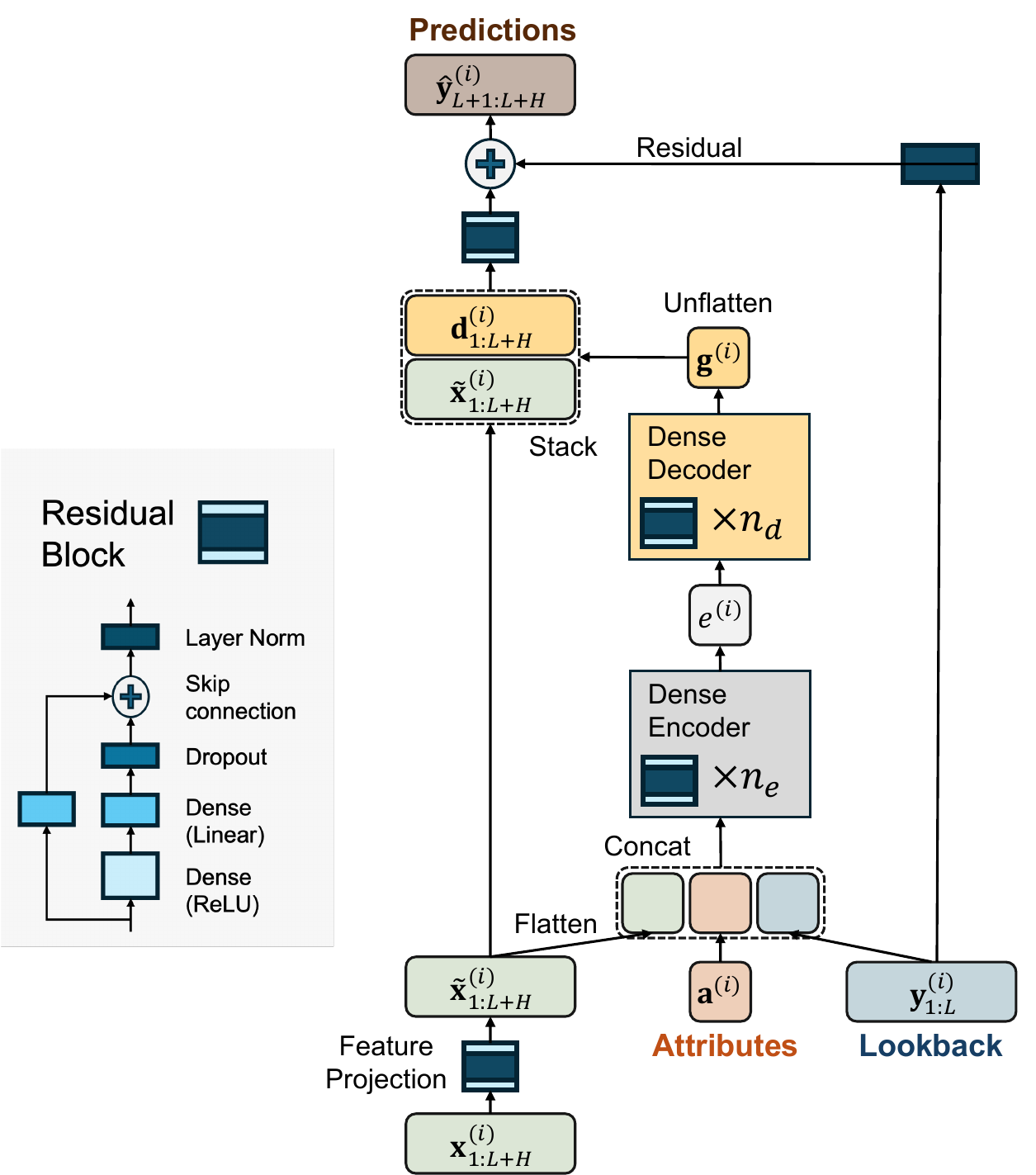}
    \caption{The network structure of Time Series Dense Encoder (TiDE), modified from \cite{das2023long}}
    \label{fig:TiDE}
\end{figure}
TiDE encodes the full look-back window and covariates into a compact latent representation and decodes the entire horizon in a single forward pass (Figure~\ref{fig:TiDE}), avoiding the sequential unrolling of autoregressive models and the quadratic cost of self-attention while filtering noise through its low-dimensional bottleneck. TiDE separates \emph{targets} (the predicted states) from \emph{covariates} (auxiliary time-varying inputs that condition but are not predicted), the latter split into past covariates (observed over the look-back window) and future covariates (known over the horizon). This maps naturally onto a control-exogenous dynamical system whose state evolution depends on its history and on applied inputs and prescribed conditions. Specifically, TiDE takes past states $\mathbf{x}_{k-w+1:k}$ as the past target, past inputs $\mathbf{u}_{k-w:k-1}$ and conditions $\{\mathbf{u}_{k-w:k-1},\mathbf{d}_{k-w:k-1}\}$ as past covariates, and planned future inputs and conditions $\{\mathbf{u}_{k:k+N-1},\mathbf{d}_{k:k+N-1}\}$ as future covariates, producing future states predictions $\hat{\mathbf{x}}_{k+1:k+N}$, where $\mathbf{d}$ are prescribed process parameters and $\mathbf{u}$ the inputs optimized by the MPC. The surrogate is
\begin{align}
    \hat{\mathbf{x}}_{k+1:k+N} = \textbf{TiDE}(\mathbf{x}_{k-w+1:k},& \;\mathbf{d}_{k-w:k-1},\; \mathbf{u}_{k-w:k-1}, \mathbf{d}_{k:k+N-1},\; \mathbf{u}_{k:k+N-1} \mid \boldsymbol{\theta}), \label{eq:TiDE_general}
\end{align}
where $\boldsymbol{\theta}$ collects all trainable parameters, allowing the MPC optimizer to query TiDE with candidate input sequences for simultaneous $N$-step predictions without sequential unrolling.

\subsection{Quantile Regression}
\label{sec:quantile_reg}
In engineering applications, a predictive model that returns only a single point estimate of future system states is often insufficient for robust decision-making, as it gives no indication of how reliable that estimate is under disturbances. Quantile regression \cite{koenker2001quantile, yu2003quantile, fan2020deep} addresses this limitation by predicting conditional quantiles rather than the mean, yielding distribution-free prediction bounds that represent aleatoric uncertainty and accommodate heteroscedasticity without assuming an error distribution. The loss function for a given quantile level $q \in (0, 1)$ and response channel $j\in{1,...,N_d}$ is defined as
\begin{equation}
    L_{q,j}(x_{j,t}, \hat{x}_{j,t}) =
\begin{cases}
q \cdot (x_{j,t} - \hat{x}_{j,t}), & \text{if } x_{j,t} \geq \hat{x}_{j,t}, \\
(1 - q) \cdot (\hat{x}_{j,t} - x_{j,t}), & \text{if } x_{j,t} < \hat{x}_{j,t},
\end{cases} \label{eq:standard_quantile}
\end{equation}
where $x_t$ and $\hat{x}_t$ are the ground-truth and predicted values at time $t$, respectively. The asymmetric weighting on the two residual terms encourages the model's output to converge to the $q$-th percentile of the conditional distribution: setting $q = 0.5$ recovers the median, while $q = 0.1$ and $q = 0.9$ yield the lower and upper bounds of an 80\% prediction interval.
By training TiDE simultaneously across multiple quantile levels using the aggregate of the $L_{q,j}$ losses, the framework produces interval predictions of future system states that can be directly propagated through the MPC formulation to enable uncertainty-aware control \cite{fan2020deep,chen2026uncertainty, chen2025real}. The aggregated quantile loss function across all channels and for multi-step-ahead prediction can therefore be denoted as:
\begin{equation}
L_Q(\mathbf{x}_{k+1:k+N},\mathbf{\hat{x}}_{k+1:k+N})= \sum_{q=1}^{N_q}\sum_{j=1}^{N_d}\sum_{i=k+1}^{k+N}L_{q,j}(x_{i,j},\hat{x}_{i,j})
\end{equation}

Together, these three components form the uncertainty-aware, closed-loop Digital Twin developed in our prior work~\cite{chen2026uncertainty}: TiDE provides a fast multi-step surrogate, quantile regression supplies distribution-free uncertainty bounds, and robust MPC translates these bounds into safe control actions. However, this framework assumes that the surrogate remains faithful to the physical system. When operating conditions drift, the learned quantiles, and thus the safety margins used by robust MPC, may no longer be valid. The next section introduces the adaptation layer, which detects drift, efficiently updates the surrogate using streaming data, and statistically validates each update before reintegrating it into the decision-making loop.

\section{The Proposed Method}
\label{sec:method}

This section presents the proposed Adaptive Digital Twin framework, summarized in Figure~\ref{fig:overall_framework}. Building on the closed-loop components reviewed in Section~\ref{sec:tech_background}, the presentation is organized around the three questions posed in the Introduction: Section~\ref{sec:when_to_update} introduces Fisher score--based drift detection and its online implementation to answer \emph{when} to update the model, Section~\ref{sec:how_to_update} introduces LoRA and the batch-wise fine-tuning procedure built around it to answer \emph{how} to update the model, and Section~\ref{sec:validation} presents the statistical validation that determines \emph{whether} an update is successful and may replace the live model. We make two assumptions on the drift that the physical system encounters:
\begin{assumption}
    The physical system encounters only abrupt drift and incremental drift, which are not recurring and infrequent.
\end{assumption}
\begin{assumption}
    Following the characterization of drift as either abrupt or gradual, it is assumed that, after model adaptation, the system does not experience an immediate subsequent drastic drift. In other words, drift evolves on a timescale that allows the updated model to remain valid for a non-negligible period.
\end{assumption}
These assumptions reflect the physics and operating limits of engineered systems. Behavior changes arise from discrete events (faults, regime switches) or continuous processes (wear, aging, thermal effects), giving abrupt and incremental drift respectively, with complex patterns interpretable as combinations of the two. Abrupt events are infrequent owing to limited failure rates, while incremental drift cannot induce successive large deviations, and monitoring/control systems should stabilize operation after each correction. Statistically, this enforces a separation of timescales between drift and adaptation, ensuring the updated model stays valid long enough for reliable estimation, validation, and decision-making. Even if rapid successive drift occurs, the validation gate simply withholds deployment, and the framework keeps re-adapting on fresh data, so a violated assumption costs responsiveness but does not push an unvalidated model into the safety-critical loop.
\begin{figure}[h!]
    \centering
    \includegraphics[width=1\linewidth]{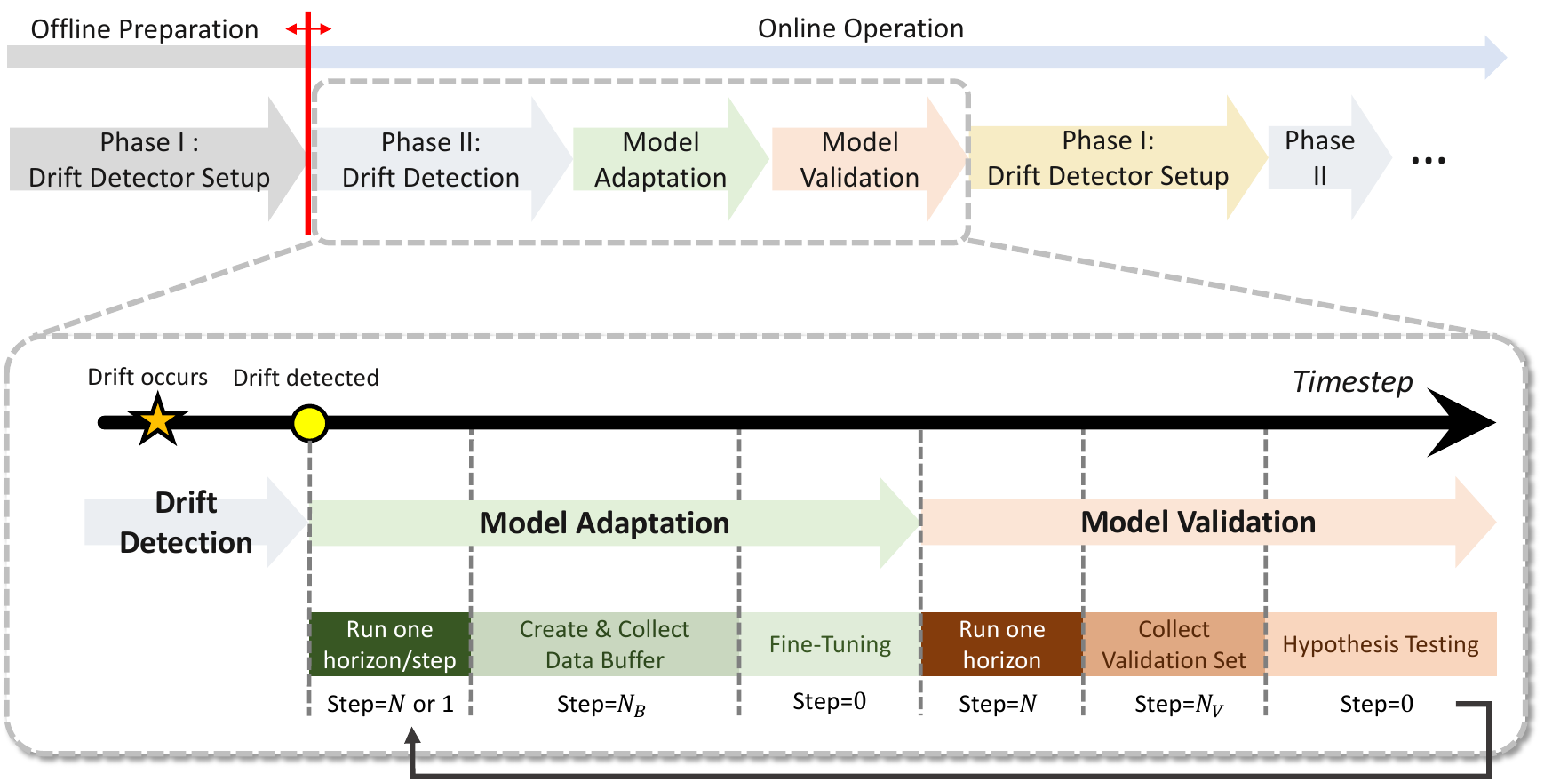}
    \caption{Timeline of the proposed framework.}
    \label{fig:timeline}
\end{figure}

\subsection{Offline Model Training}
\label{sec:offline_training}

\textcolor{black}{Data are collected to train an offline predictive model using TiDE (Section~\ref{sec:tide}) with quantile regression (Section~\ref{sec:quantile_reg}), simultaneously predicting the median and the upper (0.95) and lower (0.05) quantiles for uncertainty-aware prediction. The parameters of the pretrained model are then frozen, with LoRA layers (Section~\ref{sec:lora}) appended to all layers for efficient adaptation. If the network lacks a dedicated final linear layer for gradient monitoring (Section~\ref{sec:NN_application}), as in TiDE due to its residual connections, an identity-initialized linear layer $\mathbf{W}_{\text{lin}}$ is added after pretraining. Specifically, the layer is applied to the raw network output, represented as a two-dimensional tensor, before it is reshaped into the final prediction dimensions. Since $\mathbf{W}_{\text{lin}}$ is initialized as an identity matrix with zero bias, this transformation preserves the model output. The additional layer is then selected as the only gradient-monitored layer, providing a dedicated set of parameters for gradient-based drift detection without altering the original predictive model. Although this transformation does not alter the model outputs, the gradients of the prediction loss with respect to $\mathbf{W}_{\text{lin}}$ still reflect how incoming observations influence the model's output-level sensitivities, which is sufficient for high-dimensional NNs \cite{zhang2023concept}, as discussed in Section~\ref{sec:NN_application}. }

\subsection{Continual Decision-Making with Multi-Step Robust Model Predictive Control}
\label{sec:multistep_robust_mpc}
For uncertainty-aware decision-making, we adopt the simultaneous multi-step robust MPC framework of \cite{chen2026uncertainty}. This approach integrates multi-step-ahead predictions with deep quantile learning to reformulate the stochastic MPC problem into a tractable deterministic optimization while retaining probabilistic guarantees. Through its receding-horizon formulation, the MPC continually updates and resolves the control problem using the latest state measurements from the physical system, enabling the Digital Twin to provide adaptive and optimal decisions in real time.

Consider a nonlinear dynamical system subject to stochastic disturbances as described in Equation~\ref{eq:stochastic_system}. The robust MPC optimizes the nominal control inputs $v$ over a finite horizon while satisfying probabilistic constraints on the system states:{\color{black}
\begin{equation}
\min_{v} \ \mathbb{E}[J(v, x, r, w)], \quad
\text{s.t. } \Pr(x_{k+i} \in \mathcal{X}) \ge \alpha_{g}, \ \forall i, \quad u_{k+i} \in \mathcal{U}, 
\end{equation}}
\textcolor{black}{where $\alpha_g$ is the constraint satisfaction rate for probabilistic constraints $g$.} However, directly solving this problem is computationally intractable due to uncertainty propagation and the evaluation of probabilistic constraints.

In this work, we leverage a multi-step predictor via TiDE, $\hat{\mathbf{x}}_{k+1:k+N}=[\bar{\mathbf{x}}_{k+1:k+N},\tilde{\mathbf{x}}_{k+1:k+N},\underline{\mathbf{x}}_{k+1:k+N}]$, to directly predict the future-state distribution through learned quantiles, where $\bar{x},\underline{x}$ are the upper/lower quantiles at confidence level $\alpha_g$. Each chance constraint is then enforced deterministically on the predicted quantile, e.g.\ $\Pr(x_{j,k+i}\le x_j^{\mathrm{ub}})\ge\alpha_g \Leftrightarrow \bar{x}_{j,k+i}\le x_j^{\mathrm{ub}}$ (and analogously for the lower bound), reducing the stochastic MPC to the deterministic problem \textcolor{black}{$\min_{v} J(v,\tilde{x},r)$ subject to $\bar{x}_{k+i}(v)\le x^{\mathrm{ub}}$ and $\underline{x}_{k+i}(v)\ge x^{\mathrm{lb}}$ for all $i$, with the $v$-dependence entering through the predictive model. The aleatoric uncertainty is thus handled explicitly through the learned bounds, avoiding sample-based approaches (Monte Carlo dropout, Bayesian NNs) and iterative uncertainty propagation in which error may accumulate \cite{kohler2022state}. Moreover, an ancillary feedback policy $u_k=v_k+Ke_k$, $e_k=x_k-\tilde{x}_k$, is added and the input set tightened to $v_{k+i}\in\mathcal{U}\ominus K\mathcal{Z}_{k+i}$, where $v_k$ is the nominal control input optimized by MPC, $u_k$ is the actual input applied to the system after feedback correction,} and $\mathcal{Z}_{k+i}$ is the prediction-error bound, ensuring feasibility under disturbances. Through this reformulation, the uncertainty-aware MPC problem is converted into a deterministic optimization problem that can be solved efficiently in real time while maintaining probabilistic guarantees on constraint satisfaction via the learned quantile bounds. The reader may refer to \cite{chen2026uncertainty} for more details on the multi-step robust MPC.

{\color{black}
\subsection{\emph{When to Update the Digital Twin?} -- Gradient-Based Drift Detection}
\label{sec:when_to_update}

The first module of the framework determines \emph{when} the surrogate model may need to be updated by monitoring whether newly observed data remain consistent with the distribution represented by the fitted model. Building on the score-based drift detection framework of \cite{zhang2023concept}, we extend its empirical monitoring principle to gradients of general supervised learning loss functions, providing a multivariate mechanism for identifying distributional changes. The original score-based method monitors the sensitivity of the log-likelihood with respect to model parameters, whereas the generalized gradient-based approach measures how individual observations influence the prediction loss through the model parameters. Unlike error-based approaches that primarily monitor prediction errors, gradient-based monitoring captures multivariate information about model sensitivities, potentially identifying distributional changes that do not immediately result in increased prediction errors. The gradient vectors are aggregated into a single detection statistic using the Mahalanobis distance, which accounts for correlations among gradient components, while individual components provide additional insight into how changes are reflected in the model parameters \cite{zhang2023concept}, which has been extended to monitor the sensitivity of each parameter, providing potential applications for diagnosing parametric models. For NN-based models, gradients can be efficiently computed through backpropagation, supporting their application in online monitoring. In the following, we first review the statistical foundation of score-based detection and its generalization to supervised learning losses (Sections~\ref{sec:score_vector}--\ref{sec:NN_application}), followed by its online implementation within the proposed framework (Section~\ref{sec:online_implementation}).

\subsubsection{Gradient-Based Monitoring for Supervised Learning Models}
\label{sec:score_vector}

Motivated by the flexibility and potential interpretability of the score-based concept drift monitoring approach \cite{zhang2023concept}, we aim to generalize the same concept to monitoring the gradients of the parameters with respect to the supervised learning loss. Let $\{(\mathbf{x}_i, y_i)\}_{i=1}^{n}$ denote a training dataset of $n$ observations collected under an in-control regime, where $\mathbf{x}_i \in \mathbb{R}^p$ is the model input and $y_i \in \mathbb{R}$ is the response for the $i$-th observation. Consider a parametric conditional distribution $P(Y \mid \mathbf{X};\, \boldsymbol{\theta})$, indexed by a parameter vector $\boldsymbol{\theta} \in \mathbb{R}^q$. The supervised learning model, parameterized by $\hat{\boldsymbol{\theta}}^{(0)}$, is fitted by maximizing the log-likelihood over the training data, yielding the maximum likelihood estimate (MLE):
\begin{equation}
    \hat{\boldsymbol{\theta}}^{(0)}
    =
    \arg\max_{\boldsymbol{\theta}}
    \frac{1}{n}\sum_{i=1}^{n}
    \log P\!\left(y_i \mid \mathbf{x}_i;\, \boldsymbol{\theta}\right).
    \label{eq:mle}
\end{equation}

The \emph{Fisher score vector} (hereafter, the \emph{score vector}) for a single observation $(\mathbf{x}_t, y_t)$ evaluated at the fitted parameters $\hat{\boldsymbol{\theta}}^{(0)}$ is defined as the gradient of the log-likelihood with respect to $\boldsymbol{\theta}$:
\begin{equation}
    \mathbf{s}_t
    =
    \nabla_{\!\boldsymbol{\theta}}
    \log P\!\left(y_t \mid \mathbf{x}_t;\, \boldsymbol{\theta}\right)
    \Big|_{\boldsymbol{\theta} = \hat{\boldsymbol{\theta}}^{(0)}}.
    \label{eq:score}
\end{equation}

Under the assumption that the model is correctly specified and the true parameter vector is $\boldsymbol{\theta}^{(0)}$, the score vector satisfies the zero-mean property:
\begin{equation}
    \mathbb{E}_{\boldsymbol{\theta}^{(0)}}\!\left[
        \mathbf{s}\!\left(\boldsymbol{\theta}^{(0)};\,(\mathbf{X},Y)\right)
        \mid \mathbf{X}
    \right] = \mathbf{0}.
    \label{eq:zero_mean}
\end{equation}
Moreover, at an interior stationary point of the unregularized maximum likelihood objective, the average score vector over the training data is zero:
\begin{equation}
    \frac{1}{n}\sum_{i=1}^{n}
    \mathbf{s}\!\left(\hat{\boldsymbol{\theta}}^{(0)};\,(\mathbf{x}_i, y_i)\right)
    = \mathbf{0}.
    \label{eq:empirical_zero}
\end{equation}
These properties motivate score-based drift detection, which monitors changes in the score vectors computed from incoming observations using a fixed predictive model \cite{zhang2023concept}.

In this work, we argue that the same empirical monitoring principle can be generalized to supervised learning models trained using loss functions other than the negative log-likelihood. Let $\ell(\mathbf{x},y;\boldsymbol{\theta})$ denote a general supervised learning loss function. The corresponding gradient vector for an observation $(\mathbf{x}_t,y_t)$ is defined as
\begin{equation}
    \mathbf{g}_t
    =
    \nabla_{\!\boldsymbol{\theta}}
    \ell(\mathbf{x}_t,y_t;\boldsymbol{\theta})
    \Big|_{\boldsymbol{\theta}=\hat{\boldsymbol{\theta}}^{(0)}}.
    \label{eq:gradient_vector}
\end{equation}
When $\ell$ is the negative log-likelihood, $\mathbf{g}_t=-\mathbf{s}_t$, and the gradient-based formulation reduces to score-based monitoring up to a sign change. For general supervised learning losses, however, the theoretical zero-mean property in Equation~\eqref{eq:zero_mean} is not necessarily satisfied. In particular, limited learning capability in training the parametric model, stochastic optimization, and differences between the training and monitoring objectives may result in a nonzero mean gradient even under an in-control regime.

Nevertheless, the empirical principle of gradient-based monitoring remains applicable without requiring the reference gradient mean to be zero. Specifically, the empirical reference mean is estimated from the in-control observations as
\begin{equation}
    \bar{\mathbf{g}}^{(0)}
    =
    \frac{1}{n}
    \sum_{i=1}^{n}
    \mathbf{g}\!\left(
    \hat{\boldsymbol{\theta}}^{(0)};
    (\mathbf{x}_i,y_i)
    \right).
    \label{eq:empirical_gradient_mean}
\end{equation}

The same loss-gradient definition and fixed model parameters are used for the in-control reference data and incoming observations. Under a stationary data-generating process, the mean gradient of incoming observations is expected to remain close to $\bar{\mathbf{g}}^{(0)}$, subject to sampling variability. When the underlying data distribution changes, the mean gradient may deviate from this reference value. Therefore, monitoring the difference between the incoming and reference mean (as demonstrated in Equation~\ref{eq:t2} in the next section) provides an empirical basis for detecting distributional changes, without relying on the theoretical zero-mean property of the Fisher score. For notational consistency with the score-based framework of \cite{zhang2023concept}, we hereafter denote the gradient vector $\mathbf{g}_t$ as $\mathbf{s}_t$. Note that $\mathbf{s}_t$ represents the gradient of a general supervised learning loss and does not necessarily satisfy the theoretical properties of the Fisher score.

The gradient-based framework is intended to detect concept drift, characterized by changes in the conditional distribution $P(Y\mid\mathbf{X})$. Under a correctly specified conditional model evaluated at its true parameters, changes in the covariate distribution $P(\mathbf{X})$ alone do not alter the expected gradient, since its conditional expectation remains at the same mean gradient. However, supervised learning models trained or fine-tuned under a limited operating regime (which is assumed to be stationary) may exhibit reduced predictive accuracy when extrapolating beyond the observed input distribution, especially under non-stationary operations. Consequently, covariate shifts may expose model approximation errors that introduce changes in the monitored gradients, even when the underlying conditional distribution remains unchanged. Covariate shifts may also alter gradient variability and affect the monitoring statistic even when the gradient mean remains unchanged. Therefore, while the proposed method primarily targets concept drift, its response to covariate shifts may also reflect the limited generalization capability of the predictive model rather than changes in the underlying system dynamics. This also highlights the need for continual monitoring and fine-tuning of the surrogate model throughout the system life cycle under non-stationary conditions.
}

\subsubsection{Drift Detection Framework}
\paragraph{Phase I: Retrospective Analysis and Control Limit Estimation.}
Collect a batch of chronologically ordered observations $\mathcal{D} = \{(\mathbf{x}_i, y_i)\}_{i=1}^{m}$ from the historical record and partition it into two subsets: $\mathcal{D}_1 = \{(\mathbf{x}_i, y_i)\}_{i=1}^{m_1}$ for estimating the distribution of the score vector, and $\mathcal{D}_2 = \{(\mathbf{x}_i, y_i)\}_{i=m_1+1}^{m}$ for estimating the control limit. With a pre-trained parametric supervised model fit offline and its \textcolor{black}{learned parameters} $\hat{\boldsymbol{\theta}}^{(0)} \in \mathbb{R}^q$ (Equation~\eqref{eq:mle}), the \textcolor{black}{gradient} vector $\mathbf{s}_i$ is then computed for each $(\mathbf{x}_i, y_i) \in \mathcal{D}_1$ at the fixed $\hat{\boldsymbol{\theta}}^{(0)}$ (Equation~\eqref{eq:gradient_vector}), and further used to compute the mean of the \textcolor{black}{gradient} vector $\bar{\mathbf{s}}$ and the covariance of the \textcolor{black}{gradient} vector $\widehat{\boldsymbol{\Sigma}}$. 

Note that fitting $\bar{\mathbf{s}}$ and $\widehat{\boldsymbol{\Sigma}}$ on held-out data different from the training set avoids the bias of reusing $\mathcal{D}_1$, so the \textcolor{black}{gradient} reflect the model's sensitivity to new data rather than to the training set. Therefore, with batch $\mathcal{D}_2$ that is independent from $\mathcal{D}_1$, the \textcolor{black}{gradient} sequence using $\mathcal{D}_2$ as the streaming data with $\hat{\boldsymbol{\theta}}^{(0)}$ is smoothed using the Multivariate EWMA (MEWMA) recursion \cite{zou2011multivariate},
\begin{equation}
    \mathbf{z}_t
    \;=\;
    \lambda\, \mathbf{s}_t \;+\; (1 - \lambda)\,\mathbf{z}_{t-1},
    \qquad t = 1, 2, \ldots,\, m-m_1,
    \label{eq:mewma}
\end{equation}
where $\lambda \in (0,1]$ controls the effective memory ($\approx 1/\lambda$ observations) and $\mathbf{z}_0$ is the $\bar{\mathbf{{s}}}$. To reduce the $q$-dimensional $\mathbf{z}_t$ to a scalar monitoring statistic, the Hotelling $T^2$ statistic, i.e.\ the Mahalanobis distance of $\mathbf{z}_t$ from the reference mean $\bar{\mathbf{s}}$, is used,
\begin{equation}
    T_t^2
    \;=\;
    \left(\mathbf{z}_t - \bar{\mathbf{s}}\right)^{\!\top}
    \widehat{\boldsymbol{\Sigma}}^{-1}
    \left(\mathbf{z}_t - \bar{\mathbf{s}}\right).
    \label{eq:t2}
\end{equation}
When $\widehat{\boldsymbol{\Sigma}}$ is singular or ill-conditioned, a nugget regularization $\widehat{\boldsymbol{\Sigma}} \leftarrow \widehat{\boldsymbol{\Sigma}} + \delta \mathbf{I}$ is applied, with $\delta > 0$ the smallest value bringing the condition number below a prescribed threshold.

\textcolor{black}{Ideally}, the upper control limit (UCL) is set as the $(1-\alpha_{\textcolor{black}{\text{UCL}}})$ empirical quantile of $\{T_t^2\}$ over $\mathcal{D}_2$, where $\alpha_{\textcolor{black}{\text{UCL}}}$ is the nominal false-alarm rate. This nonparametric limit requires no distributional assumption and reflects the variability of the reference data directly; estimating it on a batch separate from model fitting and scoring prevents the underestimated variability and inflated false-alarm rates that double use would induce.

\paragraph{Phase II: Prospective Online Monitoring.}
\label{sec:phase-II}
As each time step as new observation arrives, the score $\mathbf{s}_t$ is computed at the estimate $\hat{\boldsymbol{\theta}}^{(0)}$, the MEWMA recursion~\eqref{eq:mewma} and statistic~\eqref{eq:t2} are continued, and at the hypothesis {\color{black}
\begin{align}
  H_0 &: \mathbb{E}\big[\mathbf{s}(\hat{\boldsymbol{\theta}}^{(0)}; (\mathbf{X}, Y))\big] = \bar{\mathbf{s}} \quad \text{(no drift)}, \\
  H_1 &: \mathbb{E}\big[\mathbf{s}(\hat{\boldsymbol{\theta}}^{(0)}; (\mathbf{X}, Y))\big] \neq \bar{\mathbf{s}} \quad \text{(drift present)},
\end{align}}
is evaluated by computing the $T^2$ statistics on new observations. A drift alarm is raised when $T_t^2>\mathrm{UCL}$, upon which monitoring pauses and the fine-tuning procedure of Section~\ref{sec:how_to_update} begins.
\subsubsection{Drift Detection for Neural Networks}
\label{sec:NN_application}

From a statistical perspective, monitoring the score vector of the last layer is a sufficient and efficient approach for detecting distributional changes that affect model outputs, as suggested by \cite{zhang2023concept}. When focusing on the last layer, the score captures deviations in the conditional distribution $p(y\mid z)$, where $z$ denotes the learned latent representation from preceding layers. For many engineering applications, the drift of interest is precisely a change in this conditional relationship, rather than a complete restructuring of the feature extractor. \textcolor{black}{Moreover, since the reference mean and covariance of the gradients are empirically estimated during Phase~I using the augmented model, the additional layer does not need to be included in the original training process. The same fixed layer is subsequently used during Phase~II, ensuring consistency between the reference and incoming gradients. The augmented model becomes the live model, with an identical idle copy.}

Detecting drift only on the last-layer parameters also improves statistical efficiency since estimating the covariance of the full score vector in high dimensions is often ill-conditioned and data-intensive, reducing detection power and increasing memory usage. In contrast, the lower-dimensional last-layer score enables more stable covariance estimation and more reliable test statistics. This yields a favorable trade-off in detection efficiency while ensuring robust inference in streaming, data-limited settings.

\subsubsection{Online Implementation of Gradient-Based Drift Detection for Quantile Prediction}
\label{sec:online_implementation}

\textcolor{black}{Three adjustments for the implementation of the gradient-based monitoring approach in this case have been made in this study, namely the simplification of inverting the covariance matrix, the estimation of the UCL, and the compromise on monitoring the full horizon to reduce delay calls of drift detection. }

\textcolor{black}{First, a primary bottleneck for online applications of monitoring gradients} is inverting the covariance $\hat{\boldsymbol{\Sigma}}^{-1}\in\mathbb{R}^{\tilde{q}^2\times \tilde{q}^2}$ ($\tilde{q}$ the last-layer output dimension), costing $\mathcal{O}(\tilde{q}^6)$ and being statistically unreliable when estimated from limited data. We therefore conducted an empirical study on approximating $\hat{\boldsymbol{\Sigma}}$ by its diagonal ($\mathcal{O}(\tilde{q}^2)$ inversion, more stable estimates) and also considered Kronecker-Factored Approximate Curvature (K-FAC) \cite{martens2015optimizing}. Details of this approximation and an empirical comparison of the covariance strategies are provided in the Supplementary Information (SI). Here, we compared three strategies for the covariance term $\hat{\boldsymbol{\Sigma}}^{-1}$ in the Hotelling $T^2$ statistic on a nonlinear scalar-regression benchmark with heteroscedastic, asymmetric noise: exact full-covariance inversion (\texttt{FULL}), a diagonal approximation (\texttt{DIAG}), and K-FAC (\texttt{KFAC}). Across both a mean-shift drift and a variance-only drift, \texttt{DIAG} consistently achieved the shortest detection delay and the lowest false-alarm rate, trading a small covariance bias for a large reduction in estimation variance, whereas \texttt{KFAC} offered no advantage over \texttt{FULL}. We therefore adopt \texttt{DIAG} for computing $\hat{\boldsymbol{\Sigma}}^{-1}$ throughout this work. The full experimental setup, figures, and analysis are provided in the SI. 

\textcolor{black}{Second,} estimating the UCL online is a second bottleneck for the online implementation of Phase~II. Formally, the empirical-quantile approach requires $\sim 1/\alpha_{\text{UCL}}$ samples (e.g.\ $10^4$ for $\alpha_{\text{UCL}}=10^{-4}$) to estimate the UCL, making it impractical to reset the drift detector during operation. We instead fit the upper tail of the $T^2$ distribution, assuming $T^2$ is stationary after adaptation and approximating it by a scaled non-central $\chi^2$ via moment matching. Although the Hotelling $T^2$ statistic follows an $F$ distribution under certain assumptions \cite{bilodeau1999theory}, this no longer holds here, as the MEWMA induces temporal dependence (see SI). Because tail fitting from limited data is sensitive to outliers, we additionally bootstrap the $T^2$ values to obtain multiple UCL estimates, whose median is used as the drift-detection threshold.

{\color{black}
Third, although the quantile prediction model is trained using a multi-step-ahead loss, in the rest of the work we monitor only the gradients of the one-step-ahead prediction loss as an engineering compromise to reduce detection latency. Monitoring the full prediction horizon requires waiting for all corresponding future observations to become available, introducing additional delays before a drift alarm can be raised. From a dynamical systems perspective, this simplification is reasonable because predictions at successive time steps are temporally dependent. Discrepancies in the immediate system dynamics that lead to increased one-step-ahead prediction errors may also propagate to subsequent prediction steps. Therefore, monitoring the one-step-ahead loss provides a reasonable indicator of potential changes affecting the future prediction horizon, although changes that primarily affect longer-horizon predictions may be less readily detected. Importantly, the one-step-ahead loss is a component of the multi-step training objective, and the proposed detector only monitors its gradients with respect to selected NN parameters. While these gradients are not necessarily zero-mean because the monitored loss does not match the complete training objective, the empirical monitoring principle remains applicable. Specifically, the same one-step-ahead loss and fixed model parameters are used to estimate the reference mean gradients in Phase~I and monitor their deviations in Phase~II consistently. Furthermore, because the full gradient contains parameters associated with different prediction steps, monitoring only the one-step-ahead loss results in zero gradients for parameters exclusively associated with subsequent prediction steps. Consequently, these zero-gradient components can be excluded from covariance estimation and inversion without affecting the computation of the Hotelling $T^2$. Regardless of this simplification, the diagonal covariance approximation discussed previously is not restricted to one-step-ahead monitoring and can be extended to other high-dimensional NN models.}

These modifications enable efficient online Phase~I re-initialization by excluding zero-gradient components, reducing the effective covariance dimension, avoiding full covariance inversion, and estimating the UCL using relatively small batches, making gradient-based drift detection feasible in real-time applications.

\subsection{\emph{How to Update the Digital Twin?} -- Parameter-Efficient Model Adaptation}
\label{sec:how_to_update}

Once drift is detected, the second module updates the surrogate from a limited buffer of streaming data. As discussed in the Introduction, the update mechanism must supply enough plasticity to capture the new regime while preserving the pretrained structure that remains valid, and must do so from few samples without destabilizing ongoing operation. We address the first requirement with LoRA, reviewed in Section~\ref{sec:lora}, and the second with the batch-wise fine-tuning procedure of Section~\ref{sec:fine_tune}.

\subsubsection{Low-Rank Adaptation}
\label{sec:lora}
\begin{figure}[h!]
    \centering
    \includegraphics[width=0.65\linewidth]{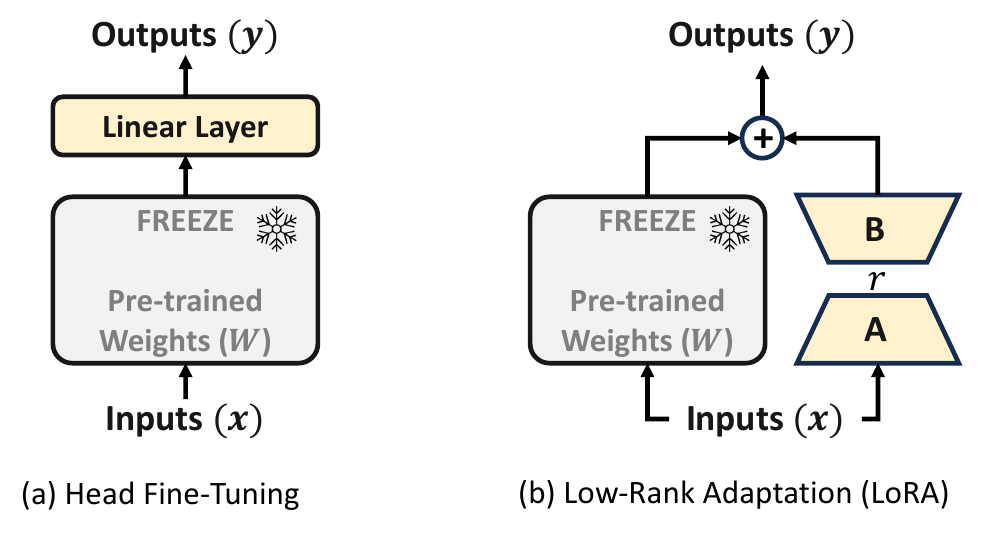}
    \caption{Illustration of fine-tuning methods.}
    \label{fig:LoRA}
\end{figure}
Low-rank adaptation (LoRA) \cite{hu2022lora}, a widely used PEFT method for large language models, constrains task-specific adaptation to a low-rank subspace of the pretrained weights, as illustrated in Figure~\ref{fig:LoRA}(b). By restricting fine-tuning to this low-dimensional subspace, LoRA improves data and parameter efficiency while providing sufficient plasticity for model adaptation, with the pretrained network retained to balance the model's stability. For a linear layer $y=\mathbf{W}x$ with $\mathbf{W}\in\mathbb{R}^{d\times k}$, LoRA freezes $\mathbf{W}$ and learns an additive low-rank update
\begin{equation}
\mathbf{W}' = \mathbf{W} + \Delta \mathbf{W}, \quad \Delta \mathbf{W} = \mathbf{B A},
\end{equation}
with $\mathbf{A}\in\mathbb{R}^{r\times k}$ and $\mathbf{B}\in\mathbb{R}^{d\times r}$, reducing the number of trainable parameters from $dk$ to $r(d+k)$ for $r \ll \min(d,k)$. Initializing $\mathbf{B}=\mathbf{0}$ makes $\Delta \mathbf{W}=\mathbf{0}$ at the start, so the model exactly reproduces the pretrained mapping before adaptation and departs from it only as evidence accumulates from the new regime.

The low-rank constraint is not merely a computational convenience but a structural hypothesis about how adaptation manifests, one that is well matched to concept drift in engineered systems. Empirical studies of over-parameterized networks show that fine-tuning updates possess low intrinsic dimensionality, i.e.\ the weight correction required to specialize a pretrained model resides in a subspace of dimension far smaller than the total number of parameters \cite{aghajanyan2021intrinsic, hu2022lora}. This hypothesis is plausible for the drift scenarios considered here, where physical drift typically arises from parameter variation, wear, or environmental change that perturbs the governing dynamics smoothly rather than restructuring them entirely. The latent representations learned by the pretrained surrogate, encoding temporal features and input--response couplings, therefore remain largely valid under drift, and adaptation reduces to reweighting how these features are combined, a correction naturally expressible as a low-rank perturbation of the weight matrices. This stands in contrast to head fine-tuning, which can only rescale and recombine the \emph{fixed} output features and thus fails when drift alters the internal dynamics that generate those features, and to full fine-tuning, whose unconstrained updates can distort the shared representation from few noisy samples.

From an estimation-theoretic perspective, restricting the update to $r(d+k)$ degrees of freedom directly addresses the low-data pathology of streaming adaptation. With a buffer of only $N_B$ samples, estimating a full-rank correction is severely underdetermined and highly sensitive to measurement noise, whereas the low-rank parameterization shrinks the effective hypothesis space, reducing estimation variance and acting as an implicit regularizer that complements the explicit weight penalty used during fine-tuning. The rank $r$ moreover serves as \textcolor{black}{hyperparameter} for the stability--plasticity trade-off central to continual learning, where small $r$ yields conservative, well-conditioned updates suitable for mild drift, while larger $r$ grants additional plasticity for severe regime changes. However, the selection of the rank highly depends on the available buffer size and fine-tuning requirements, depending on different applications.

LoRA also offers deployment properties that align closely with the architecture of the proposed framework. Because the update is additive to each layer without changing the model's structure, the adapted surrogate incurs negligible inference latency since the rank $r \ll d$, and one not shared by adapter-style methods that insert serial layers into the forward pass \cite{jiahao2023online}. Moreover, the modularity of low-rank updates suggests a natural extension in which validated $\{\mathbf{B}_j\mathbf{A}_j\}$ pairs are archived as a library of regime-specific corrections, enabling rapid re-deployment upon recurrence of previously encountered operating conditions rather than re-adaptation from scratch. This direction is left to future work. Because $\mathbf{W}$ remains frozen throughout, the pretrained backbone is preserved intact, mitigating catastrophic forgetting by construction rather than through auxiliary replay or regularization losses. To benchmark against existing methods, \textcolor{black}{we also experiment with head fine-tuning, i.e., fine-tuning the last layer only (Figure~\ref{fig:LoRA}(a)).}

\subsubsection{Batch-Wise Model Adaptation}
\label{sec:fine_tune}
Once drift is detected, the framework transitions to the model-adaptation stage. Prior to data collection, the system is operated over a full prediction horizon $N$ to ensure that all samples used for fine-tuning are obtained directly from the physical system rather than from model-generated predictions \cite{casagrande2023online}. This step is essential to prevent bias introduced by previously drifted models. Following this initialization, a data batch of size $N_B$ is collected. Adopting a batch-based strategy improves the robustness and generalizability of the fine-tuning process while reducing the risk of overfitting to transient disturbances or anomalous observations, which is equivalent to the stochastic weight averaging (SWA) concept of adapting parameters over a short horizon at every time step \cite{casagrande2023online, schena2024reinforcement, pinosky2023hybrid}. Such a design is also consistent with common practice in model-based reinforcement learning \cite{de2018experience}.

Once the data buffer is full, the collected dataset is partitioned into training and validation subsets using a 9:1 ratio. Rather than performing a contiguous split at a specific time index, validation samples are periodically extracted from the streaming data every 10 steps, with the remaining samples assigned to the training set. This interleaved splitting strategy ensures that both subsets remain representative of the same underlying data distribution, which is particularly important in non-stationary environments \cite{gasparino2023unmatched}. As a result, the validation process more reliably reflects the model's generalization performance under evolving system dynamics.

With the dataset prepared, the idle model is fine-tuned using the training set. Optimization is carried out with the \texttt{Adam} optimizer for 100 epochs using mini-batch updates. To mitigate overfitting, a regularization term is incorporated into the loss function, and the validation loss is monitored at each epoch; the model achieving the lowest validation loss is selected as the updated model. Within this framework, the three fine-tuning strategies introduced in Section~\ref{sec:lora} are evaluated.

Finally, the proposed adaptation strategy imposes minimal computational overhead for real-time deployment. Although each fine-tuning cycle uses larger batches and more epochs than per-step updates, it runs entirely on the idle model and does not interfere with ongoing operation. Moreover, because adaptation is followed by a validation stage that itself requires collecting a batch of streaming data, the fine-tuning cost is effectively hidden when executed in parallel with validation-batch collection. The framework also permits fine-tuning to run on hardware separate from the real-time controller, avoiding any additional burden or latency in time-critical optimization.

\subsection{\emph{Is the Update Successful?} -- Model Validation}
\label{sec:validation}
Once the idle model has been fine-tuned, the framework proceeds to the model-validation stage, where its predictive performance is formally assessed to determine whether it should replace the live model. Specifically, we compare the supervised-learning loss produced by the live and idle models on a validation dataset. This validation set is strictly held out from the fine-tuning process to prevent data leakage and ensure an unbiased evaluation of generalization performance. An additional batch of data of size $N_V$ is therefore collected after the adaptation stage to construct the validation set.

{\color{black}
Given the validation data, the live and idle models receive the same validation input, and their predictions are compared against the ground truth using the quantile loss. Specifically, the quantile loss is computed over the entire prediction horizon for each validation sample to evaluate the multi-step-ahead predictive performance. This yields paired loss observations from the two models. The validation task is formulated as a hypothesis test of whether the idle model achieves a statistically significant reduction in loss relative to the live model. As the loss distributions are not guaranteed to be Gaussian, we employ a one-sided paired hypothesis test using moving block bootstrapping (MBB) to generate random samples from loss sequences that account for their temporal dependence.

The paired loss differences are first calculated as $d_t = \mathcal{L}_{\text{idle},t} - \mathcal{L}_{\text{live},t}$ across the entire validation set, where $\mathcal{L}_{\text{idle},t}$ and $\mathcal{L}_{\text{live},t}$ denote the full-horizon quantile losses of the idle and live models for validation sample $t$, respectively. To account for temporal dependence among these differences, MBB is employed with a block length of three, assuming that temporal correlations induced by system disturbances diminish over time. Using MBB, $n_{\text{MBB}}=10,000$ bootstrap time series are generated by resampling consecutive blocks of the paired loss differences. The mean of each bootstrap time series is then calculated, yielding $n_{\text{MBB}}$ bootstrap estimates of the mean difference, $\bar{d}^{*(b)}$, $b\in[1,\ldots,n_{\text{MBB}}]$. The resulting bootstrap distribution is used to assess whether the idle model achieves a lower expected quantile loss than the live model through the following one-sided hypothesis test:
\begin{align}
    H_0 &: \mathbb{E}[d_t] \geq 0, \\
    H_1 &: \mathbb{E}[d_t] < 0.
\end{align}
The significance level is set to $\alpha_{\text{test}}=0.05$, corresponding to a one-sided confidence level of 95\%. Rejection of $H_0$ indicates statistically significant evidence that the idle model achieves a lower expected loss than the live model, implying improved predictive performance. In this case, the idle model replaces the live model for subsequent decision-making. Otherwise, the evidence is insufficient to justify replacement, and the framework returns to the adaptation stage, where the idle model continues to be refined as additional data become available. The flowchart of the model validation stage is illustrated in Figure~\ref{fig:U_validation}.

\begin{figure}[h!]
    \centering
    \includegraphics[width=1\linewidth]{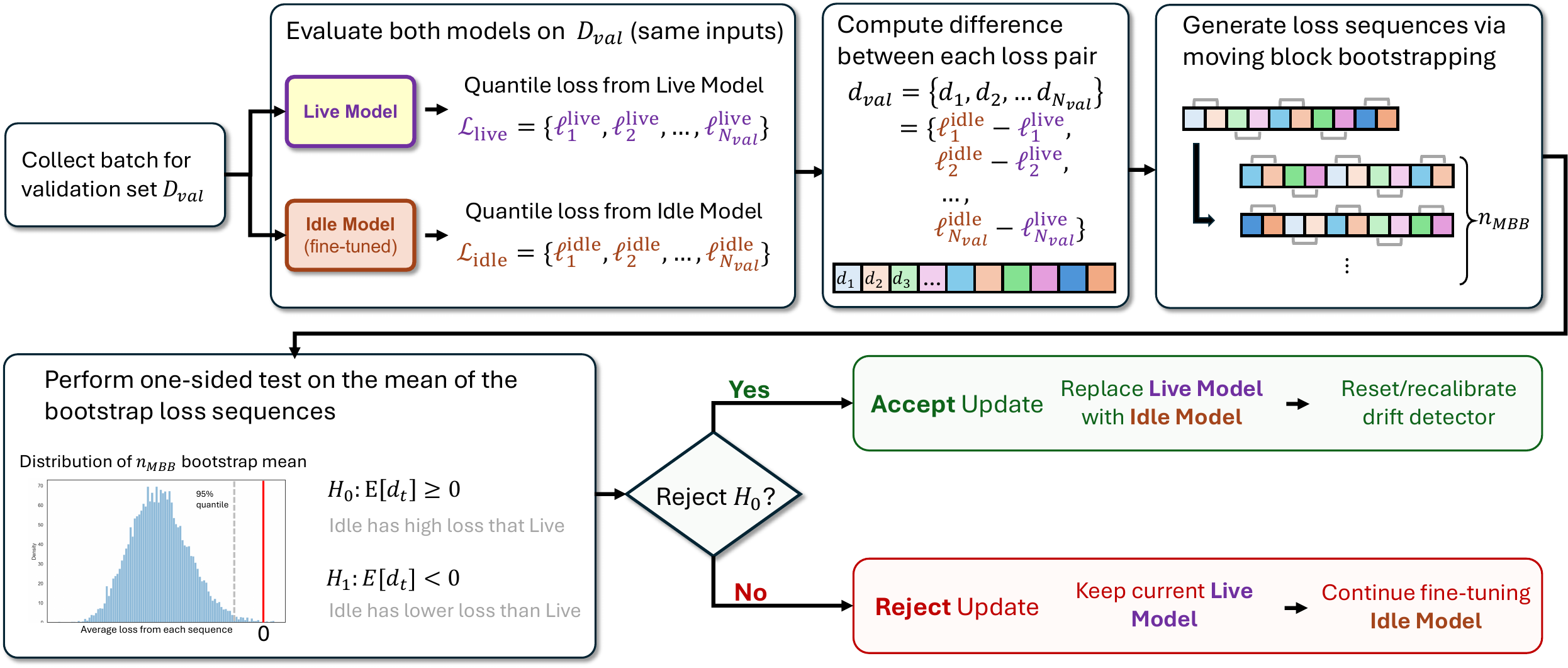}
    \caption{Flowchart of applying the one-sided paired hypothesis test using moving block bootstrapping for model validation.}
    \label{fig:U_validation}
\end{figure}

}

Ideally, one would compare the magnitudes of the score vectors or the Hotelling $T^2$ statistic rather than the loss values. However, comparing the Hotelling $T^2$ requires re-estimation of both $\bar{\mathbf{s}}$ and $\hat{\Sigma}^{-1}$, which is not sufficiently data-efficient in an online setting. Consequently, as a pragmatic compromise, we adopt direct comparison of loss values for online validation.

Following model replacement, the framework re-enters Phase~I to reinitialize the drift detector using the updated model, as described in Section~\ref{sec:online_implementation}. Collectively, the integration of drift detection, model adaptation, and statistical validation establishes an autonomous framework that enables the Digital Twin to continuously adapt to evolving operating conditions throughout its life cycle. While the proposed batch-wise adaptation strategy may exhibit slower responsiveness than step-wise updates, it provides a statistical criterion for model improvement through hypothesis testing, reducing the risk of replacing the live model with an updated model that overfits transient data.

We argue that the model-validation stage is essential even when validation loss is monitored during fine-tuning. This design follows the standard training--validation--testing paradigm in supervised learning, where the testing stage evaluates a model's ability to generalize to unseen data. Such evaluation is essential for Digital Twins operating under non-stationary conditions, where the fine-tuning and validation datasets are sampled from the same distribution, whereas the input distribution, $p(x)$, of future streaming data may drift from this distribution even when the underlying concept, $p(y \mid x)$, remains unchanged. Consequently, an additional validation stage on unseen data provides a more reliable assessment of generalization and can reveal overfitting not apparent from the fine-tuning loss alone. For instance, a monotonically decreasing validation loss during fine-tuning does not guarantee deployability; failure to reject $H_0$ indicates that there is insufficient evidence of improvement of the updated model on newly arriving data and therefore should not replace the live model. In such cases, a more appropriate strategy is to continue fine-tuning as additional streaming data becomes available, leveraging the adaptability of the pretrained network. \textcolor{black}{Furthermore, repeated failures to reject the null hypothesis, despite collecting additional fine-tuning data, may indicate that automated model adaptation cannot achieve sufficient improvement. If validation fails for three consecutive adaptation attempts (in this work), the system can be stopped and human intervention requested to investigate the underlying causes, providing an additional safeguard against continued operation with an inadequate model.} Alternative approaches, such as adaptive regularization schemes or improved data-buffer strategies to enhance generalization, are promising directions but are beyond the scope of this work.

\section{Illustrative Example}
\label{sec:score_based_ex}

In this section, the illustrative example introduced in \cite{chen2026uncertainty} is used to walk through the proposed method and to benchmark against existing fine-tuning approaches. A discrete-time linear system with states $\mathbf{x}=[x_1,x_2]^\top$ is considered for demonstration. Initially, the system is subject to additive stochastic disturbance on $x_2$, where $\epsilon_k \sim \mathcal{N}(0, 1^2)$, described by
\begin{align}
\mathbf{x}_{k+1} &= \mathbf{A}\mathbf{x}_k + \mathbf{B}u_k + \mathbf{w}_k\\ &= 
\begin{bmatrix}
0.3 & 0.1 \\
0.1 & 0.2
\end{bmatrix} \mathbf{x}_k 
+ 
\begin{bmatrix}
0.5 \\ 
1
\end{bmatrix} u_k 
+ \begin{bmatrix} 
    0\\0.1
\end{bmatrix} \epsilon_k \\
&= F_w(\mathbf{x}_k, u_k, \mathbf{w}_k),\\
\mathbf{y}_{k+1} & = \mathbf{x}_{k+1}
\label{eq:toy_problem_sys}
\end{align}
where the disturbance follows a Gaussian distribution due to the linear transformation of $\epsilon_k$.

\subsection{Building the Virtual Twin: Surrogate Modeling}

To construct training data for system identification with TiDE, an input sequence $\mathbf{D}_u = \{u_0, \ldots, u_{n-1}\}$ is generated, where $u_k \in [-5,5]$ is uniformly sampled and $n = 422{,}000$. Starting from the initial condition $\mathbf{x}_0 = [0,0]^T$, the corresponding state trajectory $\mathbf{D}_x = \{\mathbf{x}_1, \ldots, \mathbf{x}_n\}$ is obtained by forward simulation. The detailed training configuration is summarized in Table~\ref{tab:tide_training}.

The resulting sequences are segmented using a sliding window of length $w + N$, where the window size is $w = 10$ and the prediction horizon is $N = 10$. For the $l$th segment, define
\begin{align}
\mathbf{D}_{u,l} &= \{u_l, \ldots, u_{l+w+N-1}\}, \\
\mathbf{D}_{x,l} &= \{\mathbf{x}_{l+1}, \ldots, \mathbf{x}_{l+w+N}\}.
\end{align}
Each segment is then decomposed into past and future components:
\begin{align}
\mathbf{x}^p_l &= [\mathbf{x}_{l+1}, \ldots, \mathbf{x}_{l+w}], \quad
\mathbf{x}^f_l = [\mathbf{x}_{l+w+1}, \ldots, \mathbf{x}_{l+w+N}], \\
\mathbf{u}^p_l &= [u_l, \ldots, u_{l+w-1}], \quad
\mathbf{u}^f_l = [u_{l+w}, \ldots, u_{l+w+N-1}],
\end{align}
where the superscripts $p$ and $f$ denote the past and future components, respectively. The tuple $(\mathbf{x}^p_l, \mathbf{u}^p_l, \mathbf{u}^p_l)$ serves as input and $\mathbf{x}^f_l$ as the prediction target. 

All segments are randomly partitioned into training, validation, and test sets with a ratio of 8:1:1. Following Eq.~(\ref{eq:TiDE_general}), the TiDE model is trained via supervised learning using the quantile loss:
\begin{subequations}
\begin{align}
\min_{\boldsymbol{\phi}} \quad & L_Q(\mathbf{x}^f, \hat{\mathbf{x}}^f), \\
\text{s.t.} \quad & \hat{\mathbf{x}}^f = \text{TiDE}(\mathbf{x}^p, \mathbf{u}^p, \mathbf{u}^f \mid \boldsymbol{\phi}).
\end{align}
\end{subequations}

As the focus is the adaptive framework rather than surrogate accuracy, we report only that the trained TiDE attains $\approx 5\%$ MAPE and $\approx 0.04$ NRMSE on both states (see \cite{chen2026uncertainty} for full validation), which is sufficient for the following tasks.

\begin{table}[t]
\centering
\caption{\textcolor{black}{Hyperparameters} of training TiDE model}
\label{tab:tide_training}
\scriptsize
\begin{tabular}{ccccccc}
\hline \hline
\multicolumn{7}{c}{Details for TiDE model setup}                                                                                                                                                                                                                         \\ \hline
\multicolumn{1}{c|}{\# encoder layers} & \multicolumn{1}{c|}{\# decoder\_layers} & \multicolumn{1}{c|}{decoder output dim.} & \multicolumn{1}{c|}{hidden size} & \multicolumn{1}{c|}{decoder hidden size} & \multicolumn{1}{c|}{dropout rate} & layer normalization \\ \hline
\multicolumn{1}{c|}{1}                 & \multicolumn{1}{c|}{1}                  & \multicolumn{1}{c|}{16}                  & \multicolumn{1}{c|}{128}         & \multicolumn{1}{c|}{32}                           & \multicolumn{1}{c|}{0.2}          & True                \\  \hline \hline
\multicolumn{7}{c}{Details for TiDE model training}                                                                                                                                                                                                                            \\ \hline
\multicolumn{1}{c|}{learning rate}     & \multicolumn{1}{c|}{regularization}     & \multicolumn{1}{c|}{step\_size}          & \multicolumn{1}{c|}{rate decay}  & \multicolumn{1}{c|}{\# epoch}                     & \multicolumn{1}{c|}{batch size}   & shuffle data        \\ \hline
\multicolumn{1}{c|}{0.001}             & \multicolumn{1}{c|}{0.002}              & \multicolumn{1}{c|}{10}                   & \multicolumn{1}{c|}{0.95}        & \multicolumn{1}{c|}{1500}                          & \multicolumn{1}{c|}{64}          & True                \\ \hline \hline
\end{tabular}
\end{table}

\subsection{Uncertainty-Aware Decision-Making through Robust Model Predictive Control}

The physical system and its virtual twin interact through a robust MPC: at each step, measurements from the physical system update the state for predicting quantiles through TiDE, which is then used in robust MPC to build safety margins for explicit constraint handling under uncertainty, and the optimal input is applied to the physical system.

The objective of this example is to regulate the state $x_1$ to track a reference trajectory while ensuring that $x_1$, $x_2$, and the control input $u$ remain within admissible bounds in the presence of unbounded disturbances. The multi-step robust MPC formulation, leveraging quantile predictions from TiDE, is given by
\begin{subequations}
\label{eq:MPC_toy}
\begin{align}
    \min_{\mathbf{v}} \quad & J(\mathbf{v},\tilde{\hat{\mathbf{x}}},\mathbf{r}) 
    = \sum_{i=0}^{N-1} \|\tilde{\hat{x}}_{1,k+i+1} - r_{k+i+1}\|_{\mathbf{Q}}^2 
    + \|v_{k+i}\|_{\mathbf{R}}^2 \\
    \text{s.t.} \quad 
    & \hat{\mathbf{x}}_{k+1}^f = [\bar{\hat{\mathbf{x}}}_{k+1}^{f}, \tilde{\hat{\mathbf{x}}}_{k+1}^{f}, \underline{\hat{\mathbf{x}}}_{k+1}^{f}]^T  = \text{TiDE}(\mathbf{x}_{k}^p, \mathbf{u}_{k}^p, \mathbf{v}), \\
    & \mathbf{v} = [v_k,\hdots,v_{k+N-1}], \\
    & \text{Pr}\left(\hat{x}_{1,k+i} \geq -2 \right) \geq 0.95, \quad \forall i \in \mathbb{N}_{[1,N]}, \\
    & \text{Pr}\left(\hat{x}_{1,k+i} \leq 2.5 \right) \geq 0.95, \quad \forall i \in \mathbb{N}_{[1,N]}, \\
    & \text{Pr}\left(\hat{x}_{2,k+i} \geq -3.5 \right) \geq 0.95, \quad \forall i \in \mathbb{N}_{[1,N]}, \\
    & \text{Pr}\left(\hat{x}_{2,k+i} \leq 3.5 \right) \geq 0.95, \quad \forall i \in \mathbb{N}_{[1,N]}, \\
    & v_{k+i} \in \mathbb{U} \ominus \mathbf{K}\mathbb{Z}_{k+i}, \quad \forall i \in \mathbb{N}_{[0,N-1]}, \\
    & u_k = \pi(v_k) = v_k + \mathbf{K}\mathbf{e}_k, \\
    & \mathbf{u}_{k}^p = [u_{k-w},...,u_{k-1}].
\end{align}
\end{subequations}

The weighting matrices are chosen as $\mathbf{Q} = \mathbf{I}$ and $\mathbf{R} = \mathbf{I}$. The admissible input set is $\mathbb{U} \in [-5,5]$, adaptively tightened based on the prediction-error bound $\mathbb{Z}$. The feedback gain is selected as $\mathbf{K} = [-0.0621, -0.2027]$. Since the predicted upper and lower quantiles inherently capture the probabilistic bounds on the states, the chance constraints in (\ref{eq:MPC_toy}) can be equivalently reformulated as deterministic constraints:
\begin{subequations} 
\begin{align}
    & \bar{\hat{x}}_{1,k+i} \leq 2.5, \quad \underline{\hat{x}}_{1,k+i} \geq -2, \\
    & \bar{\hat{x}}_{2,k+i} \leq 3.5, \quad \underline{\hat{x}}_{2,k+i} \geq -3.5.
\end{align}
\end{subequations}

\subsection{Phase~I: Drift Detector Setup}

To enable online gradient-based detection \textcolor{black}{and LoRA for PEFT,} the pretrained TiDE (187,278 parameters) is frozen, LoRA layers are added to each layer, and an additional linear output layer producing the score vector is appended. The score is defined as the gradient of the loss with respect to this final linear layer (3,600 parameters). \textcolor{black}{These parameters are used exclusively for gradient-based monitoring and are not fine-tuned. Similarly, the LoRA parameters (2,670 parameters) remain frozen during drift detection and are unfrozen only during model adaptation.}  

To calibrate the drift detector in Phase~I, 700 undrifted steps are first collected. The first 200 estimate the score mean $\bar{\mathbf{s}}$ and covariance $\hat{\Sigma}$ (with $\hat{\Sigma}^{-1}$ from diagonal inversion), and the remaining 500 give the sequential MEWMA $T^2$ statistic (Eqs.~\ref{eq:mewma}--\ref{eq:t2}, $\lambda=0.05$). Note that although TiDE is trained with a multi-step-ahead quantile loss, the score vector is computed from the one-step-ahead quantile loss.  

\begin{figure}[h!]
    \centering
    \includegraphics[width=1\linewidth]{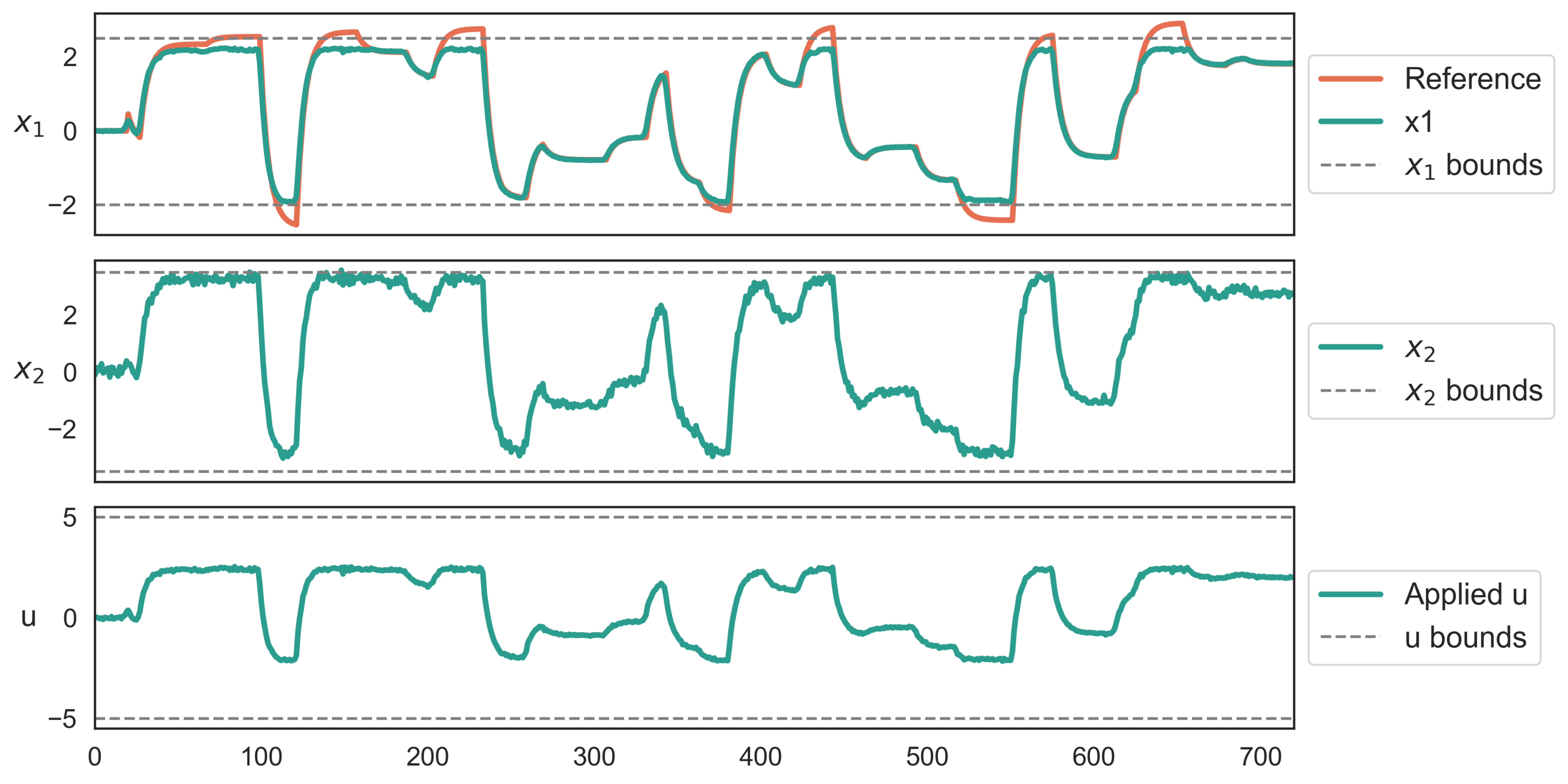}
    \caption{Trajectory of RMPC in the in-control condition for setting up the drift detector in Phase~I}
    \label{fig:phase_1_traj}
\end{figure}

Figure~\ref{fig:phase_1_traj} illustrates the trajectory of the multi-step RMPC under in-control conditions. The proposed RMPC framework effectively trades off reference-tracking performance to maintain constraint satisfaction in the presence of fluctuations, enabled by the safety margins provided by the learned quantiles.

\begin{figure}[h!]
    \centering
    \includegraphics[width=0.8\linewidth]{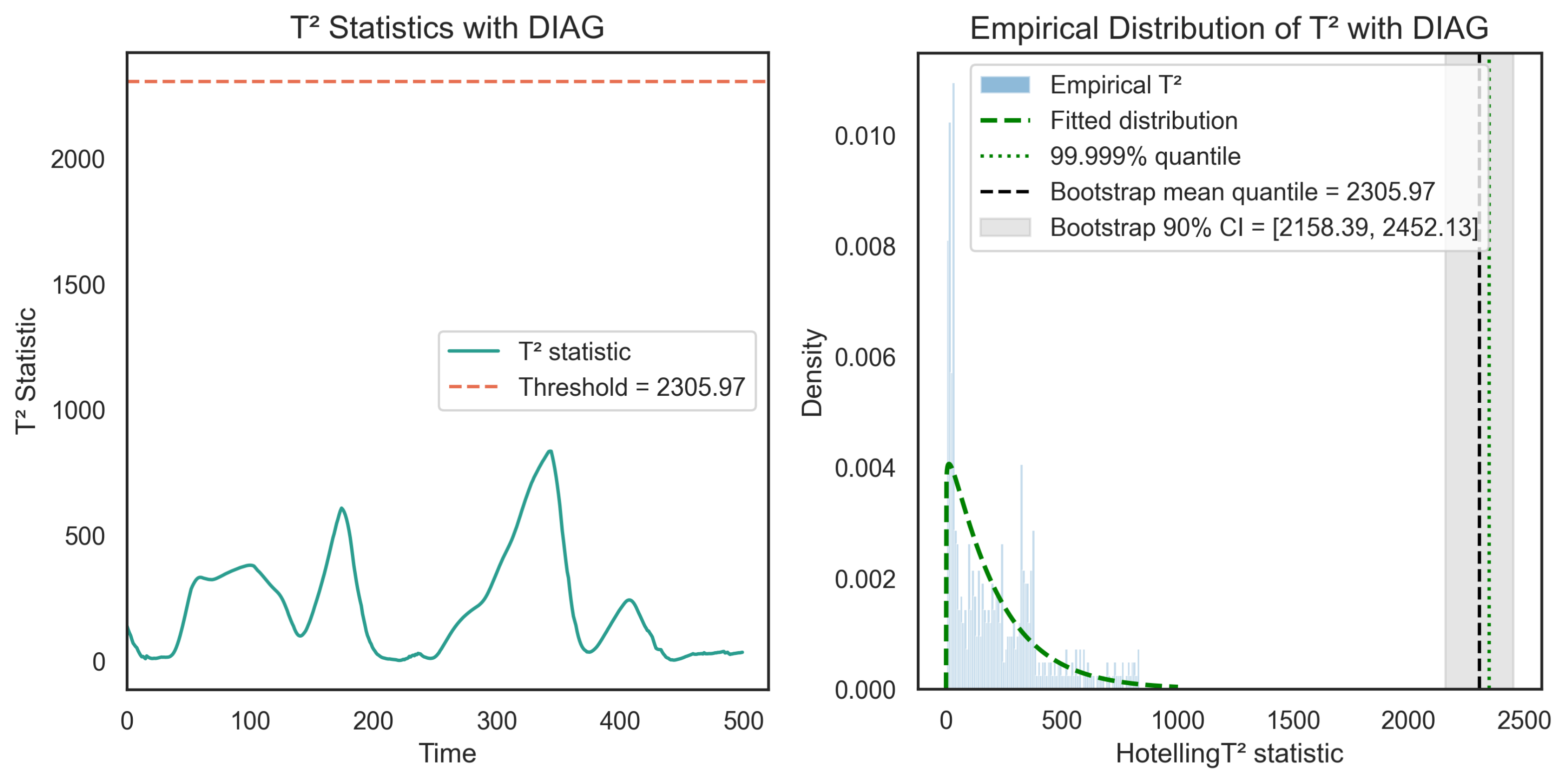}
    \caption{Hotelling $T^2$ statistics under in-control conditions for \textcolor{black}{estimated UCL}. (a) The $T^2$ trajectory of the streaming data. Under stationary conditions, the $T^2$ statistic fluctuates around a stable level and remains below the threshold, indicating the absence of false alarms. (b) Empirical distribution of $T^2$ in Phase~I, well approximated by the fitted scaled noncentral $\chi^2$ distribution.}
    \label{fig:phase_1_diag_dist}
\end{figure}

The corresponding $T^2$ trajectory and its histogram are presented in Figures~\ref{fig:phase_1_diag_dist}(a) and \ref{fig:phase_1_diag_dist}(b), respectively. In Figure~\ref{fig:phase_1_diag_dist}(a), the horizontal dashed line represents the detection threshold. Under stationary conditions, the $T^2$ statistic fluctuates around a stable level and remains below the threshold. Figure~\ref{fig:phase_1_diag_dist}(b) further shows that the empirical distribution of $T^2$ is well approximated by the fitted scaled noncentral $\chi^2$ distribution. 

\subsection{Phase~II: Online Drift Detection, Adaptation, and Validation}
\label{sec:phase2}

In Phase~II the tracking task runs for 3,000 steps with abrupt changes at $k=200$ and $k=1500$. At $k=200$ the system enters concept $P_1(y\mid x)$, injecting three uncertainties: nonlinear terms for previously unmodeled dynamics, modified $\mathbf{A},\mathbf{B}$ for changed intrinsic dynamics, and a noise reduction $\mathcal{N}(0,0.1^2)\!\to\!\mathcal{N}(0,0)$. \textcolor{black}{The noise reduction under concept $P_1$ was intentionally introduced to evaluate the framework's ability to adapt to changes in aleatoric uncertainty, in addition to changes in the underlying system dynamics.} The resulting state evolution under concept $P_1(y \mid x)$ is given by
\begin{align}
\mathbf{x}_{k+1}
&=
\begin{bmatrix}
0.5 & 0.04\\
0.2 & 0.2
\end{bmatrix}\mathbf{x}_k
+
\begin{bmatrix}
0.5\\
1
\end{bmatrix}u_k
+
0.3
\begin{bmatrix}
1 & 0\\
0 & 1
\end{bmatrix}
\tanh(\mathbf{x}_k)
+
0.1
\begin{bmatrix}
1\\
1
\end{bmatrix}
\tanh(u_k)
+
\begin{bmatrix}
0\\
0
\end{bmatrix}\epsilon_k, 
\label{eq:toy_nonlinear_matrix_p1}
\end{align}
where $\epsilon_k \sim \mathcal{N}(0, 1)$.

At $k = 1500$, the system transitions to another regime, concept $P_2(y \mid x)$. Here the disturbance is allowed to affect both state variables, representing an increase in aleatoric uncertainty. In addition, the system matrices are altered again, introducing further epistemic uncertainty. The resulting system under concept $P_2(y \mid x)$ is
\begin{align}
\mathbf{x}_{k+1}
&=
\begin{bmatrix}
0.6 & 0.25\\
0.2 & 0.4
\end{bmatrix}\mathbf{x}_k
+
\begin{bmatrix}
0.5\\
1
\end{bmatrix}u_k
+
0.3
\begin{bmatrix}
1 & 0\\
0 & 1
\end{bmatrix}
\tanh(\mathbf{x}_k)
+
0.1
\begin{bmatrix}
1\\
1
\end{bmatrix}
\tanh(u_k)
+
\begin{bmatrix}
0.1\\
0.1
\end{bmatrix}\epsilon_k.
\label{eq:toy_nonlinear_matrix_p2}
\end{align}

\begin{figure}[h!]
    \centering
    \includegraphics[width=1\linewidth]{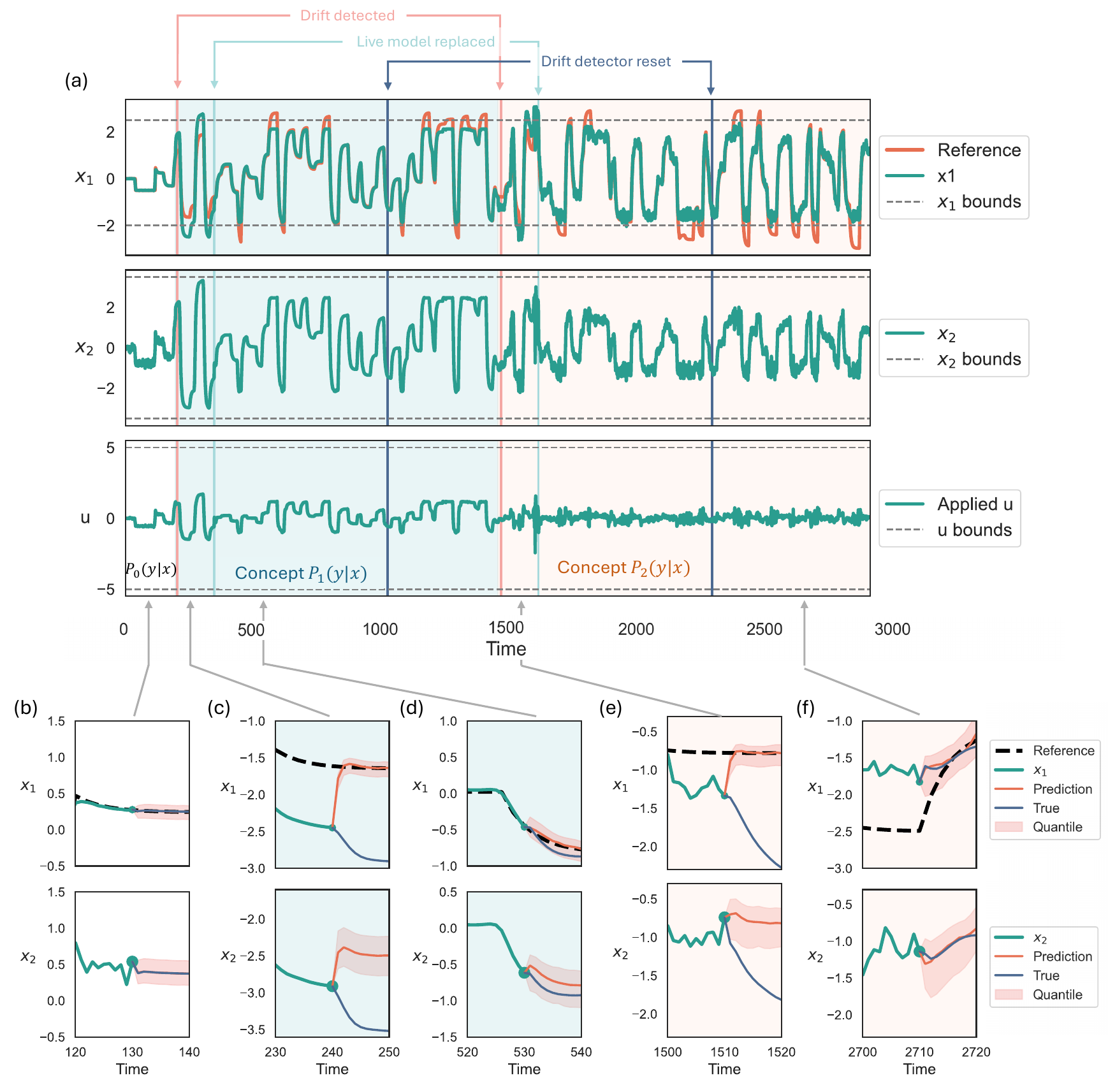}
    \caption{Trajectories of the adaptive Digital Twin in Phase~II. The figure shows the trajectories of $x_1$, $x_2$, and $u$ under the proposed framework with two different injected uncertainties. The white background denotes the in-control region, the green background a different concept $P_1(y|x)$, and the orange background denotes another concept $P_2(y|x)$. The results demonstrate that the system effectively detects the drift and adapts to restore control performance under both types of drift.}
    \label{fig:toy_all_traj}
\end{figure}

{\color{black}

We show the resulting trajectory using \texttt{ADT-LoRA}, with LoRA of rank $r=1$, with buffer $N_B=100$ and validation set $N_V=30$. During fine-tuning, the learning rate is set to $10^{-3}$ using the \texttt{Adam} optimizer. The model is trained for 100 epochs with a mini-batch size of 32. In addition, a regularization with coefficient $1 \times 10^{-4}$ is incorporated into the loss function to mitigate overfitting. These hyperparameters are selected from preliminary experiments to ensure stable convergence and effective adaptation but are not exhaustively optimized; their effect on adaptation performance can be case-dependent and is left to future work.

Figure~\ref{fig:toy_all_traj}(a) shows the trajectories of $x_1$ and $x_2$ under the proposed adaptive Digital Twin framework. The results demonstrate effective drift detection and adaptation, restoring control performance under the injected uncertainties. Shaded regions denote different concept regimes, while drift detection, model replacement, and detector reset events are explicitly marked. The first drift (at $k=200$) is detected at $k=207$ and the live model is replaced at $k=356$, after which the detector resets at $k=1056$. The second drift (at $k=1500$) is detected at $k=1513$, and the live model is replaced at $k=1663$, with a detector reset at $k=2363$.


Figure~\ref{fig:toy_all_traj}(b)--(f) present the trajectories and quantile predictions in different regimes. In particular, the model predictions given the optimal control inputs are compared with the noise-free ground-truth state trajectories from Equation~\ref{eq:toy_problem_sys}. Before drift, the system accurately tracks the reference, and the TiDE predictions closely match the true trajectories (Figure~\ref{fig:toy_all_traj}(b)). After the first concept drift, the pretrained model fails to capture the new dynamics, resulting in large prediction errors (Figure~\ref{fig:toy_all_traj}(c)). Once drift is detected, the model is updated with data from the new regime, providing improved mean and quantile predictions (Figure~\ref{fig:toy_all_traj}(d)). 

At the second drift stage, the model again fails to capture the new dynamics (Figure~\ref{fig:toy_all_traj}(e)) before being updated. After drift detection and adaptation, prediction accuracy is significantly improved (Figure~\ref{fig:toy_all_traj}(f)). Despite the increased noise variance, the updated TiDE can predict the nominal trajectory while widening its predictive quantiles to reflect the higher aleatoric uncertainty. Overall, these results demonstrate that the proposed adaptive Digital Twin framework effectively detects and adapts under both types of drift.

\begin{figure}[h!]
    \centering
    \includegraphics[width=\linewidth]{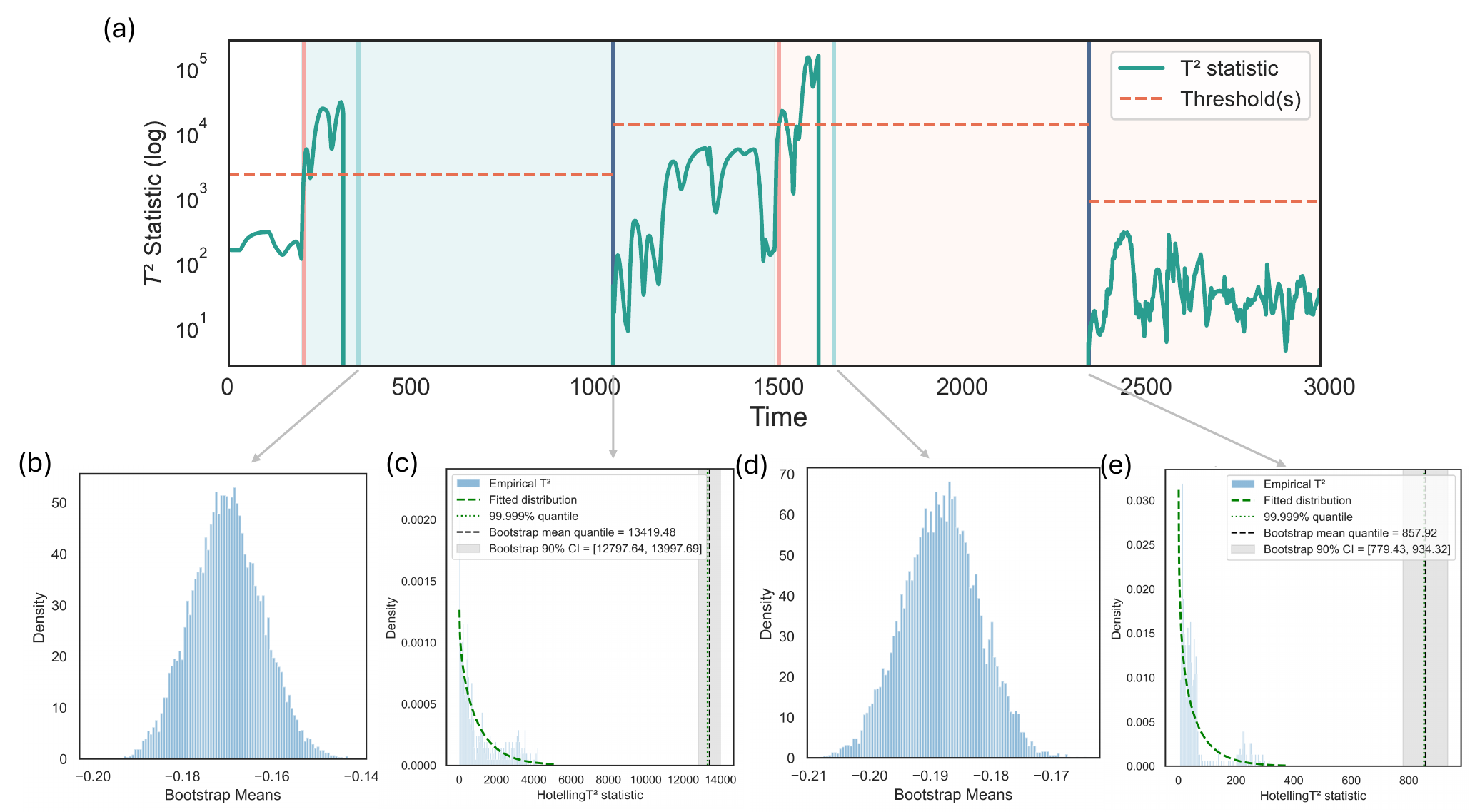}
    \caption{The trajectory and estimated UCL of Hotelling $T^2$, and the validation performance of the closed-loop operation under each stage. (a) shows the trajectory of Hotelling $T^2$ under each concept once the drift detector is reset. (b) and (d) show the one-sided paired test result after the idle model is fine-tuned under $P_1$ and $P_2$. (c) and (e) show the fitted non-central $\chi^2$ distribution of the Hotelling $T^2$ and the estimated UCL after the model is successfully fine-tuned under $P_1$ and $P_2$, respectively. }
    \label{fig:t2_toy}
\end{figure}

Figure~\ref{fig:t2_toy}(a)--(e) presents the evolution of the Hotelling $T^2$ statistic, the online validation results, and the estimated $T^2$ distributions following successful model updates. As shown in Figure~\ref{fig:t2_toy}(a), the $T^2$ statistic increases rapidly upon transitions to new operating concepts, successfully triggering drift alarms. Following model updates and detector recalibration at $k=1056$ and $k=2363$, the subsequent $T^2$ values remain largely below the estimated UCLs, as further illustrated in Figures~\ref{fig:t2_toy}(c) and (e). These results suggest that, with an appropriately updated model, the recalibrated detector can effectively characterize the in-control distribution of the $T^2$ statistic, provided that the underlying gradient statistics remain approximately stationary. Furthermore, Figures~\ref{fig:t2_toy}(b) and (d) present the bootstrap distributions of the mean paired loss differences obtained using MBB, supporting the successful validation of the updated models.

Collectively, these results demonstrate the capability of the proposed Adaptive Digital Twin framework to autonomously detect concept drift, update and validate the surrogate model, and recalibrate the drift detector using streaming data. The framework accommodates changes in both system dynamics (epistemic uncertainty) and disturbance characteristics (aleatoric uncertainty), highlighting its potential for reliable, long-term operation under evolving conditions.

}

\subsection{Benchmarking on Model Accuracy}

To jointly assess deterministic accuracy and the quality of the learned quantiles, we report four metrics evaluated against the nominal ground-truth responses: the normalized root mean square error (NRMSE), empirical coverage of the $90\%$ prediction interval, the normalized negatively oriented interval score (NNOIS)~\cite{negarandeh2026non}, and the quantile loss. \textcolor{black}{Although coverage against nominal responses does not strictly measure probabilistic calibration, these metrics remain useful for comparing fine-tuning strategies. Specifically, since the pretrained model is adapted using limited streaming data rather than retrained on the new concept, residual prediction errors may persist, and the learned quantiles may not converge to the nominal response even under noise-free conditions. Therefore, coverage evaluates the inclusion of nominal responses within the predicted intervals, while NNOIS additionally penalizes interval width and deviations outside the intervals. These metrics provide complementary measures of interval quality during adaptation, rather than rigorous assessments of statistical calibration.} Together, they characterize median accuracy, interval sharpness, and online learning quality; their full definitions are given in the Supplementary Information (SI).

\newcolumntype{C}[1]{>{\centering\arraybackslash}m{#1}}

\begin{table}[h!]
\caption{Benchmark approaches for the illustrative example}
{\color{black}
\label{table:benchmark_list}
    \centering
    \scriptsize
    \begin{tabularx}{\textwidth}{
    p{2.5cm}
    >{\raggedright\arraybackslash}X
    C{2.5cm}
    C{1.5cm}
    C{2cm}
    C{1.5cm}
}
\hline\hline
Method & \multicolumn{1}{c}{Description}
       & \multicolumn{1}{c}{Parameters}
       & Detection
       & Fine-Tuning Frequency
       & Validation \\ \hline

\texttt{No FT}
& No fine-tuning approach applied.
& \multicolumn{1}{c}{---}
& \xmark
& \xmark
& \xmark \\ \hline

\texttt{Head FT}
& Fine-tuning the last layer every iteration, which is a common practice in most of the existing literature.
& \begin{tabular}[c]{@{}l@{}}
    lr=1e-3\\
    $\lambda_{\text{reg}}$=1e-4
  \end{tabular}
& \xmark
& Stepwise
& \xmark \\ \hline

\texttt{Full}
& Fine-tuning the entire model every iteration.
& \begin{tabular}[c]{@{}l@{}}
    lr=1e-4\\
    $\lambda_{\text{reg}}$=1e-4
  \end{tabular}
& \xmark
& Stepwise
& \xmark \\ \hline

\texttt{LoRA}
& Applying LoRA ($r=1$) and fine-tuning the model every iteration.
& \begin{tabular}[c]{@{}l@{}}
    lr=1e-3\\
    $\lambda_{\text{reg}}$=1e-4
  \end{tabular}
& \xmark
& Stepwise
& \xmark \\ \hline

\texttt{ADT-Head FT-N}
& Applying the proposed method with head fine-tuning as the fine-tuning approach, but without online validation after fine-tuning.
& \begin{tabular}[c]{@{}l@{}}
    lr=1e-3\\
    $\lambda_{\text{reg}}$=1e-4\\
    $N_B$=100\\
    $N_{\text{epoch}}$=100
  \end{tabular}
& \cmark
& Batch
& \xmark \\ \hline

\texttt{ADT-Head FT}
& Applying the proposed method with head fine-tuning as the fine-tuning approach, with online validation after fine-tuning.
& \begin{tabular}[c]{@{}l@{}}
    lr=1e-3\\
    $\lambda_{\text{reg}}$=1e-4\\
    $N_B$=100\\
    $N_V$=30\\
    $N_{\text{epoch}}$=100
  \end{tabular}
& \cmark
& Batch
& \cmark \\ \hline

\texttt{ADT-Full-N}
& Applying the proposed method to fine-tune the entire model, but without online validation after fine-tuning.
& \begin{tabular}[c]{@{}l@{}}
    lr=1e-4\\
    $\lambda_{\text{reg}}$=1e-4\\
    $N_B$=100\\
    $N_{\text{epoch}}$=100
  \end{tabular}
& \cmark
& Batch
& \xmark \\ \hline

\texttt{ADT-Full}
& Applying the proposed method to fine-tune the entire model, with online validation after fine-tuning.
& \begin{tabular}[c]{@{}l@{}}
    lr=1e-4\\
    $\lambda_{\text{reg}}$=1e-4\\
    $N_B$=100\\
    $N_V$=30\\
    $N_{\text{epoch}}$=100
  \end{tabular}
& \cmark
& Batch
& \cmark \\ \hline

\texttt{ADT-LoRA-N}
& Applying the proposed method with LoRA ($r=1$) as the fine-tuning approach, but without online validation after fine-tuning.
& \begin{tabular}[c]{@{}l@{}}
    lr=1e-3\\
    $\lambda_{\text{reg}}$=1e-4\\
    $N_B$=100\\
    $N_{\text{epoch}}$=100
  \end{tabular}
& \cmark
& Batch
& \xmark \\ \hline

\texttt{ADT-LoRA}
& Applying the proposed method with LoRA ($r=1$) as the fine-tuning approach, with online validation after fine-tuning.
& \begin{tabular}[c]{@{}l@{}}
    lr=1e-3\\
    $\lambda_{\text{reg}}$=1e-4\\
    $N_B$=100\\
    $N_V$=30\\
    $N_{\text{epoch}}$=100
  \end{tabular}
& \cmark
& Batch
& \cmark \\ \hline\hline

    \end{tabularx}

\vspace{2pt}
\begin{minipage}{\textwidth}
\footnotesize
\cmark = Enabling a specific module, \xmark = Disabling a specific module,
"lr" in this table denotes the learning rate, and $\lambda_{\text{reg}}$ is the regularization.
\end{minipage}}
\end{table}

We benchmark the proposed Adaptive Digital Twin (\texttt{ADT}) framework, together with several of its variants, against existing online fine-tuning methods, as listed in Table~\ref{table:benchmark_list}. These benchmarks isolate the effect of four design dimensions:
\begin{itemize}
    \item \texttt{ADT} vs.\ non-\texttt{ADT}: isolates the effect of the proposed framework relative to conventional fine-tuning. The key distinction is that \texttt{ADT} embed drift detection, batch-wise adaptation, whereas the baseline methods update the model stepwise as each new observation becomes available.
    \item \texttt{LoRA} vs.\ \texttt{Full} vs.\ \texttt{Head FT}: compares \texttt{LoRA}, which remains underexplored for adaptive model-based control, against conventional head fine-tuning \textcolor{black}{and fine-tuning the entire model}. The same comparison is repeated within the \texttt{ADT} framework by swapping only the fine-tuning module while holding all other components fixed.
    \item With validation vs.\ without validation: isolates the contribution of the proposed online validation stage, which safeguards against deploying an update that degrades predictive performance.
\end{itemize}

\begin{figure}[h!]
    \centering
    \includegraphics[width=0.9\linewidth]{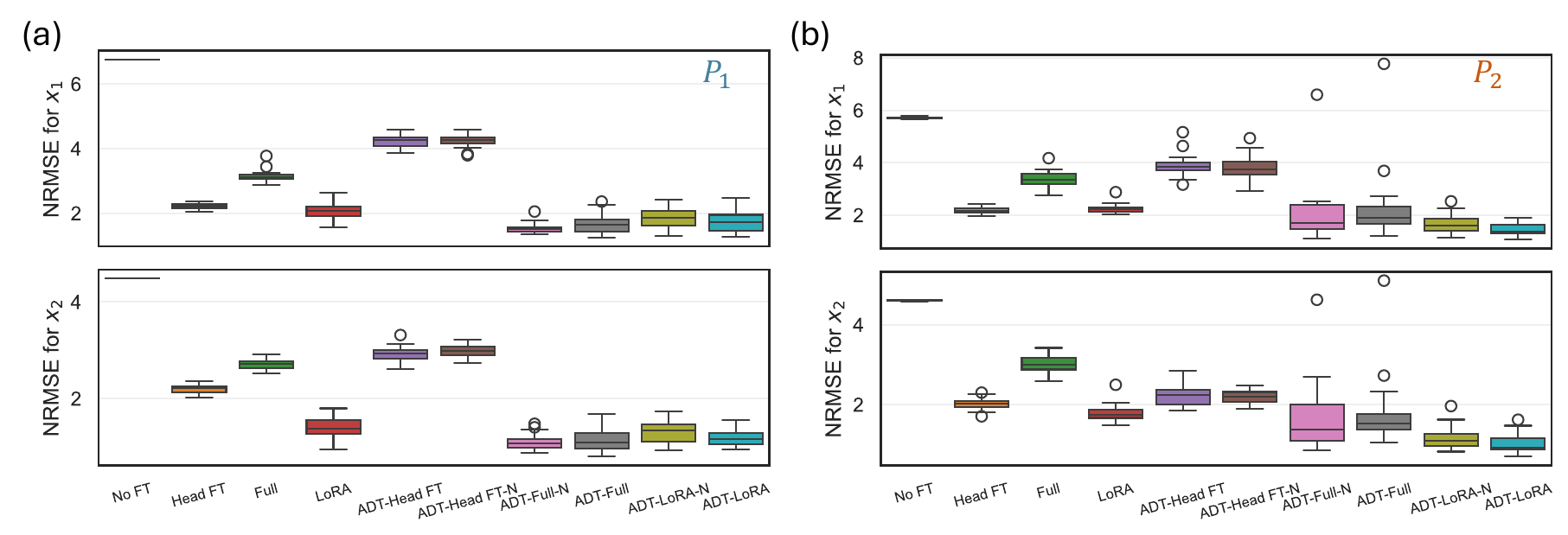}
    \caption{Distribution of the normalized root mean square error (NRMSE) across qualified replications for each benchmark method, evaluated for states $x_1$ (top) and $x_2$ (bottom) under (a) concept $P_1(y\mid x)$ and (b) concept $P_2(y\mid x)$. Lower values indicate more accurate median predictions.}
\label{fig:NRMSE}
\end{figure}

\begin{figure}[h!]
    \centering
    \includegraphics[width=0.9\linewidth]{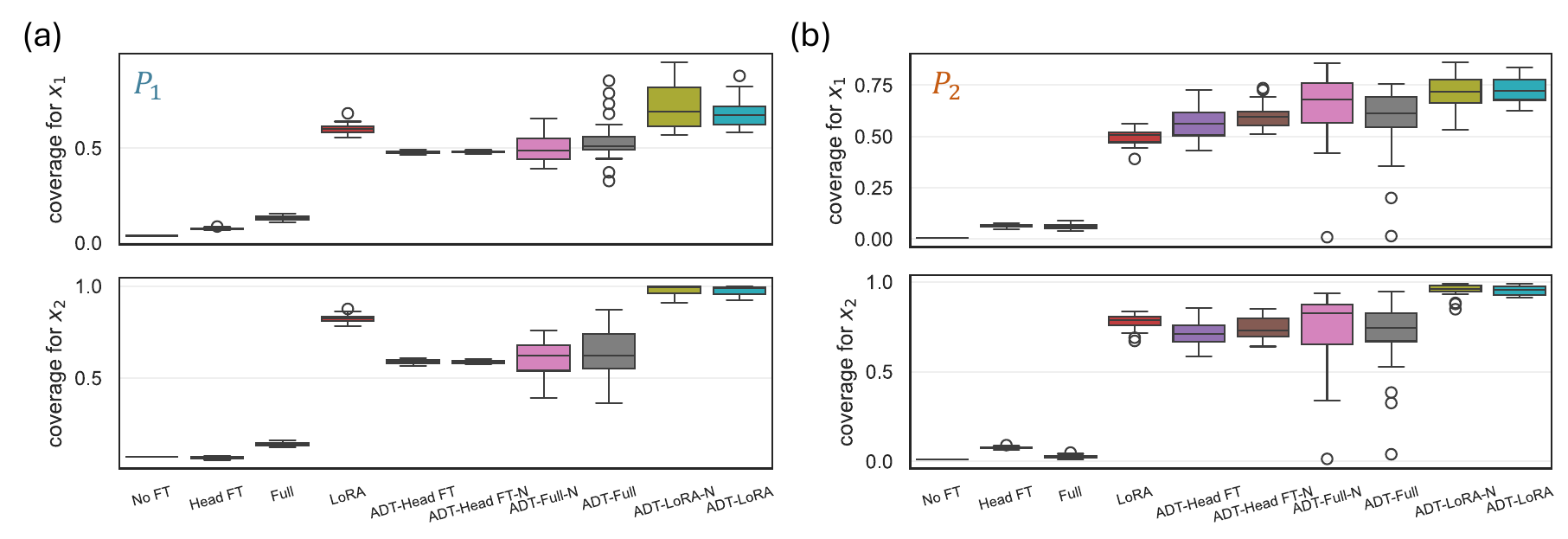}
    \caption{Empirical coverage of the $90\%$ prediction interval across qualified replications for each method, for states $x_1$ (top) and $x_2$ (bottom) under (a) $P_1(y\mid x)$ and (b) $P_2(y\mid x)$. \textcolor{black}{Higher coverage indicates more frequent inclusion of the nominal response but does not necessarily imply better probabilistic calibration.}}
\label{fig:Coverage}
\end{figure}

\begin{figure}[h!]
    \centering
    \includegraphics[width=0.9\linewidth]{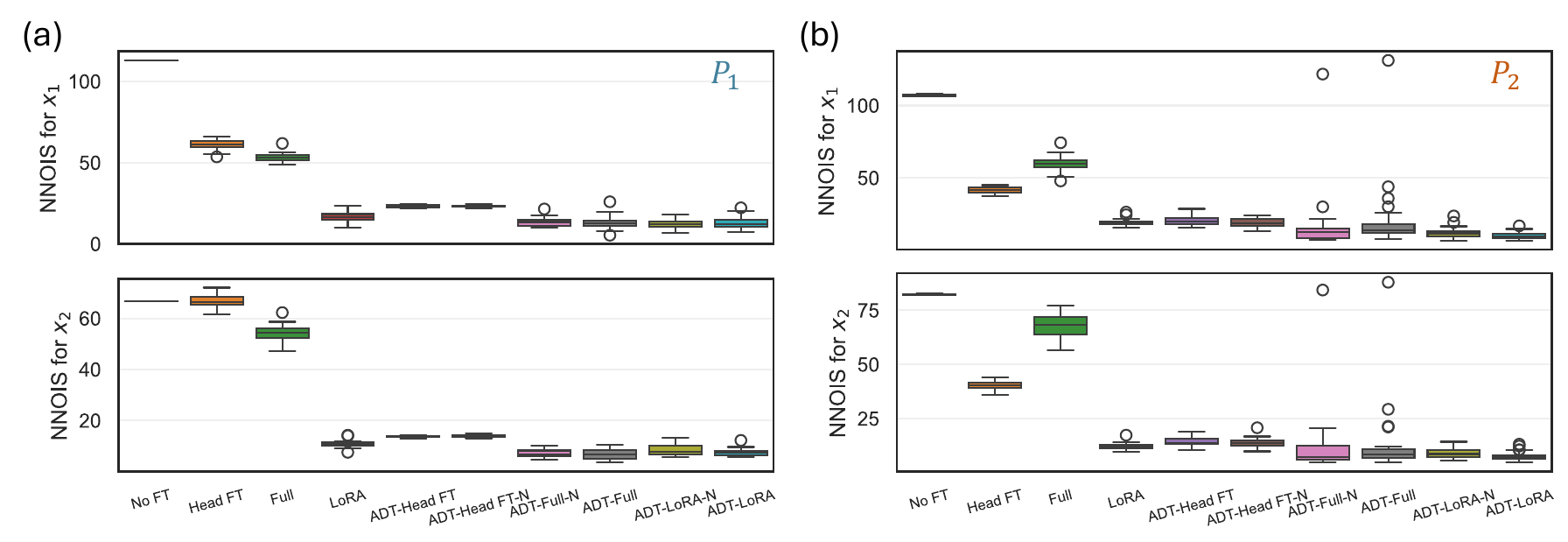}
    \caption{Distribution of the normalized negatively oriented interval score (NNOIS) across qualified replications for each method, for states $x_1$ (top) and $x_2$ (bottom) under (a) $P_1(y\mid x)$ and (b) $P_2(y\mid x)$. \textcolor{black}{Lower values indicate a more favorable trade-off between interval width and deviations of the nominal response outside the predicted intervals.}}
\label{fig:NNOIS}
\end{figure}

\begin{figure}[h!]
    \centering
    \includegraphics[width=0.9\linewidth]{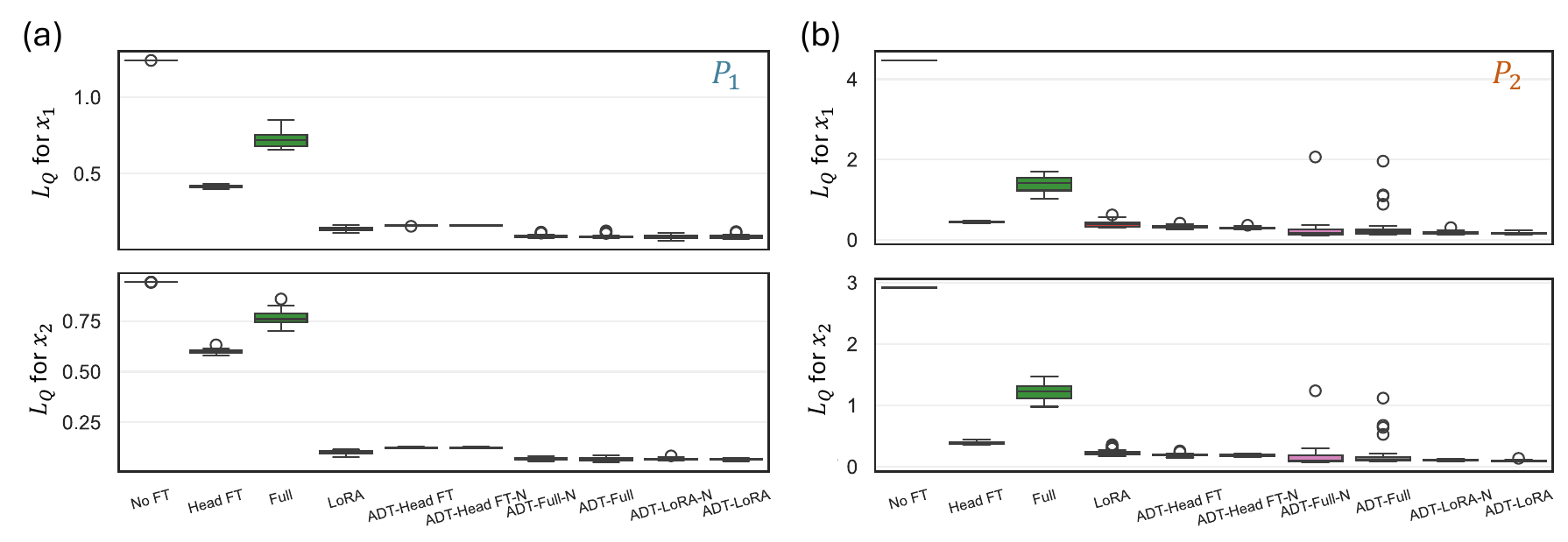}
    \caption{Distribution of the quantile loss $L_Q$ across qualified replications for each method, for states $x_1$ (top) and $x_2$ (bottom) under (a) $P_1(y\mid x)$ and (b) $P_2(y\mid x)$. Lower values indicate closer agreement between the predicted quantiles and the realized responses.}
\label{fig:quantile_loss}
\end{figure}

For each approach, 30 replication simulations are conducted with different random seeds, and the distributions of NRMSE, Coverage, NNOIS, and the quantile loss $L_Q$ of $x_1$ and $x_2$ under adaptation to $P_1$ and $P_2$ are compared. For the \texttt{ADT} approaches, \textcolor{black}{the evaluation metrics are computed only over the intervals between the deployment of an updated live model and the subsequent drift detection under each operating concept. This allows the predictive performance to be assessed during periods in which the model has passed the proposed validation procedure and is considered reliable under the current operating conditions. In contrast, stepwise adaptation approaches are evaluated over the entire operating period, as they do not incorporate an explicit mechanism for validating model updates. Replications in which drift remains undetected throughout a new operating concept are excluded from the predictive performance evaluation, but are accounted for separately in the subsequent analysis of Type I and Type II error rates. Specifically, only 18 and 21 replications of \texttt{ADT-Full-N} and \texttt{ADT-Full}, respectively, are included in this analysis, as the remaining 12 and 9 replications failed to detect the drift at the transition to $P_2$, as shown in Figure~\ref{fig:false_alarm}. Consequently, the reported predictive performance represents a conditional evaluation rather than a direct comparison of overall performance across the approaches.} The results across all metrics are summarized in Figures~\ref{fig:NRMSE}--\ref{fig:quantile_loss}.

{\color{black}

First, adaptation is essential for maintaining predictive accuracy under distributional shifts. \texttt{No FT} consistently exhibits the largest NRMSE, NNOIS, and quantile loss, while its empirical coverage approaches zero under both $P_1$ and $P_2$ (Figures~\ref{fig:NRMSE}--\ref{fig:quantile_loss}). These results demonstrate that an unadapted Digital Twin loses both predictive accuracy and uncertainty calibration as operating conditions change.

Second, comparing the approaches with and without the proposed \texttt{ADT} framework reveals improvements in several metrics, although the extent depends on the fine-tuning strategy. Compared with stepwise \texttt{Full}, \texttt{ADT-Full} substantially reduces NRMSE and quantile loss under both concepts. Meanwhile, stepwise \texttt{LoRA} already achieves competitive predictive accuracy, but \texttt{ADT-LoRA} provides higher empirical coverage and lower NNOIS, particularly for $x_2$ under $P_1$ and both states under $P_2$. The improvements in NRMSE are less pronounced under $P_1$ but become more apparent under $P_2$. These findings suggest that drift-triggered batch-wise adaptation can improve uncertainty estimation even when stepwise adaptation already provides reasonable predictive accuracy.

Third, the choice of fine-tuning strategy substantially influences predictive performance. Among stepwise approaches, \texttt{LoRA} consistently outperforms \texttt{Full} and \texttt{Head FT} across most metrics. Within the \texttt{ADT} framework, both LoRA and full-model fine-tuning achieve relatively low NRMSE and quantile loss, whereas head fine-tuning generally exhibits larger errors and lower coverage. Notably, \texttt{ADT-Full} achieves competitive predictive accuracy but exhibits greater variability and occasional extreme outliers under $P_2$. This motivates further investigation of the stability of different fine-tuning strategies through the subsequent analysis of Type I and Type II error rates.

Fourth, the differences between validated and nonvalidated \texttt{ADT} approaches are generally modest across the evaluated metrics. In particular, \texttt{ADT-LoRA} and \texttt{ADT-LoRA-N} exhibit similar NRMSE, coverage, NNOIS, and quantile loss, while validation does not consistently improve the performance of the other fine-tuning strategies. These results suggest that the primary role of online validation is not necessarily to improve predictive accuracy, but to provide a statistical safeguard against deploying updates without demonstrated improvement over the existing live model.

Overall, the results demonstrate that the proposed \texttt{ADT} framework can improve predictive accuracy and uncertainty estimation across different fine-tuning strategies, with LoRA providing the superior performance over the others under both operating concepts.

\subsection{Benchmarking on Run Performance Analysis}

To further evaluate the trade-offs between batch fine-tuning and the control performance due to the delayed model replacement, in this section we compare the tracking performance using all 30 replicates from each method. Figure~\ref{fig:traj_compare} compares the closed-loop trajectories of $x_1$ obtained via selected methods for the ease of visualization, with enlarged views under $P_1$ and $P_2$. As shown in Figure~\ref{fig:traj_compare}(a), most methods generally follow the reference trajectory, although noticeable differences emerge near abrupt transitions and constraint boundaries.

\begin{figure}[h!]
    \centering
    \includegraphics[width=0.8\linewidth]{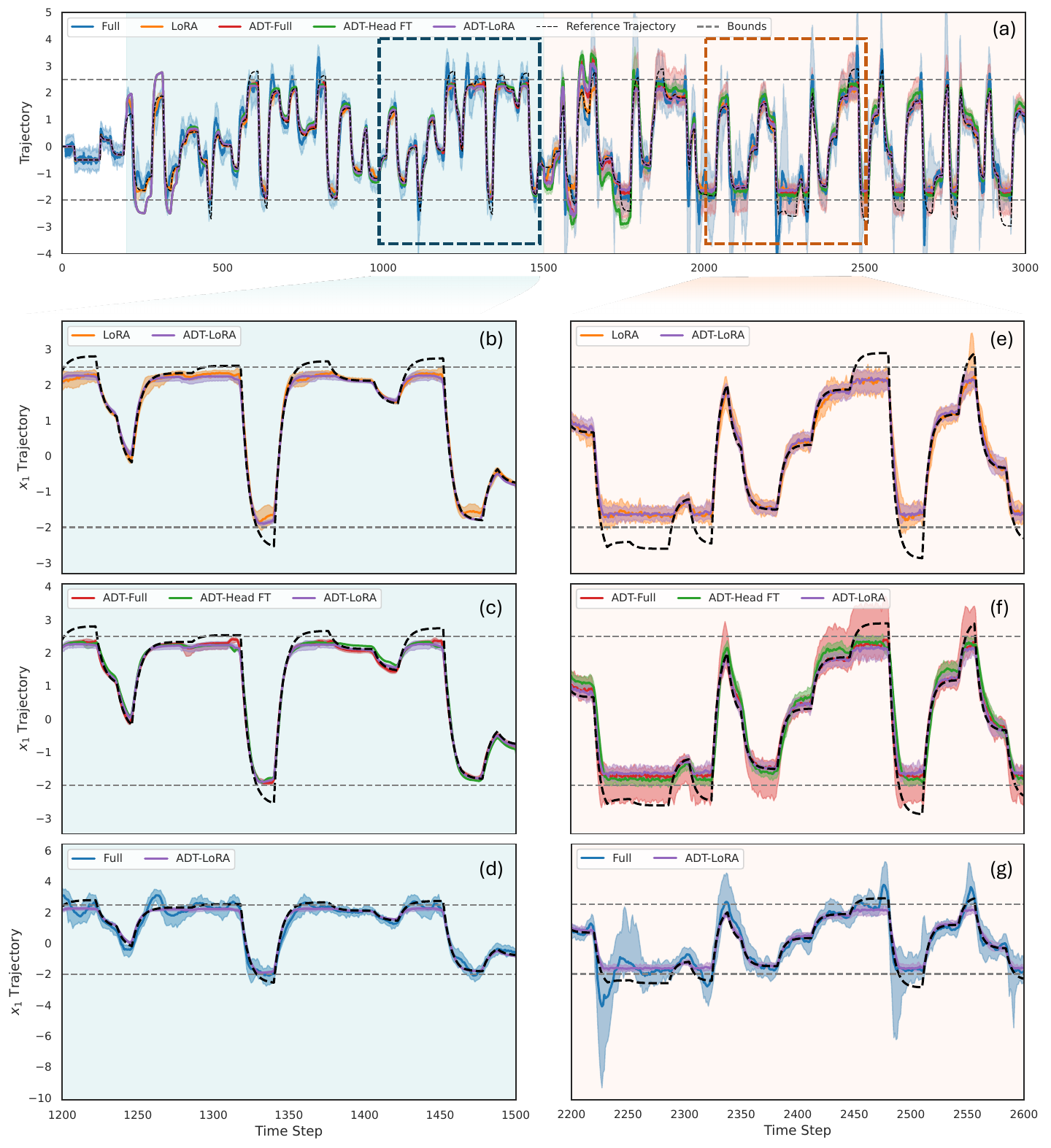}
    \caption{Comparison of closed-loop reference tracking and constraint satisfaction among different fine-tuning strategies. (a) Overall trajectories across the two operating conditions. Enlarged views of trajectory distributions under (b)-(d) the noiseless condition ($P_1$) and (e)-(g) the noisy condition ($P_2$). Panels (b) and (e) compare \texttt{LoRA} and \texttt{ADT-LoRA}, while panels (c) and (f) compare \texttt{ADT-Full}, \texttt{ADT-Head FT}, and \texttt{ADT-LoRA}. }
    \label{fig:traj_compare}
\end{figure}

Figures~\ref{fig:traj_compare}(b) and (e) compare the trajectory distributions obtained using \texttt{LoRA} and \texttt{ADT-LoRA} under $P_1$ and $P_2$, respectively. Under $P_1$, Figure~\ref{fig:traj_compare}(b) shows that \texttt{ADT-LoRA} produces more consistent trajectories across replications, whereas \texttt{LoRA} exhibits greater variability. Moreover, \texttt{LoRA} occasionally exhibits minor constraint violations and maintains a larger margin from the lower bound around $k=1340$. In contrast, \texttt{ADT-LoRA} approaches the constraint boundary more closely without violating it, while achieving improved reference tracking around $k=1340$ and $k=1470$. Similar observations can be made under $P_2$, as shown in Figure~\ref{fig:traj_compare}(e). Although the median trajectories of the two methods are nearly identical, \texttt{ADT-LoRA} demonstrates slightly better constraint handling, whereas \texttt{LoRA} occasionally exhibits minor constraint violations. Nevertheless, the differences in reference tracking performance between the two methods remain relatively modest.

Figures~\ref{fig:traj_compare}(c) and (f) further compare the trajectory distributions obtained using \texttt{ADT-Full}, \texttt{ADT-Head FT}, and \texttt{ADT-LoRA} to examine the influence of different fine-tuning strategies within the ADT framework. Under the noiseless conditions of $P_1$, Figure~\ref{fig:traj_compare}(c) shows that all three methods produce nearly identical trajectory distributions. In contrast, their differences become substantially more pronounced under the noisy conditions of $P_2$, as shown in Figure~\ref{fig:traj_compare}(f). Specifically, \texttt{ADT-Full} exhibits substantial constraint violations, whereas \texttt{ADT-Head FT} and \texttt{ADT-LoRA} demonstrate comparable constraint-handling capabilities. However, \texttt{ADT-LoRA} achieves better reference tracking than \texttt{ADT-Head FT}, as evidenced by the noticeable deviation between the reference trajectory and the median trajectory of \texttt{ADT-Head FT}. These observations further support the superior adaptation capability of \texttt{ADT-LoRA} under increased process noise, achieving a more favorable balance between constraint satisfaction and reference tracking than the other fine-tuning strategies.

Lastly, Figures~\ref{fig:traj_compare}(d) and (g) compare \texttt{Full} and \texttt{ADT-LoRA} to illustrate the combined benefits of drift-triggered adaptation, parameter-efficient fine-tuning, and model validation. While \texttt{Full} demonstrates reasonable reference tracking under $P_1$, it exhibits substantial variability and fails to consistently satisfy the constraints under both operating conditions. In contrast, \texttt{ADT-LoRA} maintains more consistent reference tracking across replications while demonstrating better constraint-handling capabilities under both $P_1$ and $P_2$. The contrast becomes particularly pronounced under the noisy conditions of $P_2$, where \texttt{Full} exhibits considerably larger trajectory deviations. These results highlight the advantages of integrating the proposed adaptation and validation mechanisms with parameter-efficient fine-tuning to improve the robustness and reliability of closed-loop MPC under changing operating conditions.

\begin{figure}[h!]
    \centering
    \includegraphics[width=1\linewidth]{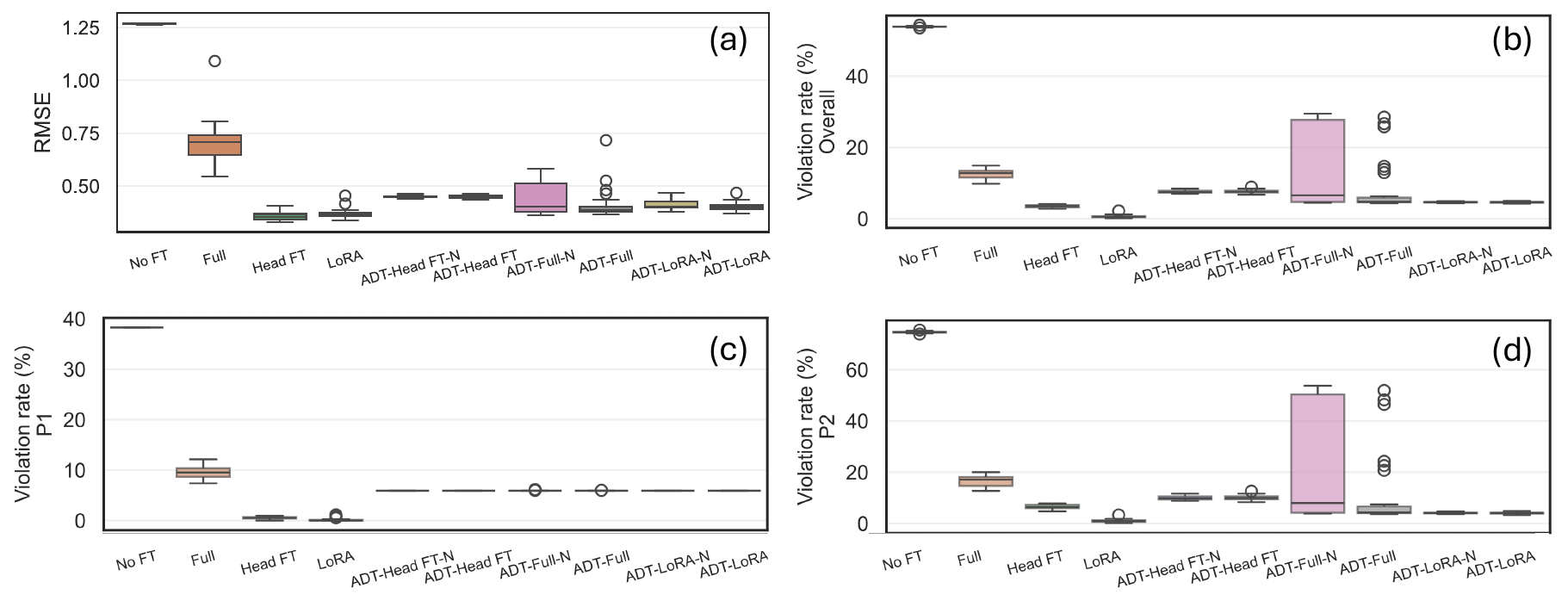}
    \caption{Comparison of closed-loop control performance across different fine-tuning strategies and ADT configurations. (a) RMSE between the resulting trajectories and the reference trajectory over the entire operation. Per-response constraint violation rates of $x_1$ under (b) the entire operation, (c) the noiseless condition ($P_1$), and (d) the noisy condition ($P_2$).}
    \label{fig:traj_RMSE_violation}
\end{figure}

Figure~\ref{fig:traj_RMSE_violation} quantitatively evaluates the closed-loop control performance across replications in terms of reference tracking RMSE and constraint violation rates of $x_1$. Specifically, Figure~\ref{fig:traj_RMSE_violation}(a) presents the RMSE over the entire trajectory, while Figures~\ref{fig:traj_RMSE_violation}(b)--(d) show the constraint violation rates over the entire operation and under $P_1$ and $P_2$, respectively. Among the evaluated methods, \texttt{No FT} exhibits the poorest overall performance, while \texttt{Full} yields the highest RMSE and constraint violation rates among the fine-tuning methods. These results further emphasize the importance of effective fine-tuning strategies for online model adaptation.

Among the \texttt{ADT}-based methods, \texttt{ADT-Full}, \texttt{ADT-LoRA-N}, and \texttt{ADT-LoRA} achieve comparable RMSE, whereas \texttt{ADT-Full-N} exhibits substantially greater variability across replications. Similar observations can be made from the constraint violation rates, where the differences among the \texttt{ADT}-based methods are relatively minor under $P_1$ but become more pronounced under the noisy conditions of $P_2$. In particular, \texttt{ADT-Full-N} exhibits considerable variability and occasionally high violation rates, whereas \texttt{ADT-Full} demonstrates more consistent performance. This comparison suggests that the proposed validation mechanism effectively safeguards against unsuccessful model updates, particularly when fine-tuning the entire network, thereby improving the stability of closed-loop performance. Moreover, \texttt{ADT-LoRA} achieves relatively low violation rates with limited variability, demonstrating more consistent constraint handling than \texttt{ADT-Full} and better performance than \texttt{ADT-Head FT} under $P_2$.

The stepwise \texttt{Head FT} and \texttt{LoRA} methods generally achieve lower RMSE and constraint violation rates than the \texttt{ADT}-based approaches. This can be partially attributed to the delayed model replacement inherent in the proposed \texttt{ADT} framework. Specifically, both metrics are evaluated over the entire operation, including the period between drift occurrence and model replacement, during which the existing live model continues operating while sufficient data are collected for batch fine-tuning and the updated idle model undergoes validation. The temporary model mismatch during this interval can introduce substantial tracking errors and constraint violations, increasing the overall RMSE and violation rates. Overall, these findings reveal an inherent trade-off between the responsiveness of continuous fine-tuning and the reliability afforded by drift-triggered batch adaptation and model validation.

\subsection{Benchmarking on Drift Detection Performance}

Finally, we compare the drift detection performance across the \texttt{ADT}-based methods, especially after the drift detector is re-calibrated online. Since the same model is used for each method for offline drift detector setup by monitoring the gradients of the parameters in the last layer, the offline calibration of the drift detector (UCL, $\mathbf{\bar{s}}, \hat{\Sigma}$) is also identical, resulting in the same delay in drift detection when transitioning from $P_0$ to $P_1$, i.e., the alarm is raised at $k=207$ while the drift is injected at $k=200$. Therefore, more meaningful comparisons would be to look at whether the drift detector can be effectively reset online.

\begin{figure}[h!]
    \centering
    \includegraphics[width=0.9\linewidth]{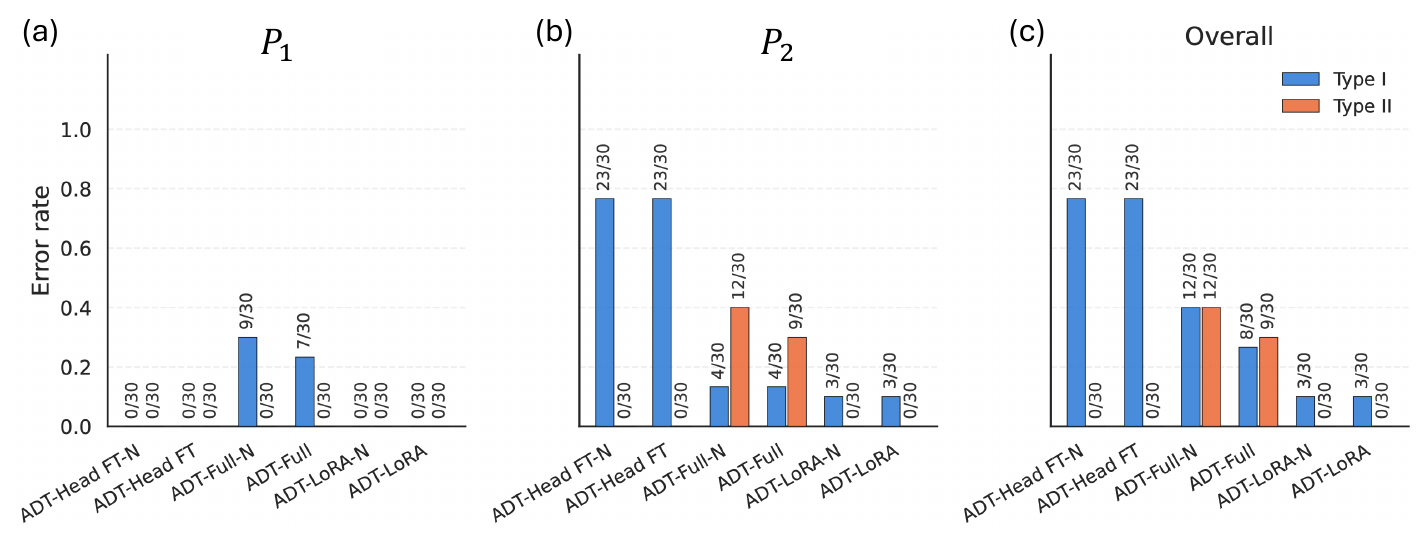}
    \caption{Comparison of Type I (false alarm) and Type II (missed detection) error rates across different \texttt{ADT}-based methods over 30 replicates.}
    \label{fig:false_alarm}
\end{figure}

\begin{figure}[h!]
    \centering
    \includegraphics[width=0.8\linewidth]{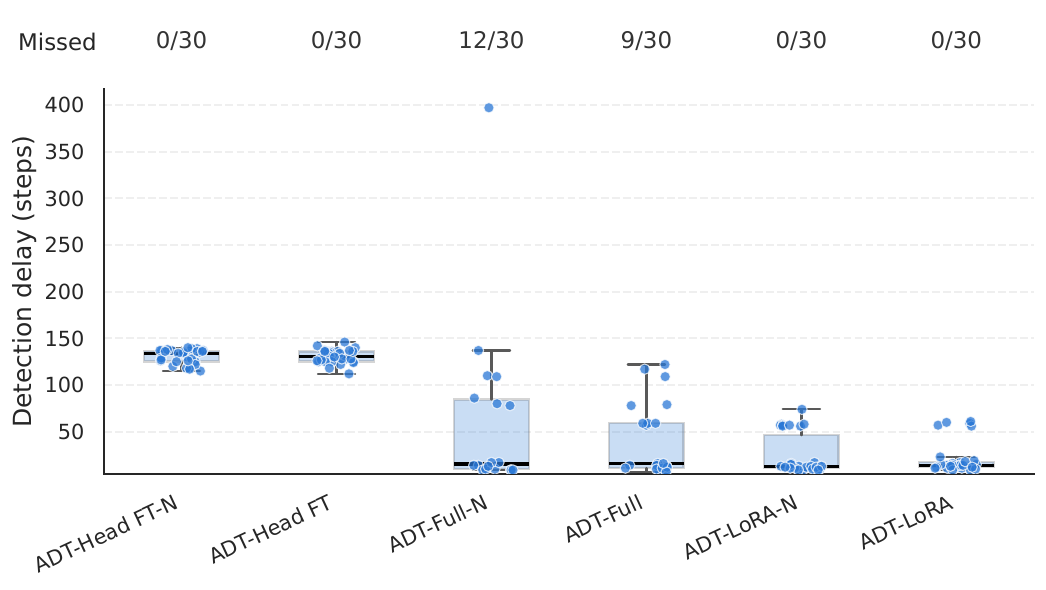}
    \caption{Comparison of drift detection delays under $P_2$ across different \texttt{ADT}-based methods over 30 replicates. Detection delay is measured as the number of time steps between the onset of drift and its successful detection.}
    \label{fig:delay_average}
\end{figure}

Figures~\ref{fig:false_alarm} and~\ref{fig:delay_average} further evaluate the performance of the proposed drift detection framework in terms of Type I and Type II error rates and detection delays across 30 replications. It should be noted that the drift detector is recalibrated using the updated model following each successful adaptation, such that the reported false alarms under $P_1$ and $P_2$ occur after detector recalibration under the corresponding operating conditions. As shown in Figure~\ref{fig:false_alarm}, the error patterns differ considerably across the fine-tuning strategies and operating conditions. Under the noiseless condition of $P_1$, \texttt{ADT-Full} and \texttt{ADT-Full-N} exhibit false alarms in 7 and 9 out of 30 replications, respectively, whereas the other configurations exhibit no false alarms. In contrast, under the noisy condition of $P_2$, \texttt{ADT-Head FT} and \texttt{ADT-Head FT-N} exhibit substantially higher false alarm incidences, both reaching 23 out of 30 replications. Moreover, \texttt{ADT-Full} and \texttt{ADT-Full-N} exhibit missed detections at the transition to $P_2$ in 9 and 12 replications, respectively. In comparison, \texttt{ADT-LoRA} and \texttt{ADT-LoRA-N} demonstrate the most consistent detection performance, with only three replications exhibiting false alarms under $P_2$ and no missed detections under either operating condition. Figure~\ref{fig:delay_average} further shows that the LoRA-based configurations achieve relatively short detection delays, whereas the head fine-tuning configurations exhibit substantially longer delays.

These differences suggest that the fine-tuning strategy may influence not only predictive performance but also the gradient statistics used for subsequent drift detection. Under $P_1$, the elevated false alarm incidence of \texttt{ADT-Full} may result from overfitting or insufficient generalization, causing gradient variations that are not adequately captured during detector recalibration. Such variability may also reduce the distinguishability of the subsequent drift to $P_2$, contributing to missed detections. In contrast, although \texttt{ADT-Head FT} exhibits no false alarms under $P_1$, it experiences substantially longer detection delays at the transition to $P_2$. One possible explanation is its relatively high estimated UCL, which may reflect greater variability in the monitored gradient statistics under $P_1$, making drift-induced changes less distinguishable from normal fluctuations and requiring more time for the monitoring statistic to exceed the threshold. Furthermore, the limited adaptation flexibility of \texttt{ADT-Head FT} may introduce residual model mismatch under noisy $P_2$, resulting in gradient variations that are not adequately characterized during recalibration and consequently frequent false alarms. In comparison, \texttt{ADT-LoRA} may achieve a better balance between adaptation flexibility and parameter stability, maintaining more consistent gradient behavior while preserving sensitivity to subsequent concept drift.

Nevertheless, these explanations remain hypotheses, and the relationships among model fidelity, gradient distributions, UCL estimation, and detection reliability require further investigation beyond the scope of this work. Moreover, limited data availability during online operation makes accurate estimation of the extreme-tail UCL challenging, while collecting additional data would delay model adaptation. Therefore, the proposed recalibration procedure represents a practical compromise between statistical accuracy and the need for continuous, lifelong model monitoring and adaptation. The estimated UCL thus serves as a statistically motivated reference rather than a strict guarantee of the nominal false alarm rate.

Overall, the benchmarks support \texttt{LoRA} as an effective online fine-tuning strategy, but may risk the stability during step-wise fine-tuning. On the other hand, \texttt{ADT-LoRA} offers improved model prediction during the evaluated post-adaptation intervals and favorable drift-detection performance among the \texttt{ADT} configurations. However, delayed model replacement can increase tracking errors and constraint violations, highlighting a trade-off between adaptation responsiveness and the safeguards provided by online validation. Motivated by these findings, the subsequent engineering examples focus exclusively on stepwise \texttt{LoRA} and \texttt{ADT-LoRA}.

}

\section{Engineering Case Study on Additive Manufacturing}
\label{sec:AM}

In this section, we apply the framework to DED as a case study, building on the testbed and dataset of \cite{chen2025real}, where the melt pool is a complex, highly nonlinear temporal system. Drift is injected via temperature-dependent material properties (e.g.\ thermal conductivity, specific heat), representing batch-to-batch feedstock variation. Unlike the explicit injection of the illustrative example, here the drift is embedded in the governing equations, producing nonlinear, non-additive effects on the dynamics and giving a more realistic and demanding test. \textcolor{black}{We begin by briefly summarizing the experimental setup, surrogate modeling, and robust MPC formulation from our previous work \cite{chen2025real, chen2026uncertainty} in n Section~\ref{sec:sys_overview}-\ref{sec:virtual_sys}, and then the validation setup and results.} The reader is referred to \cite{chen2025real} for full DED Digital Twin details.

\subsection{System Overview and Scope}
\label{sec:sys_overview}
The Digital Twin is designed for controlling melt pool dynamics, the most critical and controllable region of DED, where thermal history governs defects (porosity, lack of fusion, residual stress). The state is the melt pool temperature and depth, $\mathbf{x}_k=[x_{\text{temp},k},x_{\text{depth},k}]^\top$, assumed observable via in-situ sensing, and the control input is the laser power $u_k$.

The twin is a closed loop of (i) a physical system producing state observations, (ii) a virtual system producing multi-step predictions conditioned on candidate inputs, and (iii) an MPC computing the optimal action applied back to the plant. To maintain computational tractability for real-time decision-making, the Digital Twin is constructed at the melt pool level rather than the full part scale, reducing the dimensionality of the system while preserving the dominant dynamics relevant to control.

Detailed setup, data acquisition, and feature extraction are given in \cite{chen2025real, chen2026uncertainty}. Here we focus on integrating the DED system with the proposed framework, continuously monitoring and updating the surrogate under evolving conditions.

\subsection{Physical System: GAMMA Simulation of DED Process}

\begin{figure}[ht]
  \centering
  \begin{minipage}{0.48\linewidth}  
    \centering
    \includegraphics[width=0.7\linewidth]{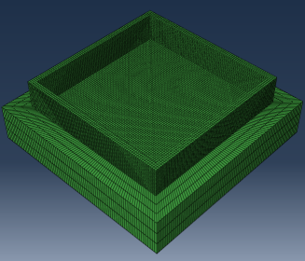}
    \caption{Single-track square}
    \label{fig:printed_square}
  \end{minipage}%
  \hfill
  \begin{minipage}{0.5\linewidth}  
    \centering
    \captionsetup{type=table}  
    \caption{Specification of the printed square} 
    \label{tab:square_spec}
    \footnotesize
    \begin{tabular}{c|c}
    \hline \hline
    Item             & Quantity  \\ \hline
    Side length      & 40 mm     \\
    Track width      & 1.5 mm    \\
    Layer height     & 0.75 mm   \\
    Num. of layers   & 10 layers \\
    Element size     & 0.375mm   \\
    Num. of elements & 40540     \\
    Substrate height & 10 mm       \\
    Scanning speed   & 7 mm/sec  \\ \hline \hline
    
    \end{tabular}
  \end{minipage}
\end{figure}

The physical DED process is represented by GAMMA, an in-house GPU-accelerated explicit FEA solver \cite{liao2023efficient} performing part-scale transient heat-transfer simulation, providing a controlled environment \cite{chen2025real} for generating trajectories and injecting uncertainties for virtual experiments. A single-track square geometry is selected as the target structure, illustrated in Figure~\ref{fig:printed_square}, with detailed specifications provided in Table~\ref{tab:square_spec}. Here, an AISI 1018 substrate is used, and 316L stainless steel serves as the printing material. This setup enables efficient simulation of complex thermal dynamics while preserving sufficient fidelity in capturing melt pool behavior relevant to control. The single-track structure can also be used for future experimental verification of melt pool depths. For comprehensive details regarding the DED setup, feature extraction, and data-processing pipeline, the reader is referred to our previous work~\cite{chen2025real}.

\subsection{Virtual System: Surrogate Modeling via TiDE}
\label{sec:virtual_sys}
To construct a sufficiently rich set of laser power trajectories, a time-series design of experiments is adopted to systematically explore the input space. Following the approach of \cite{karkaria2024towards}, each laser power profile is parameterized in a reduced-dimensional representation using a Fourier basis, resulting in a 10-dimensional coefficient space. Latin Hypercube Sampling is then employed to generate diverse trajectories within this space, enabling broad coverage of admissible process variations. Additional implementation details can be found in prior studies \cite{karkaria2024towards, chen2025real}. Based on this strategy, 100 distinct simulations are performed using the finite element model. At each timestep, key thermal responses, including melt pool temperature and depth, are recorded. The resulting dataset spans a wide range of operating conditions and provides sufficient variability for learning the underlying process--response mapping.

Using this dataset, a TiDE is trained as a multivariate, multi-step predictor. The model is configured with an input window size of $w = 50$ and a prediction horizon of $N = 50$, and predicts the 0.05 and 0.95 quantiles that define the lower and upper bounds, respectively. Detailed training procedures and hyperparameter settings are provided in \cite{chen2025real}. The trained model achieves a MAPE of 1.29\% and an RRMSE of 0.054 for melt pool temperature, and 4.25\% and 0.0441, respectively, for melt pool depth.

The dominant source of aleatoric uncertainty in the generated data arises from numerical artifacts in the FEA. Specifically, the mismatch between the laser traversal distance per timestep and the mesh discretization produces inconsistent thermal exposure across elements, leading to significant fluctuations in temperature and melt pool geometry that can exceed those caused by external disturbances such as material variability. Although these fluctuations are reproducible under identical simulation settings, they cannot be readily reduced through additional sampling. The TiDE model addresses this by extracting dominant temporal patterns through its dense encoder to generate a smoothed median prediction, while capturing the remaining variability through learned quantile bounds.

\subsection{Decision-Making: Robust MPC Formulation}

The deployment of MPC in DED aims to enable anticipatory regulation of the thermal process so that porosity defects can be mitigated even under arbitrarily specified melt pool temperature references. Prior studies suggest that maintaining the melt pool depth within a dilution window of 10\% to 30\% is essential for preventing both interlayer and intralayer porosity \cite{dass2019state}. Identically to our formulation in \cite{chen2026uncertainty}, the robust MPC is formulated to track the desired temperature trajectory while enforcing chance constraints on melt pool depth:

\begin{subequations}
\label{eq:DED_MPC}
\begin{align}
\min\limits_{\mathbf{u}} & \sum_{i=0}^{N-1} \left( \left|\hat{x}_{\text{temp},k+i+1} - r_{\text{temp},k+i+1}\right|_{\mathbf{Q}}^2 + \left|\Delta u_{k+i}\right|_{\mathbf{R}}^2 \right) \\
\text{s.t.} & \mathbb{P}\left(\hat{x}_{\text{depth},k+i} \geq x_{\text{depth,lb}}\right) \geq 0.95, \quad \forall i \in \mathbb{N}{[1,N]}, \label{ep:mpc_ded_ub}\\
& \mathbb{P}\left(\hat{x}_{\text{depth},k+i} \leq x_{\text{depth,ub}}\right) \geq 0.95, \quad \forall i \in \mathbb{N}{[1,N]}, \label{ep:mpc_ded_lb}\\
& [\hat{\mathbf{x}}_{\text{temp},k+1}^{f}, \hat{\mathbf{x}}_{\text{depth},k+1}^{f}]^T = \text{TiDE}(\mathbf{x}_{\text{temp},k}^{p}, \mathbf{x}_{\text{depth},k}^{p}, \mathbf{d}_{x,k}^{p:f}, \mathbf{d}_{y,k}^{p:f}, \mathbf{z}_k^{p:f}, \mathbf{u}_k^{p:f}), \\
& u_{k+i} \in \mathbb{U} := \left\{ u \in \mathbb{R} \mid 504~\text{W} \leq u_{k+i} \leq 750~\text{W} \right\}
\end{align}
\end{subequations}

In this formulation, $\Delta u_{k+i}$ penalizes variations in consecutive control inputs to encourage smooth laser power modulation. The auxiliary variables $\mathbf{d}_x$ and $\mathbf{d}_y$ denote the distances from the laser position to the nearest geometric boundaries along the $x$- and $y$-directions, while $\mathbf{z}$ represents the nozzle height. These geometry-dependent quantities are incorporated as exogenous inputs to the TiDE model, improving its ability to capture spatially varying process dynamics. Unlike conservative robust MPC formulations that tighten the input constraints, no such modification is applied here, as the admissible input range remains inactive throughout the operating regime considered. Note that the constraint is enforced only from the fourth layer onward, as the melt pool temperature and depth are lower in earlier layers due to the boundary conditions imposed by the substrate.

\subsection{Phase~I: Drift Detector Setup}

As before, the pretrained TiDE (796,594 parameters) is augmented with a linear head and LoRA layers. Here, we fine-tune only the LoRA layers (7,476 parameters with $r=1$) for parameter efficiency. \textcolor{black}{The value of $r$ can be adjusted according to specific application needs. In this work, however, we fix $r=1$ to limit the number of tunable parameters, as increasing $r$ would introduce additional parameters that must be adapted online. A systematic investigation of the trade-off between the number of tunable parameters and the amount of data required for online adaptation is beyond the scope of this work.}

\begin{figure}[h!]
    \centering
    \includegraphics[width=1\linewidth]{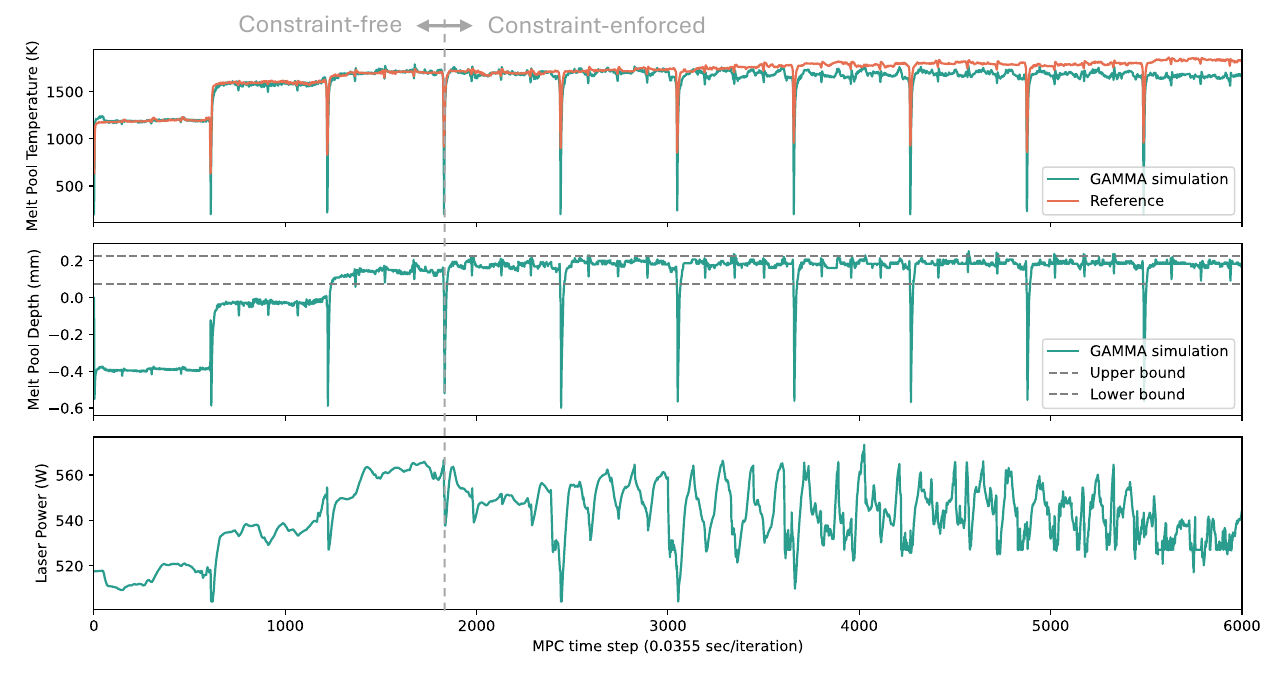}
    \caption{Trajectory of RMPC in the in-control condition for setting up the drift detector in Phase~I. The RMPC framework effectively trades off reference-tracking performance to maintain constraint satisfaction in the presence of fluctuations, enabled by the safety margins provided by the learned quantiles.}
    \label{fig:DED_phase_1_traj}
\end{figure}

\begin{figure}[h!]
    \centering
    \includegraphics[width=1\linewidth]{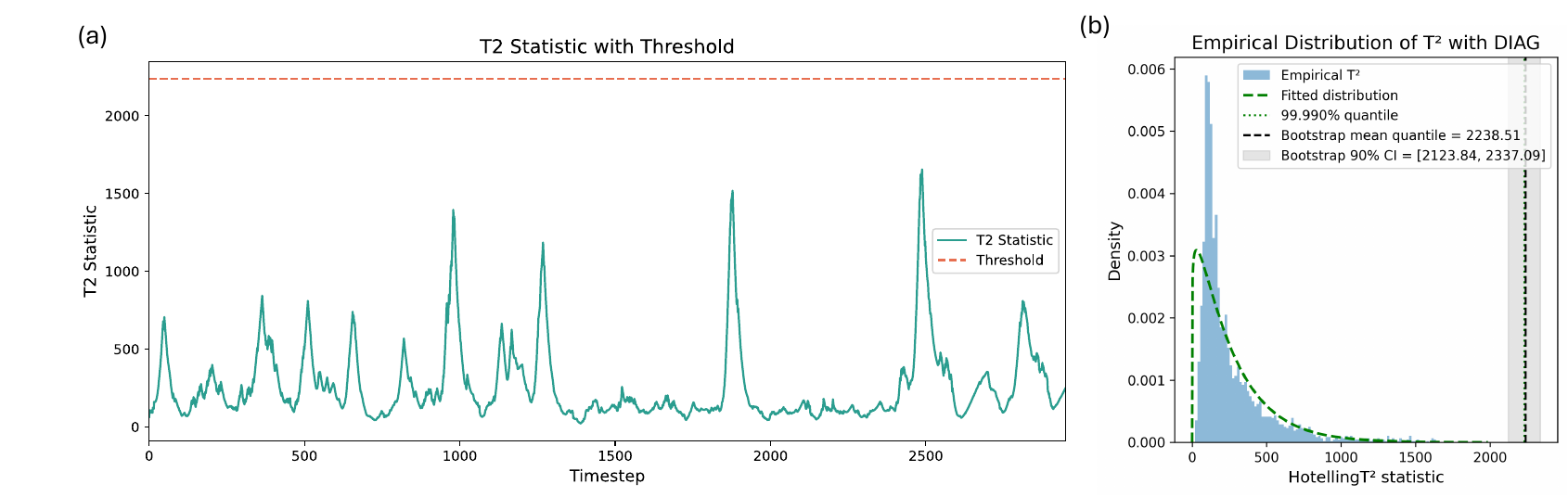}
    \caption{Hotelling $T^2$ statistics under in-control conditions for \textcolor{black}{estimated UCL}. (a) The $T^2$ trajectory of the streaming data. (b) Empirical distribution of $T^2$ in Phase~I, well approximated by the fitted scaled noncentral $\chi^2$ distribution.}
    \label{fig:DED_phase_1_T2}
\end{figure}

To establish the reference distribution of the MEWMA statistic under in-control conditions, a separate dataset of size 6,000 is generated by running GAMMA under robust MPC with a temperature reference trajectory distinct from that used in online testing. The corresponding trajectories are shown in Figure~\ref{fig:DED_phase_1_traj}. The first 3,000 samples are used to estimate the in-control score mean $\bar{\mathbf{s}}$ and covariance matrix $\hat{\Sigma}$, where $\hat{\Sigma}^{-1}$ is approximated using only the inverse of its diagonal elements. The remaining 3,000 samples are then used to sequentially compute the Hotelling $T^2$ statistic via MEWMA with $\lambda = 0.02$, based on the estimated $\bar{\mathbf{s}}$ and $\hat{\Sigma}^{-1}$. The empirical distribution of $T^2$ is subsequently approximated, and the detection threshold $h$ is determined from the fitted distribution with bootstrapping at significance level $\alpha_{\text{UCL}} = 10^{-4}$. The resulting $T^2$ trajectory and its empirical distribution are shown in Figure~\ref{fig:DED_phase_1_T2}(a)--(b). The $T^2$ statistic shows periodic peaks (Figure~\ref{fig:DED_phase_1_T2}(a)) induced by layer transitions, where prediction error rises; these inherent variations must be reflected in $\bar{\mathbf{s}}$, $\hat{\Sigma}$, and the threshold to avoid excessive false alarms at layer switches.

\subsection{Drift Injection on Material Properties}

The primary source of drift in the DED process arises from uncertainties in material properties. To evaluate the proposed framework under such conditions, variations are introduced into key material parameters, including density, solidus and liquidus temperatures, latent heat, and temperature-dependent thermal conductivity and specific heat capacity. Since 316L stainless steel is the printing material, the drift is applied to its properties to induce changes in the underlying system dynamics.

\begin{figure}[h!]
    \centering
    \includegraphics[width=0.8\linewidth]{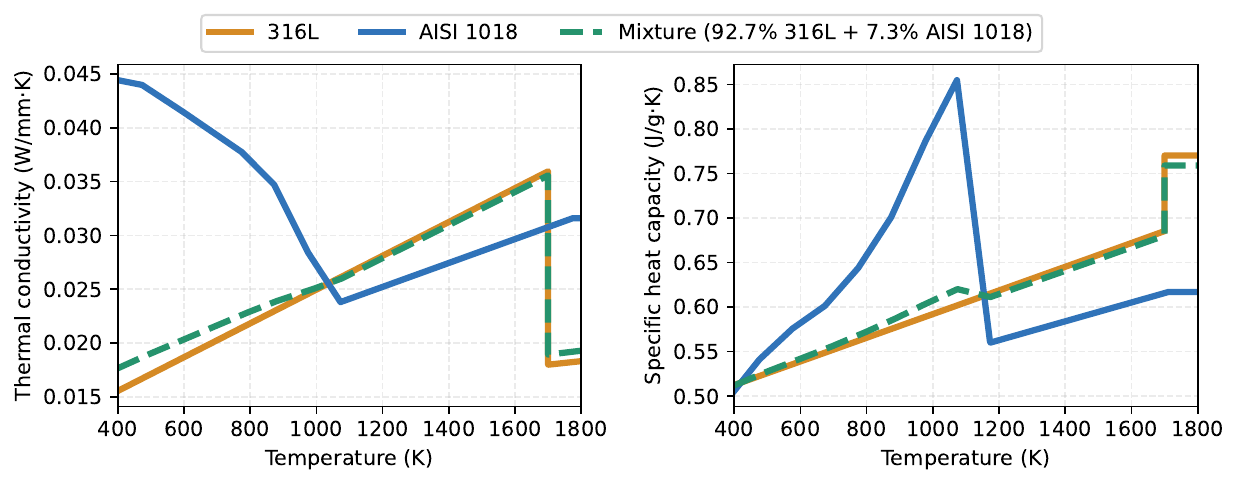}
    \caption{Temperature-dependent material properties for stainless steel 316L, AISI 1018, and their mixture. (a) Temperature-dependent thermal conductivity. (b) Temperature-dependent specific heat capacity. The discontinuities in the curves are intentionally introduced to calibrate the GAMMA simulation against experimental results.}
    \label{fig:mat_properties}
\end{figure}

To realize this drift, a virtual mixture of 316L stainless steel and AISI 1018 is considered, where the effective material properties are computed as a weighted average of the two constituents. The mixing process is initiated at time step 2000, during which the weight of 316L is \textcolor{black}{relatively reduced by 1.5\% and replaced by AISI 1018 every 40 steps.} This reduction continues until time step 2200, reaching a total decrease of \textcolor{black}{7.3\%}, after which the composition remains constant. The resulting static material properties of the mixture are summarized in Table~\ref{tab:mat_properties}, while the temperature-dependent thermal conductivity and specific heat capacity of both materials and their mixture are presented in Figure~\ref{fig:mat_properties}. 

The discontinuities observed in the curves are intentionally introduced to calibrate the GAMMA simulation against experimental results. This adjustment is necessary because the GAMMA simulation accounts only for heat conduction, without incorporating phase change or fluid-flow effects, leading to a systematic overestimation of the melt pool temperature. It is important to note that this induced drift affects not only the material near the melt pool but the entire simulation domain, resulting in more pronounced changes than those observed experimentally. The purpose of this setup is not to replicate a specific physical scenario but to introduce a substantial, non-additive, and highly nonlinear perturbation to the system dynamics for rigorous evaluation of the adaptive framework.

\begin{table}[h!]
\centering
\caption{Temperature-independent material properties}
    \small
    {\color{black}
\begin{tabular}{ccccc}
\hline \hline
Material                            & Density (g/mm$^3$) & Solidus temp. (K) & Liquidus temp. (K) & Latent heat (J/g) \\ \hline
SS 316L                             & 0.00798             & 1648.15           & 1673.15            & 290               \\
AISI 1018                           & 0.00787             & 1693              & 1733               & 270               \\
Mixture (92.7\% 316L, 7.3\% AISI 1018) & 0.00797             & 1651.424         & 1677.519          & 288.54               \\ \hline \hline
\end{tabular}}
\label{tab:mat_properties}
\end{table}

\subsection{Phase~II: Online Drift Detection, Adaptation, and Validation}

Leveraging $\bar{\mathbf{s}}$ and $\hat{\Sigma}$ estimated from Phase~I, the drift detector is deployed for online monitoring. As in the previous case, to reduce detection latency, only the quantile loss of the one-step-ahead prediction is used to represent the loss of the model. Once drift is detected, a data buffer of size 300 is collected for model fine-tuning. The fine-tuning procedure and associated hyperparameters remain identical to those used in the illustrative example. After successful fine-tuning, an additional 50 data points are collected during the validation stage to compare the quantile-loss distributions of the live and idle models. If the idle model passes the hypothesis test, it replaces the live model online, and the drift detector is reset. For the online reinitialization of Phase~I, 200 data points are first collected to estimate $\bar{\mathbf{s}}$ and $\hat{\Sigma}$, followed by an additional 500 data points to estimate the distribution of the $T^2$ statistic and its estimated UCL.

\subsubsection{Unconstrained Model Predictive Control}

\begin{figure}[h!]
    \centering
    \includegraphics[width=0.9\linewidth]{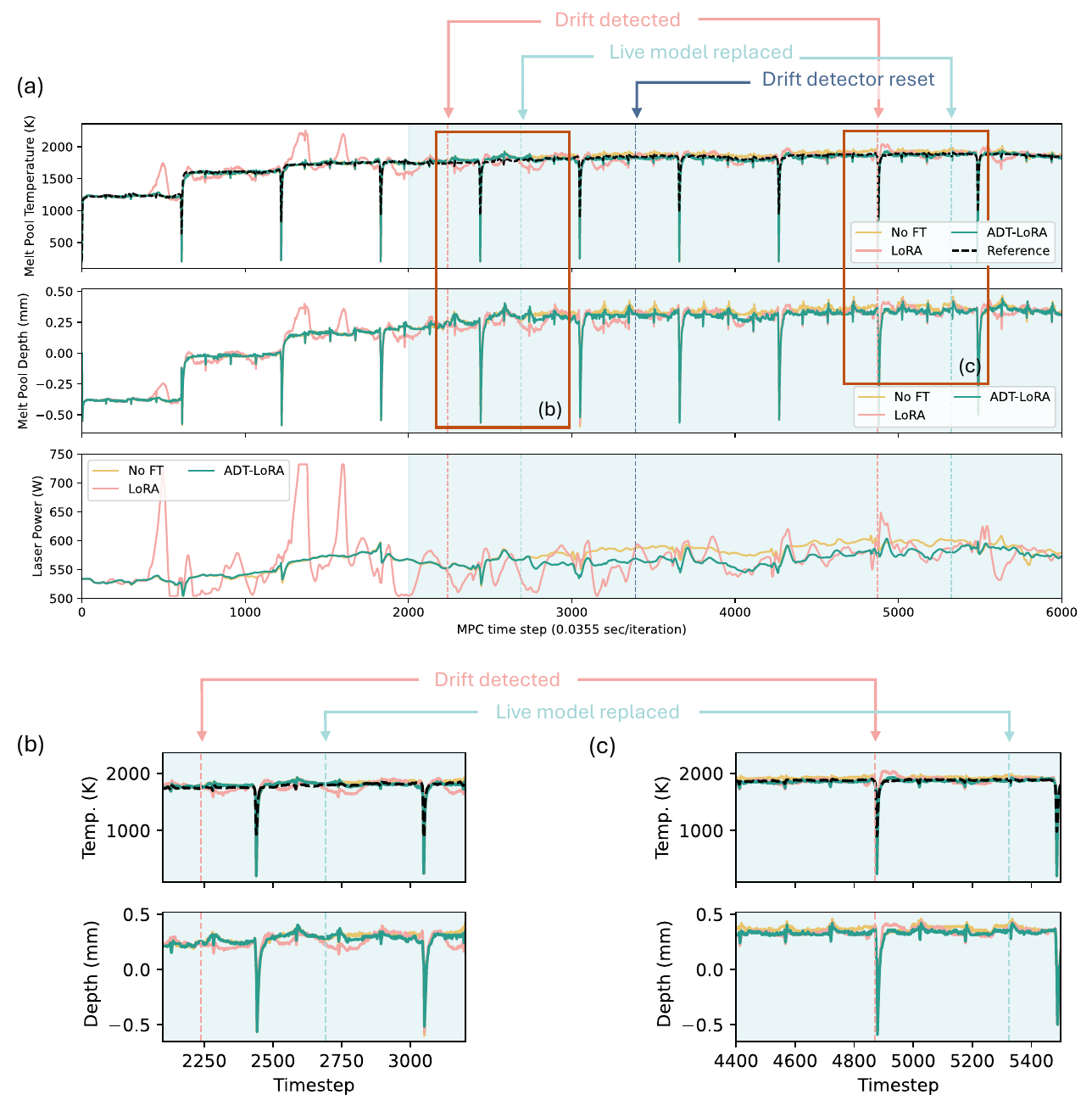}
    \caption{Resulting trajectories of melt pool temperature, melt pool depth, and laser power input in the DED printing process. (a) shows the complete trajectories over all 10 layers, while (b) and (c) highlight regions where drift is detected, and model updates are performed. The time steps corresponding to drift detection, live-model replacement, and completion of the drift-detector reset are indicated by dashed lines. The shaded background indicates the time steps operated under drifted material properties.}
    \label{fig:DED_traj}
\end{figure}

To begin with, we demonstrate the proposed method in unconstrained MPC, treating the process as a reference tracking problem by ignoring the melt pool depth constraints in Equation~\ref{eq:DED_MPC}. Figure~\ref{fig:DED_traj} presents the trajectories of melt pool temperature, melt pool depth, and applied laser power throughout the drift-injected DED process simulated in GAMMA. The results of \texttt{ADT-LoRA} are shown, along with \texttt{No-FT} and \texttt{LoRA} as benchmarks. As shown in Figure~\ref{fig:DED_traj}(a), drift is first detected 239 time steps after the material perturbation is introduced. Since the material drift is imposed gradually and the effect of the graded material requires several time steps to become significant in the simulation, the detector remains effective, identifying drift before the discrepancy between the temperature trajectory and the reference becomes pronounced (around time step 2,250). Although the comparison between the reference and state trajectories does not directly quantify prediction error, it reflects the degradation in control performance caused by an inaccurate predictive model within MPC. Figures~\ref{fig:DED_traj}(b) and~\ref{fig:DED_traj}(c) highlight the region where drift is detected, and the live model is subsequently replaced. As observed in these subfigures, the reference-tracking performance improves significantly after the model update, demonstrating the capability of the proposed framework to maintain decision quality through effective online model adaptation.

\begin{figure}[h!]
    \centering
    \includegraphics[width=0.9\linewidth]{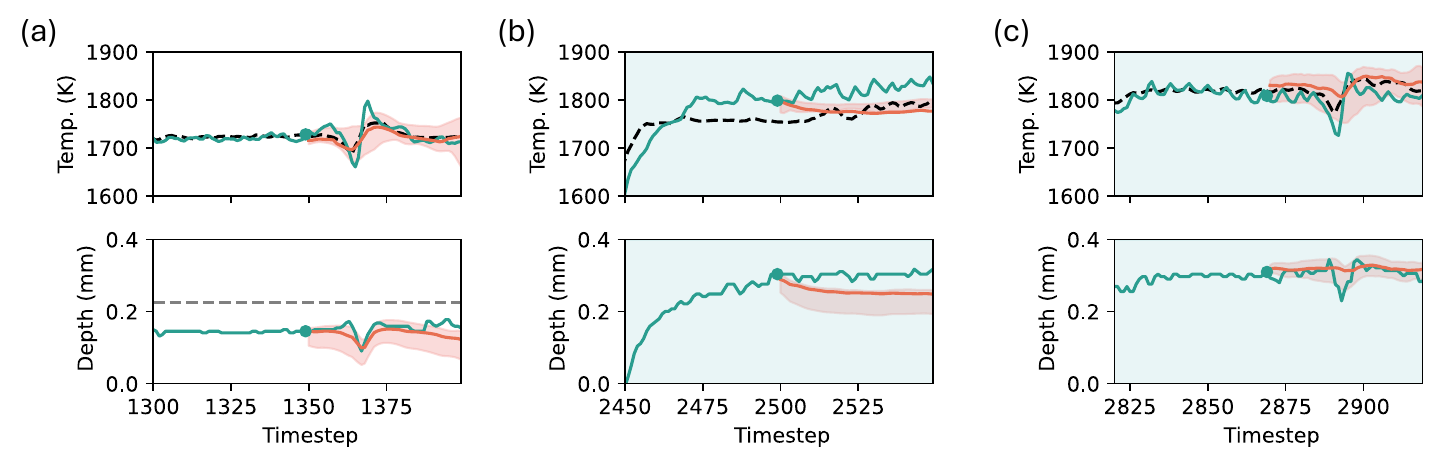}
    \caption{TiDE-predicted median and quantiles under different operating stages. (a) shows the predictions under nominal, in-control conditions prior to drift injection. (b) presents the predictions under drifted conditions before model updating, while (c) illustrates the predictions of the updated model under drifted conditions. The orange line and shaded regions represent the TiDE predictions under the optimal control actions generated by MPC. Note that the ground-truth realizations of melt pool temperature and depth under these optimal inputs are not accessible; the TiDE predictions can therefore only be compared with the resulting state trajectories observed during the process.}
    \label{fig:DED_local_pred}
\end{figure}

\begin{figure}[h!]
    \centering
    \includegraphics[width=0.9\linewidth]{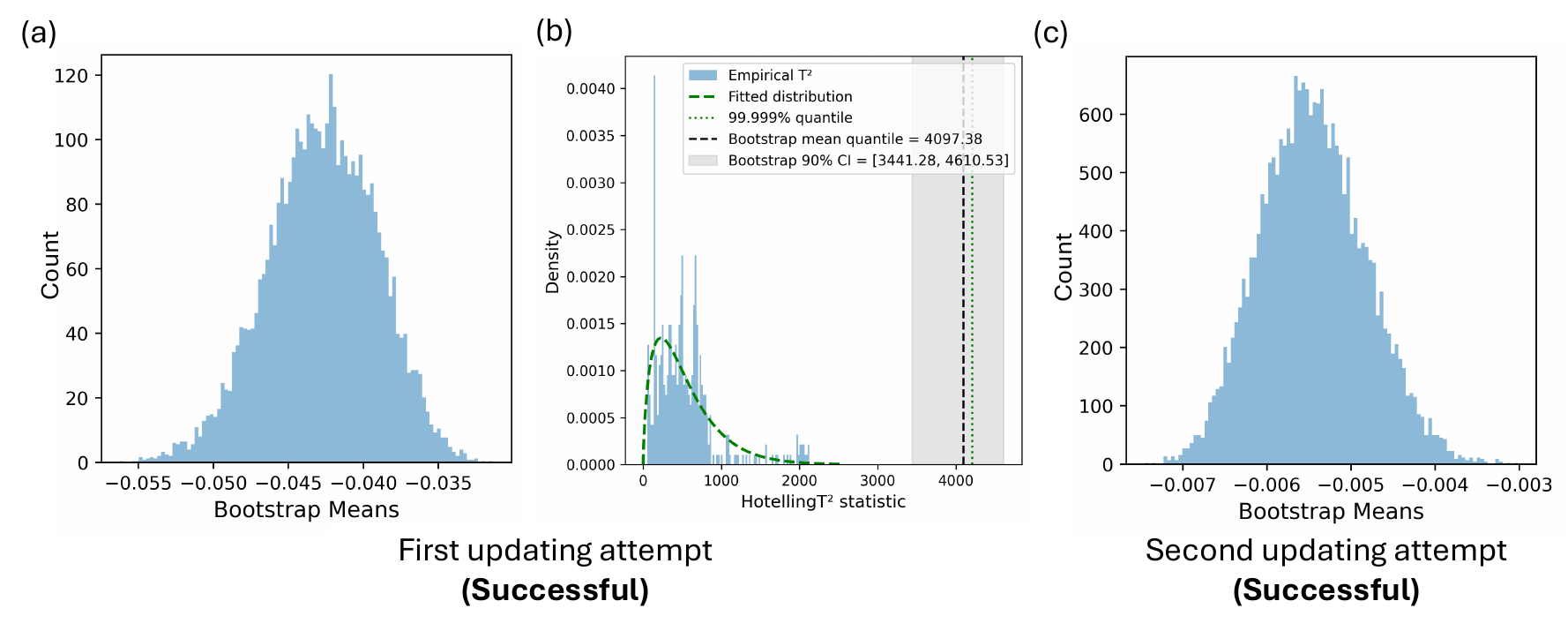}
    \caption{\textcolor{black}{(a)(c)Histograms of the means of the bootstrapped loss sequences at the validation stage, corresponding to the validation performed under the first and second updating attempts, respectively. (b) The estimated $T^2$ distribution after successful model updating. }}
    \label{fig:DED_stages}
\end{figure}

Compared to \texttt{No FT} and \texttt{LoRA}, \texttt{ADT-LoRA} offers several advantages. First, it achieves the best reference-tracking performance, attaining \textcolor{black}{$R^2 = 0.9162$ and $\mathrm{MAE} = 34.6688$K} against the temperature reference trajectory, while \texttt{No FT} and \texttt{LoRA} yield $R^2 = 0.9064$ and $0.8580$ with $\mathrm{MAE} = 56.5141$ and $43.2749$, respectively, reflecting larger tracking errors. The degradation is especially pronounced for \texttt{LoRA}, where stepwise fine-tuning introduces substantial instability even before the drift occurs, producing severe overshoot and poor tracking. Moreover, once the drift is injected, the iterative updates degrade the surrogate further, amplifying the fluctuations in both melt pool temperature and laser power. The improvement over \texttt{No FT} is more modest, but Figure~\ref{fig:DED_traj} shows that its melt pool temperature remains mostly above the reference, consistent with its larger $\mathrm{MAE}$, and its laser power exceeds that of \texttt{ADT-LoRA}, since the unadapted model is unaware of the material change that lowers the energy required to heat the melt pool. Overall, the proposed method delivers both superior tracking accuracy and greater stability under continuous model updating.

Figure~\ref{fig:DED_local_pred}(a)--(c) shows predictions before drift, under drift before updating, and after adaptation. Because the realized and multi-step-predicted trajectories use different inputs, the comparison is qualitative. In Figure~\ref{fig:DED_local_pred}(a) the quantiles capture the variability, and the median follows the trend. In Figure~\ref{fig:DED_local_pred}(b) the mismatched model underestimates temperature and depth, as the injected drift lowers solidus temperature, conductivity, and specific heat, raising melt-pool temperature and depth at fixed power. After successful model updating, the mismatch is mitigated, as shown in Figure~\ref{fig:DED_local_pred}(c): the predictive quantiles once again capture the fluctuations of the state trajectories, and reference-tracking performance is effectively restored.

Note that drift is also detected at time step 4871 despite the absence of additional perturbations. As shown in Figure~\ref{fig:DED_stages}(a)--(c), fine-tuning successfully improves the model predictions, as evidenced by \textcolor{black}{the paired differences between the validation loss from the two models}. This behavior stems from the inherently non-stationary DED process, where the nozzle height continuously increases while model updates rely on data from a localized region. \textcolor{black}{This is further supported by the observation that when the model extrapolates beyond the operating regime represented by most of the fine-tuning samples under non-stationary conditions, the resulting degradation in prediction accuracy can also trigger the drift detector. Although the proposed approach does not explicitly distinguish between concept drift and covariate drift, it can still detect distributional changes that lead to degraded predictive performance. Furthermore, while the drift detection framework relies on stationarity assumptions for control limit estimation, the demonstrated capability for repeated drift detection and model fine-tuning provides a potential means of mitigating this limitation by enabling continual adaptation across changing operating regimes, thereby supporting lifelong model adaptation.}

\subsubsection{Constrained Robust MPC}

\begin{figure}[h!]
    \centering
    \includegraphics[width=0.85\linewidth]{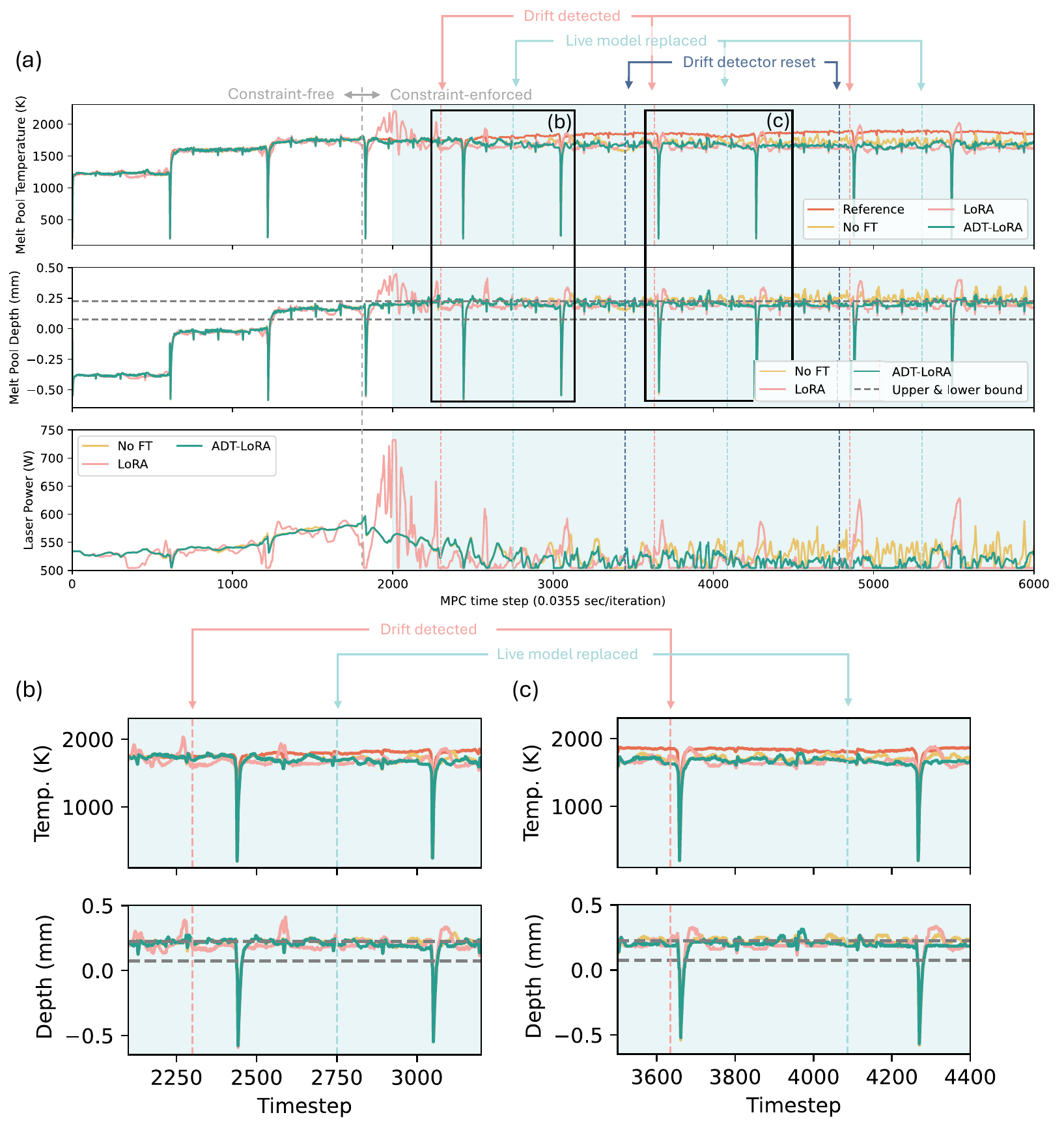}
    \caption{Resulting trajectories of melt pool temperature, melt pool depth, and laser power under constrained robust MPC, comparing \texttt{No FT}, \texttt{LoRA}, and the proposed \texttt{ADT-LoRA}. (a) shows the complete trajectories over all 10 layers, partitioned into constraint-free and constraint-enforced phases, with the melt pool depth bounds ($10\%$--$30\%$ dilution) indicated by dashed lines. (b) and (c) zoom in on the intervals where drift is detected and the live model is replaced. The time steps corresponding to drift detection, live-model replacement, and completion of the drift-detector reset are marked by dashed lines, and the shaded background indicates the time steps operated under drifted material properties.}
\label{fig:DED_constrained_traj}
\end{figure}

We revisit the same drift-injected process under constrained robust MPC, where chance constraints restrict the melt-pool depth to the $10\%$--$30\%$ dilution window, enforced from the fourth layer, using \texttt{ADT-LoRA}, with \texttt{No FT} and \texttt{LoRA} as benchmarks. With constraints enforced, drift is detected $\approx 300$ steps after the material perturbation begins.

Figure~\ref{fig:DED_constrained_traj}(a) compares the three approaches. \textcolor{black}{To compare the performance of the three methods, the constraint violation rate is calculated as the percentage of the number of time steps at which the melt pool depth exceeds its upper bound within the 4,200 time steps over which the constraints are enforced ($k=1800$ to $k=6000$). Since the lower-bound constraint is inactive during layer transitions, lower-bound violations are excluded from the calculation for all three methods to ensure a consistent comparison.} As a result, \texttt{ADT-LoRA} handles constraints best, keeping the depth below the upper bound at the lowest violation rate \textcolor{black}{($15.5\%$)} throughout the entire printing process, including the adaptation period. The zoomed views in Figure~\ref{fig:DED_constrained_traj}(b)--(c) show that after each model replacement, the depth is driven away from the active constraint, creating margin for fluctuations and reflecting successful adaptation of the constraint-enforcing quantiles. In contrast, although \texttt{LoRA} attains an $18\%$ violation rate, its iterative updates cause repeated overshoots at layer transitions and hold the depth overly far from the constraint, exhibiting over-conservativeness. Lastly, \texttt{No FT} underestimates the melt pool depth once drift occurs, incurring a $46\%$ violation rate. Thus, \texttt{ADT-LoRA} adapts more robustly while maintaining constraint satisfaction.

\begin{figure}[h!]
    \centering
    \includegraphics[width=1\linewidth]{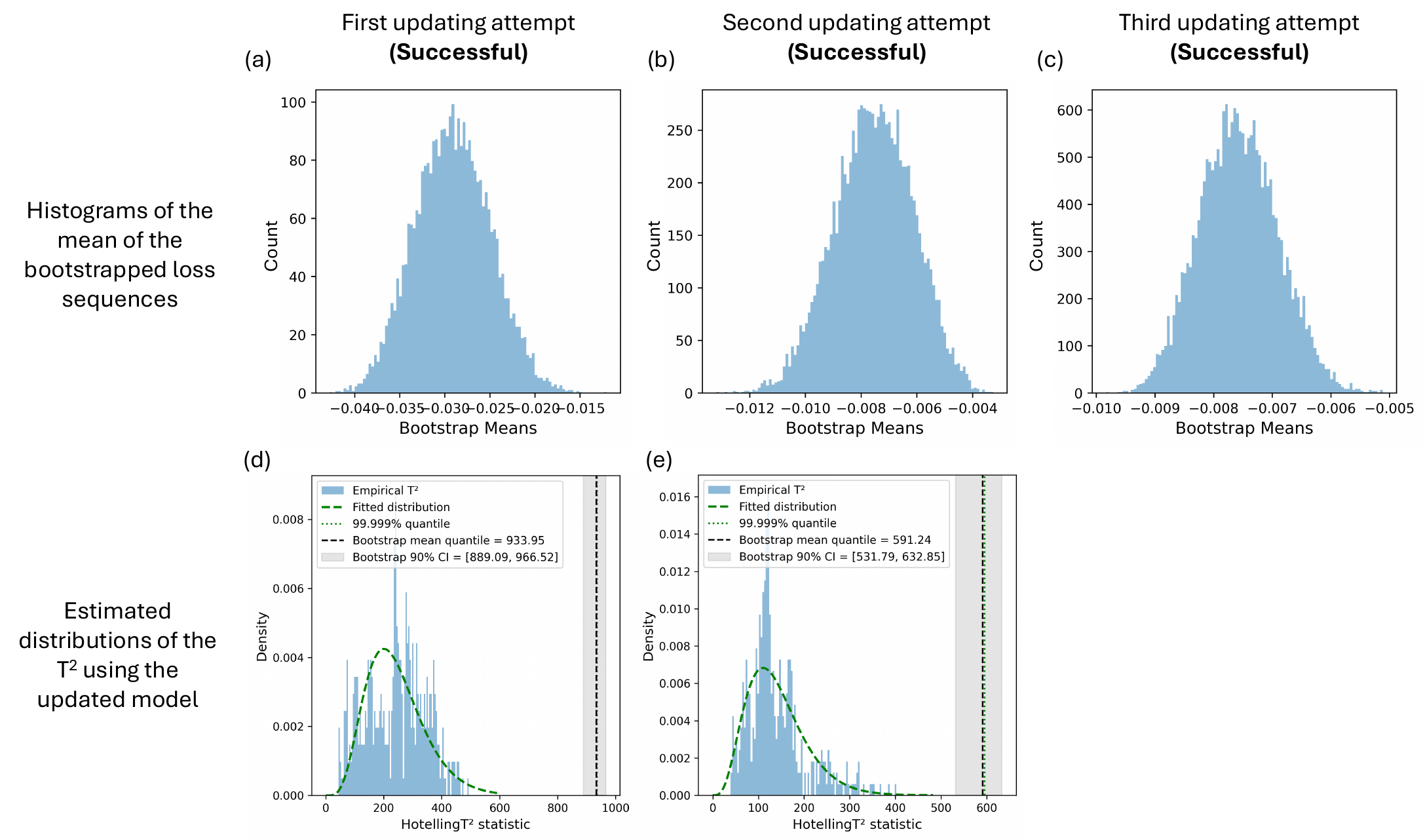}
    \caption{Histograms of the means of the bootstrapped loss sequences at the validation stage and the estimated $T^2$ distribution after successful model updating, for the constrained robust MPC case. The columns correspond to the three model-updating attempts performed over the print. The $T^2$ distribution is shown only for the resets that complete, as the print ends before the final reset concludes.}
\label{fig:DED_constrained_loss}
\end{figure}

\textcolor{black}{Figure~\ref{fig:DED_constrained_loss}(a)--(e) presents the histograms for the loss difference between live and idle models, and the fitted $T^2$ distributions used to reset the drift detector after successful updates.} Throughout the print, \texttt{ADT-LoRA} detects drift \textcolor{black}{three times and performs three successful fine-tunings.} The frequent detections are partly due to increased out-of-distribution (OOD) inputs, as the unconstrained offline training data differ from the operating distribution under constrained control, causing the detector to respond to both covariate and concept drift.

\textcolor{black}{It is worth noting that, despite experimenting with different learning rates (with the best one, which is identical to that used in the proposed method, reported here), \texttt{LoRA} still exhibits instability throughout the printing process, potentially due to overfitting during iterative fine-tuning. This behavior is observed in both unconstrained and constrained cases, where the melt pool temperature consistently overshoots at the beginning of new layers. These observations suggest that iterative fine-tuning may produce a model that performs well under recently encountered operating conditions but struggles to generalize to abrupt changes in system dynamics. In particular, the melt pool dynamics during in-layer printing differ substantially from those encountered at corners and layer transitions, explaining why \texttt{LoRA} performs normally during in-layer printing but exhibits degraded performance at these transitions. This issue may be further exacerbated by noisy feedback, as stepwise fine-tuning using stochastic gradient descent can overfit transient measurement fluctuations. Moreover, frequent model updates continuously alter the predictive mapping used by MPC, potentially introducing variability in the optimization landscape and resulting control decisions. For multi-step-ahead prediction, historical data are also collected under control decisions made using different model versions, which may further complicate subsequent fine-tuning. Consequently, improved local predictive accuracy does not necessarily translate into consistent closed-loop control performance. In contrast, \texttt{ADT-LoRA} employs batch fine-tuning triggered by detected drift, which may better preserve the predictive capabilities of the pretrained model across different operating regimes, thereby enabling smoother adaptation with reduced catastrophic forgetting. By maintaining a fixed predictive model between validated updates, \texttt{ADT-LoRA} may also improve the temporal consistency of MPC optimization. These findings further highlight the potential risks of indiscriminate iterative fine-tuning and underscore the importance of incorporating model validation into continual adaptation to ensure reliable predictive performance.}

\begin{figure}[h!]
    \centering
    \includegraphics[width=0.7\linewidth]{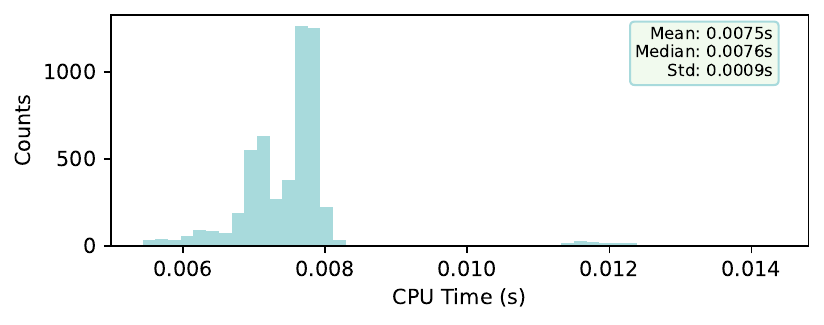}
    \caption{Histogram of the CPU time of the gradient-based drift detector}
    \label{fig:hist_cpu_time}
\end{figure}

Finally, Figure~\ref{fig:hist_cpu_time} shows the drift-detection computation time measured on an AMD Ryzen Threadripper PRO 3975WX CPU with a single NVIDIA A6000 GPU. Fine-tuning time is excluded, as it can be performed in parallel on separate hardware without affecting system operation. Gradient-based detection typically completes in under $0.008$~s, with a few instances reaching $0.012$~s, yielding an average detection time of $0.0075$~s per step, well within the 35.5ms control interval. Since the computation mainly involves a single forward and backward pass, its inference cost scales with the size and structure of the model. 

Taken together, the unconstrained and constrained DED studies show that the proposed framework transfers from the illustrative linear system to a complex, highly nonlinear, non-stationary manufacturing process in which the injected drift is embedded in the governing physics and perturbs the dynamics non-additively. \textcolor{black}{Nevertheless, although the proposed method outperforms iterative fine-tuning in both constrained and unconstrained cases, none of the investigated adaptation strategies fully restores or consistently maintains the desired uncertainty quantification performance and constraint violation rate. This highlights a fundamental challenge in online adaptation, where streaming data are often limited, noisy, and potentially biased under non-stationary operating conditions, making reliable adaptation to new concepts particularly difficult. Specifically, an inherent trade-off exists between the quality of model fine-tuning, which generally improves with the amount of available adaptation data, and the delay required to accumulate sufficient data and validate the updated model. Consequently, determining an appropriate validation criterion remains an open question. For instance, while requiring the updated model to achieve the nominal quantile coverage may provide a more stringent validation criterion for uncertainty quantification, satisfying this requirement may demand substantially more representative data or may even be infeasible under certain operating conditions. Conversely, relaxing the validation criterion may enable faster adaptation at the expense of predictive reliability. This dilemma makes the selection of validation criteria an important engineering decision that must balance adaptation quality, responsiveness, and operational safety.}

\textcolor{black}{On the other hand, rather than evaluating fine-tuning quality solely through the predictive performance of the surrogate model, an alternative validation criterion is whether the updated model can achieve satisfactory closed-loop control performance. However, directly validating control performance presents additional challenges, as empirical verification requires deploying the updated model and collecting feedback from its interactions with the physical system. Although simulation-based closed-loop evaluations may provide preliminary evidence, their reliability depends on the fidelity of the virtual environment, potentially undermining the objective of validating the model before deployment. Alternatively, theoretical safeguards, such as input-to-state stability, recursive feasibility, and convergence guarantees, may be established through model-based analysis under appropriate assumptions. Nevertheless, these properties do not necessarily reflect the actual control performance, particularly in terms of tracking accuracy and constraint satisfaction under model mismatch. Therefore, despite the demonstrated advantages of the proposed framework, establishing principled validation criteria that balance predictive fidelity, closed-loop control performance, and adaptation delay remains an important direction for future research.}

\section{Closure}
\label{sec:closure}

In this work, an online adaptation framework for Digital Twins is proposed. Building upon real-time decision-making algorithms via MPC, the framework incorporates a drift-detection mechanism to address \emph{when to update the model}, parameter-efficient fine-tuning to determine \emph{how to update the model}, and hypothesis testing for online validation to assess \emph{whether the update is successful}. A generalizable realization is demonstrated through the integration of \textcolor{black}{gradient}-based drift detection, LoRA-based adaptation, and the one-sided paired test with moving block bootstrapping, with neural networks serving as surrogate models. 

Two case studies verify the effectiveness of the proposed approach, both employing quantile-based robust MPC as the uncertainty-aware decision-making algorithm. In an illustrative example with injected abrupt uncertainty, the model is effectively fine-tuned during operation to accommodate both epistemic and aleatoric uncertainties. In an engineering case study on DED additive manufacturing, the framework adapts to incremental drift in material properties while maintaining control performance. These case studies collectively indicate that the proposed framework provides an effective approach for maintaining the performance of Digital Twins under previously unseen conditions, thereby enhancing their ability to handle unknown uncertainties.

Although the proposed framework establishes a pipeline for online adaptation, \textcolor{black}{several opportunities remain to further enhance its capabilities and practical applicability. First,} the proposed method introduces a trade-off in responsiveness, as the adaptation speed to drift is slower than that of stepwise, iterative model-updating methods commonly used in adaptive MPC with high-frequency decision requirements. To mitigate the impact of model mismatch during the adaptation phase, ad hoc compensation strategies such as offset-free control algorithms \cite{bamimore2021offset} can be incorporated. In addition, \cite{malikopoulos2026approximate} presents a promising perspective, demonstrating that accurate control performance may still be achieved with an imperfect model through alignment of the optimality conditions governing control actions. These approaches are suited as interim compensation strategies when model accuracy is degraded. \textcolor{black}{Second, as} the primary focus of this work is model adaptation from a fine-tuning perspective, control-oriented considerations, such as enforcing stability through Lyapunov-based constraints \cite{castellano2025data, wu2019real}, are not incorporated into the model-updating criteria. Consequently, the proposed framework does not provide formal guarantees on MPC stability. Thus, control-theoretic certification and validated model adaptation may play complementary roles, where the former supports closed-loop guarantees, while the latter helps maintain the Digital Twin's fidelity to an evolving physical system. \textcolor{black}{Third, several hyperparameters across the proposed framework, including the fine-tuning batch size, validation dataset size, block length for MBB, and upper-tail probability used for UCL estimation, have not been systematically optimized. Although jointly optimizing these hyperparameters could be formulated as a co-design problem, the large number of interacting parameters may introduce substantial computational costs. Therefore, practical compromises based on reasonable engineering assumptions are necessary and need to be introduced in domain applications. Lastly, although this work establishes a foundation for model adaptation through a gradient-based drift-detection approach, the potential for diagnosing parametric black-box models while introducing a degree of interpretability has not yet been demonstrated. With the capability to monitor individual parameters or groups of parameters, it can naturally be extended to enable autonomous adaptation and drift management in Digital Twins of complex systems, or even systems of systems. In particular, it provides a pathway to identify and localize drifting parameters, thereby focusing model updates on the most relevant subsets of the system.}

Such capabilities can further enhance the trustworthiness of Digital Twins by improving interpretability, especially when coupled with more transparent model representations such as differentiable causal modeling and analysis. Building on this foundation, methods such as reinforcement learning \cite{sun2024adaptive} can be incorporated to scale the framework to more complex decision-making settings, enabling the system to handle emergent behaviors through integrated reasoning and diagnosis within the surrogate model. This progression ultimately contributes to advancing machine intelligence in autonomous Digital Twin systems.

{\color{black}
\section*{Data Availability} 

The code and data used for this work are available upon request. 
}

\section*{Acknowledgement}
We are grateful for the grant support from the National Science Foundation's Engineering Research Center for Hybrid Autonomous Manufacturing: Moving from Evolution to Revolution (ERC-HAMMER), under Award Number EEC-2133630, and the Manufacturing ADvanced Electronics through Printing Using Biobased and Locally Identifiable Compounds (MADE PUBLIC) Future Manufacturing Research Grant Program under Award Number CMMI-2037026. Yi-Ping Chen acknowledges the Taiwan-Northwestern Doctoral Scholarship, and Seul Lee acknowledges the Samsung Electronics Doctoral Scholarship.


\addcontentsline{toc}{section}{Reference}
\bibliographystyle{asmems4}
\bibliography{main}

\titleformat{\section}[hang]{\fontsize{11pt}{0pt}\selectfont\bf}{\thesection}{0cm}{}

\titleformat{\section}[hang]{\centering\fontsize{11pt}{0pt}\selectfont\bf}{}{0.1cm}{}

\end{document}